\documentclass[sigconf, screen, nonacm]{acmart}
\renewcommand\footnotetextcopyrightpermission[1]{} %remove copyright
\usepackage{multirow}
\usepackage{colortbl}
\usepackage{pifont}
\usepackage{makecell}
\AtBeginDocument{%
  }

\setcopyright{acmlicensed}
\copyrightyear{2026}
\acmYear{2026}
\acmDOI{XXXXXXX.XXXXXXX}
\begin{document}

\title{Seg2Change: Adapting Open-Vocabulary Semantic Segmentation Model for Remote Sensing Change Detection}

\author{You Su}
\affiliation{%
  \institution{Xi'an Jiaotong University}
  \city{Xi'an}
  \state{Shaanxi}
  \country{China}
}
\email{yousu@stu.xjtu.edu.cn}

\author{Yonghong Song}
\affiliation{%
  \institution{Xi'an Jiaotong University}
  \city{Xi'an}
  \state{Shaanxi}
  \country{China}
}
\email{songyh@xjtu.edu.cn}

% \author{Jingtao Chen}
% \affiliation{%
%   \institution{Jiaxun Feihong Electrical Co., Ltd.}
%   \city{Beijing}
%   \country{China}
% }
% \email{chenjt@jiaxun.com}

\author{Jingqi Chen}
\affiliation{%
  \institution{Xi'an Jiaotong University}
  \city{Xi'an}
  \state{Shaanxi}
  \country{China}
}
\email{chenjingqi@stu.xjtu.edu.cn}

\author{Zehan Wen}
\affiliation{%
  \institution{Xi'an Jiaotong University}
  \city{Xi'an}
  \state{Shaanxi}
  \country{China}
}
\email{wenzehan@stu.xjtu.edu.cn}

% \author{Hao Lin}
% \affiliation{%
%   \institution{Xi'an Jiaotong University}
%   \city{Xi'an}
%   \state{Shaanxi}
%   \country{China}
% }
% \email{4123158007@stu.xjtu.edu.cn}

\renewcommand{\shortauthors}{Su et al.}

\begin{abstract}
Change detection is a fundamental task in remote sensing, aiming to quantify the impacts of human activities and ecological dynamics on land-cover changes. Existing change detection methods are limited to predefined classes in training datasets, which constrains their scalability in real-world scenarios. In recent years, numerous advanced open-vocabulary semantic segmentation models have emerged for remote sensing imagery. However, there is still a lack of an effective framework for directly applying these models to open-vocabulary change detection (OVCD), a novel task that integrates vision and language to detect changes across arbitrary categories. To address these challenges, we first construct a category-agnostic change detection dataset, termed CA-CDD. Further, we design a category-agnostic change head to detect the transitions of arbitrary categories and index them to specific classes. Based on them, we propose Seg2Change, an adapter designed to adapt open-vocabulary semantic segmentation models to change detection task. Without bells and whistles, this simple yet effective framework achieves state-of-the-art OVCD performance (+9.52 IoU$^c$ on WHU-CD and +5.50 mIoU$^c$ on SECOND). Our code is released at \href{https://github.com/yogurts-sy/Seg2Change}{https://github.com/yogurts-sy/Seg2Change}.

% under open-vocabulary settings on building, land-cover, and semantic change detection datasets. 

% 

\end{abstract}


% \ccsdesc[400]{Computing methodologies~Image segmentation}


% 123123
\begin{teaserfigure}
  \includegraphics[width=\textwidth]{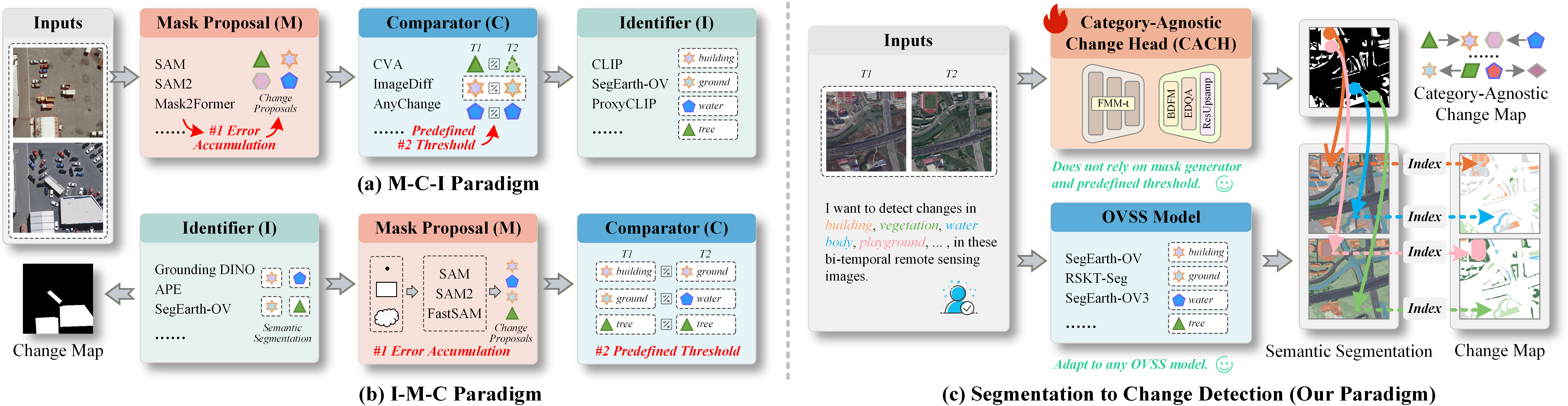}
  \caption{Previous paradigms vs. our paradigm. Previous paradigms rely on change proposals and are constrained by the segmentation performance of the proposal generator. Meanwhile, both paradigms (a–b) distinguish changed instances using a predefined threshold, which makes it difficult to accurately delineate change boundaries. In contrast, our paradigm, based on category-agnostic change maps, can effectively leverage more powerful open-vocabulary segmentation models.}
  \label{page}
\end{teaserfigure}

% \begin{figure}[t]
%   \centering
%   \includegraphics[width=\linewidth]{Figs/Paradigm4.pdf}
%   \caption{ Previous paradigm vs. our paradigm. Previous paradigms distinguish changed pixels using a fixed threshold, which makes it difficult to accurately delineate change boundaries. Meanwhile, paradigm (b), which relies on change proposals, is constrained by the segmentation performance of the proposal generator. In contrast, our paradigm based on category-agnostic change maps can readily leverage more powerful open-vocabulary segmentation models.}
%   \label{page}
% \end{figure}


\maketitle

\section{Introduction}
The Earth, our shared home, is increasingly shaped by human activities and climate change. Observing land-cover changes can help us better understand the Earth. Monitoring land-cover dynamics plays a critical role in advancing urban planning \cite{MISHRA2020133, doi:10.1080/014311600210128} and natural resource management \cite{SONG201426, 8641484, 6297453, Xu2019BuildingDD}. Within this context, change detection is an essential task that leverages bi-temporal or multi-temporal remote sensing images (RSIs) to detect both the spatial locations and semantic categories of land-cover transitions.

% With the rapid development of deep learning, mainstream change detection methods have demonstrated the superior performance of fully-supervised deep neural networks \cite{NIPS2012_c399862d, 7780459, 7478072}.

Within the supervised learning framework \cite{NIPS2012_c399862d, 7780459, 7478072}, accurate change detection generally relies on large-scale annotated datasets for training. In addition, these datasets usually cover only a small number of predefined categories, making them incompatible with open-vocabulary settings. Recently, the emergence of vision–language models (VLMs) has inspired efforts to transfer these models to the change detection task \cite{anychange, ucd_scm, instceg, li2025dynamic_earth}, leading to the concept of OVCD \cite{li2025dynamic_earth}. OVCD aims to localize and identify changes between bi-temporal images, where change categories are not predefined but can instead be described using arbitrary textual or semantic labels \cite{li2025dynamic_earth}. By introducing the ability to generalize beyond a fixed set of predefined change classes, OVCD extends traditional unsupervised change detection \cite{pca_km, cva, dcva} to more flexible and open-world scenarios. As illustrated in Fig. \ref{page}(a–b), existing OVCD methods generally follow two paradigms: \textbf{M}ask proposal–\textbf{C}omparator–\textbf{I}dentifier (\textbf{M}–\textbf{C}–\textbf{I}) and \textbf{I}dentifier–\textbf{M}ask proposal–\textbf{C}omparator (\textbf{I}–\textbf{M}–\textbf{C}). (1) \textbf{M}–\textbf{C}–\textbf{I} \cite{ucd_scm,anychange,li2025dynamic_earth}: A mask generator (\textit{e.g.}, SAM \cite{sam1} or Mask2Former \cite{cheng2021mask2former}) first generates all candidate proposals from bi-temporal remote sensing images. A comparator (\textit{e.g.}, CVA \cite{cva} or AnyChange \cite{anychange}) then evaluates change occurrence within these regions, and an identifier (\textit{e.g.}, CLIP \cite{clip} or SegEarth-OV \cite{li2025segearthov}) finally assigns semantic labels to the detected changes. (2) \textbf{I}–\textbf{M}–\textbf{C} \cite{instceg, li2025dynamic_earth}: This paradigm first identifies all proposals of interest, converts them into instance masks through the mask proposal generator, and then compares the masks to determine whether a change has occurred.

While encouraging results have been observed, these paradigms have the following issues: (1) Both the \textbf{M}–\textbf{C}–\textbf{I} and \textbf{I}–\textbf{M}–\textbf{C} paradigms rely on a mask generator (\textit{e.g.}, SAM) to produce change proposals. As a result, subsequent change determination is inevitably affected by the accumulation of segmentation errors from SAM. (2) UCD-SCM employs OTSU \cite{4310076} to estimate a global threshold based on feature distances. However, a linear threshold is often insufficient to clearly distinguish changed pixels. Although the subsequent DynamicEarth framework upgrades the comparator to DINOv2 \cite{dinov2}, it still determines whether a proposal has changed based on a predefined threshold. (3) According to the results reported by DynamicEarth, the \textbf{M}–\textbf{C}–\textbf{I} performs well in land-cover change detection, whereas \textbf{I}–\textbf{M}–\textbf{C} achieves better performance in building change detection. However, the applicability of these two paradigms is limited, requiring frequent switching to accommodate different tasks. We advocate a simple yet effective paradigm that can guide VLMs to achieve optimal performance across multiple tasks, including building, land-cover, and semantic change detection.

To this end, we propose \textbf{Seg2Change}, which serves as a competitive alternative for adapting open-vocabulary semantic segmentation (OVSS) models to change detection. The core idea is to eliminate error propagation inherent in mask proposal and avoid reliance on predefined thresholds for change decisions. As illustrated in Fig. \ref{page}(c), we instead identify a category-agnostic change map, enabling changes to be extracted directly from the segmentation masks of bi-temporal images. First, we construct a category-agnostic change detection dataset, \textbf{CA-CDD}, which expands the scope of change categories by extending category-constrained datasets (\textit{e.g.}, WHU-CD \cite{whucd} focuses on buildings, and DSIFN \cite{DSIFN} focuses on land-cover changes) to arbitrary categories. In particular, we design a category-agnostic change head (\textbf{CACH}) with feature difference enhancement and calibration. By fusing bi-temporal feature differences and performing efficient difference retrieval with a sliding-window strategy, the influence of pseudo-changes on real changes is effectively suppressed. CACH generates high-quality category-agnostic change maps without relying on any predefined thresholds. Finally, the images and target categories of interest are fed into an OVSS model to obtain the corresponding semantic segmentation maps, from which the change detection results for specific categories are indexed using the category-agnostic change maps.

Extensive experiments on three change detection tasks in open-vocabulary settings (\textit{i.e.}, building, land-cover, and semantic change detection) and six widely recognized datasets (\textit{i.e.}, WHU-CD \cite{whucd}, LEVIR-CD \cite{levircd}, DSIFN \cite{DSIFN}, CLCD \cite{clcd}, SC-SCD \cite{SC-SCD}, and SECOND \cite{second}) show that Seg2Change achieves state-of-the-art performance across the board. Notably, despite being simpler than the more complex \textbf{M}–\textbf{C}–\textbf{I} and \textbf{I}–\textbf{M}–\textbf{C} paradigms, the proposed method achieves an optimal trade-off between open-vocabulary change detection performance and inference efficiency.

In summary, the contributions of this work are as follows: 
\begin{itemize}
\item We revisit previous open-vocabulary change detection approaches from the perspective of instance change thresholds and introduce a new paradigm termed Seg2Change, which provides a simple yet effective approach for bi-temporal open-vocabulary change detection.

\item By constructing a category-agnostic change detection dataset, CA-CDD, and introducing a category-agnostic change head with feature differences fusion and calibration, Seg2Change can be adapted to any OVSS model.

\item Seg2Change achieves state-of-the-art performance with an efficient OVSS model (\textit{i.e.}, SegEarth-OV3) across multiple benchmarks. Moreover, both quantitative and qualitative analyses validate the efficacy of the proposed paradigm, further solidifying the effectiveness of our approach.
\end{itemize}

\section{Related Work}
\subsection{Vision-Language Models}
% VLMs integrate visual and textual understanding, and early studies including CLIP and ALIGN pioneered large-scale contrastive pretraining using paired image–text data. BLIP and Flamingo extended VLM capabilities to captioning, VQA, and multimodal dialogue, strengthening alignment between vision and language. GroundingDINO bridges detection and language understanding, allowing objects to be localized and recognized through textual prompts. Building on this foundation, multimodal foundation models such as DINO-X and Florence-2 have been developed with capabilities in image–text dialogue and segmentation. APE and SegEarth-OV further enhance pixel-wise segmentation and extend visual perception models to multiple tasks, including detection and localization. RemoteCLIP pioneers the adaptation of VLMs to remote sensing by collecting extensive domain-specific data to support downstream tasks. RSKT-Seg further advances open-vocabulary segmentation by integrating DINOv2 and RemoteCLIP features, enabling effective domain knowledge transfer for remote sensing imagery. SegEarth-OV3 produces high-quality segmentation maps by integrating SAM3-based semantic and instance heads, and further filters low-confidence categories using presence scores.

\begin{figure}[t]
  \centering
\includegraphics[width=0.95\linewidth]{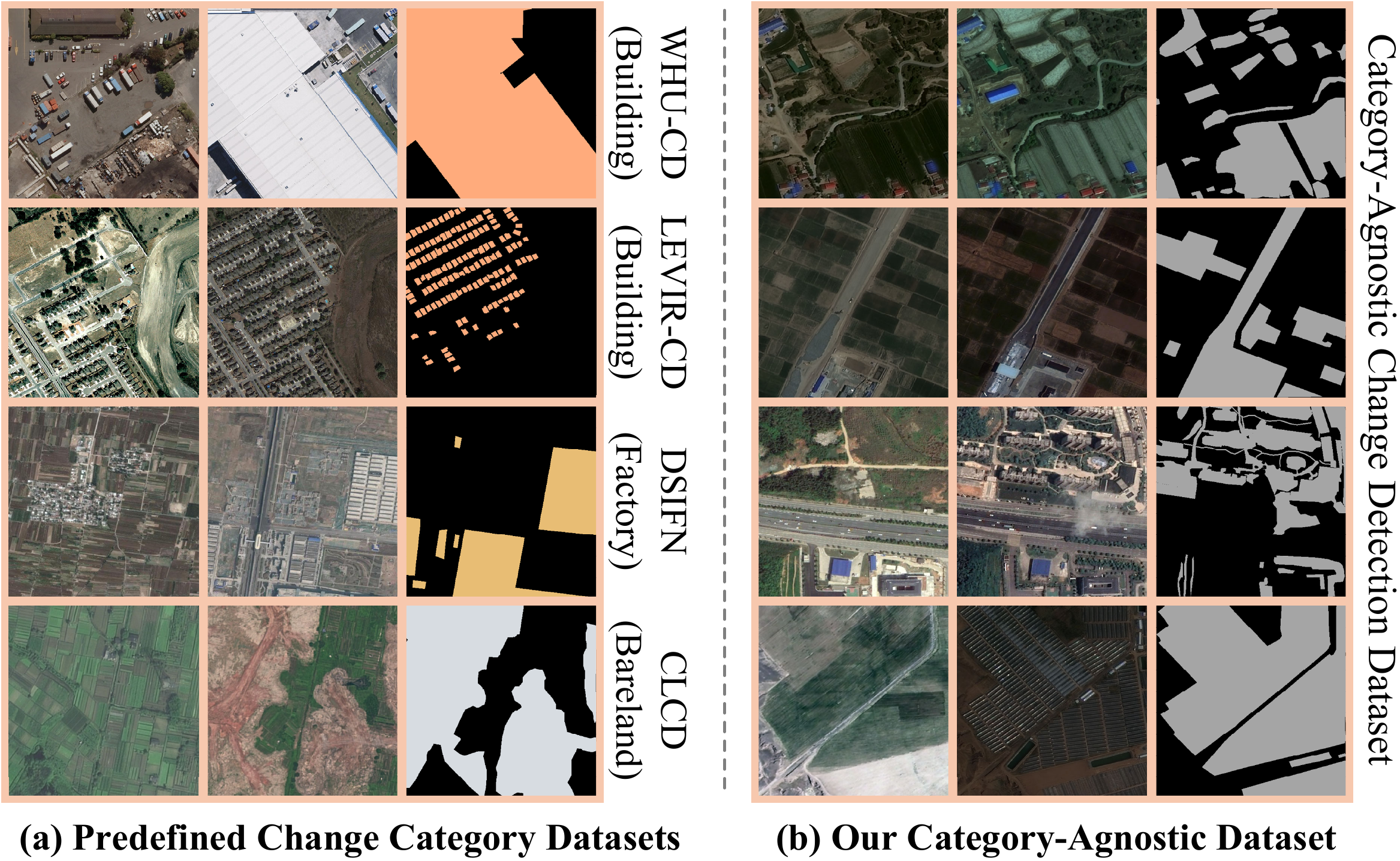}
  \caption{Visual comparison between predefined change category datasets (a) and our category-agnostic change detection dataset (b). The former focuses only on limited categories.}
\label{datasets1}
\end{figure}

\begin{figure}[t]
  \centering
\includegraphics[width=0.95\linewidth]{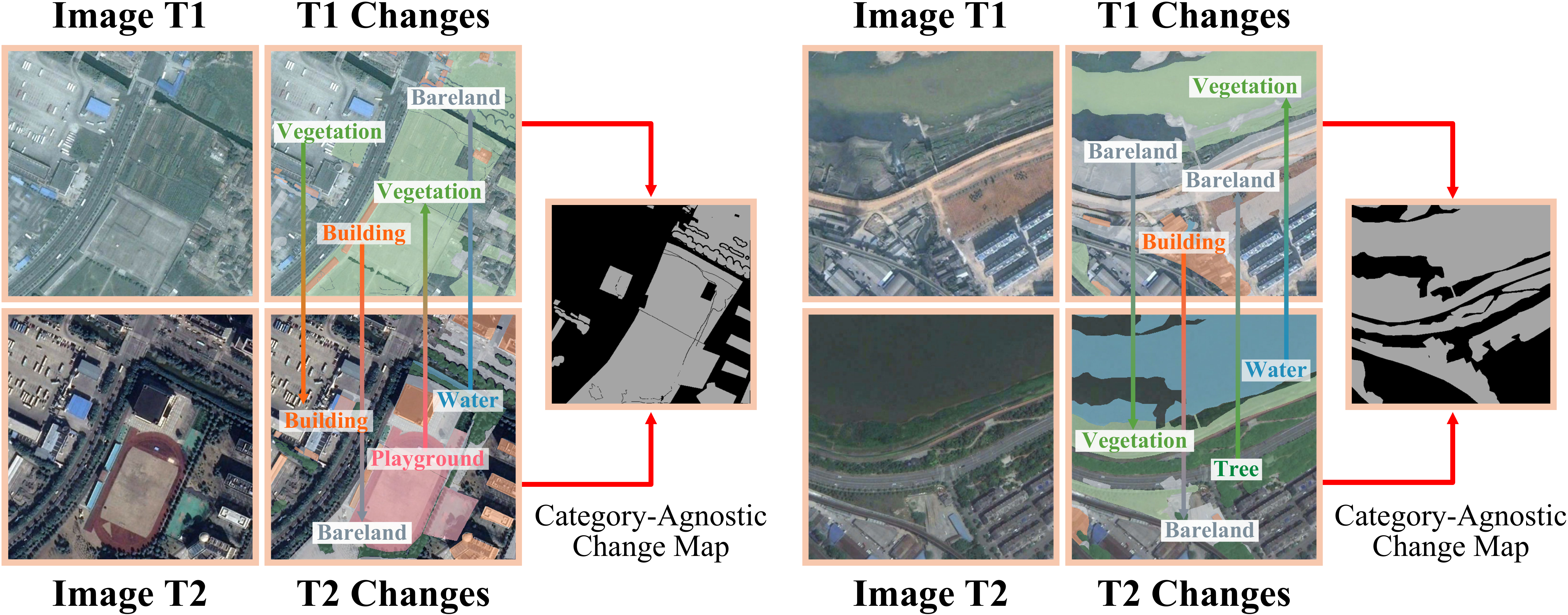}
  \caption{Annotation process of the category-agnostic change detection dataset. Regions where the category changes between the bi-temporal images are annotated as change maps. The actual recorded categories are not limited to those shown in the figure.}
\label{datasets2}
\end{figure}

Vision–Language Models (VLMs) are multimodal models capable of jointly understanding visual and semantic information. Pioneering works, such as CLIP \cite{clip} and ALIGN \cite{pmlr-v139-jia21b}, employ web-scale image–text pairs for contrastive learning, laying the groundwork for VLM pre-training. Subsequently, BLIP \cite{li2022blip} and Flamingo \cite{Flamingo} further explore applications, including image captioning, visual question answering, and dialogue generation. Grounding DINO \cite{liu2023grounding} combines object detection with natural language understanding to achieve text-driven object grounding and recognition. Further, APE \cite{APE} and SegEarth-OV \cite{li2025segearthov} enhance pixel-level semantic segmentation by leveraging vision-aware models capable of performing multiple tasks, including object detection, segmentation, and localization. In the remote sensing community, RemoteCLIP \cite{liu2024remoteclip} is the first vision–language model tailored for remote sensing imagery, built upon large-scale datasets to support a wide range of downstream applications. Recently, RSKT-Seg \cite{rskt_seg} integrates features from DINOv2 \cite{dinov2} and RemoteCLIP \cite{liu2024remoteclip}, effectively transferring remote sensing knowledge to open-vocabulary segmentation and improving the performance of remote sensing image segmentation tasks. SegEarth-OV3 \cite{li2025segearthov3} obtains high-quality segmentation maps by combining the semantic segmentation head and instance head of SAM3 \cite{carion2025sam3segmentconcepts}, and further filters out low-confidence categories in the scene using presence scores, achieving new advances in open-vocabulary segmentation for remote sensing.

\subsection{Unsupervised \& Open-Vocabulary CD}

Unsupervised change detection operates without annotated data and can generally be divided into four categories: image algebra methods \cite{cva, 4039609, dcva, DU2020278}, image transformation methods \cite{NIELSEN19981, Deng01082008, 5196726, 6553145, CHEN202399, Chen2023Exchange}, post-classification comparison methods \cite{XIAN20091133, HUSSAIN201391, GILYEPES201677, 4539638, 6841049, 1036009}, and foundation-model-based methods \cite{ucd_scm, anychange, instceg, li2025dynamic_earth}. A representative image algebra method is change vector analysis (CVA) \cite{cva}, which measures changes through direct comparison of spectral bands across bi-temporal images. Later, DCVA \cite{dcva} incorporates convolutional neural network (CNN) representations to enhance the discrimination of change-related features. Within image transformation approaches, I3PE \cite{Chen2023Exchange} serves as a representative method that generates pseudo bi-temporal image pairs and associated change labels through patch exchange from single-temporal images, enabling effective utilization of large-scale unlabeled and unregistered remote sensing data. Recently, open-vocabulary change detection has enabled category-specific change identification through the integration of VLMs. AnyChange \cite{anychange} utilizes latent matching of bi-temporal SAM image features, combined with manually annotated points, to cluster change categories of interest. UCD-SCM \cite{ucd_scm} proposes a novel piecewise semantic attention (PSA) mechanism that leverages semantic representations from FastSAM \cite{zhao2023fast} to reduce pseudo-changes. Inst-CEG \cite{instceg} extracts proposals whose categories change between bi-temporal images through a change event generation (CEG) to generate change maps. DynamicEarth \cite{li2025dynamic_earth} proposes two open-vocabulary change detection paradigms, namely \textbf{M}–\textbf{C}–\textbf{I} and \textbf{I}–\textbf{M}–\textbf{C}. The \textbf{M}–\textbf{C}–\textbf{I} paradigm first identifies all potential changes and then classifies them, whereas the \textbf{I}–\textbf{M}–\textbf{C} paradigm first detects all objects of interest and subsequently determines whether their categories have changed.

Through in-depth analysis and experimentation on existing approaches, we observe that current OVCD paradigms still rely on instance-level change estimation, making it difficult to determine reliable thresholds for identifying changed instances. Meanwhile, existing paradigms remain dependent on mask generators to provide change proposals. Subsequent segmentation and identification of change areas are susceptible to error propagation originating from the mask generator. Motivated by the rapid progress of open-vocabulary semantic segmentation models in remote sensing, we design a change detection adapter for such models. By explicitly decoupling segmentation from change reasoning, this design allows us to fully leverage the continually improving capabilities of open-vocabulary semantic segmentation models.

\begin{figure*}[t]
\centering
\includegraphics[width=0.98\textwidth]{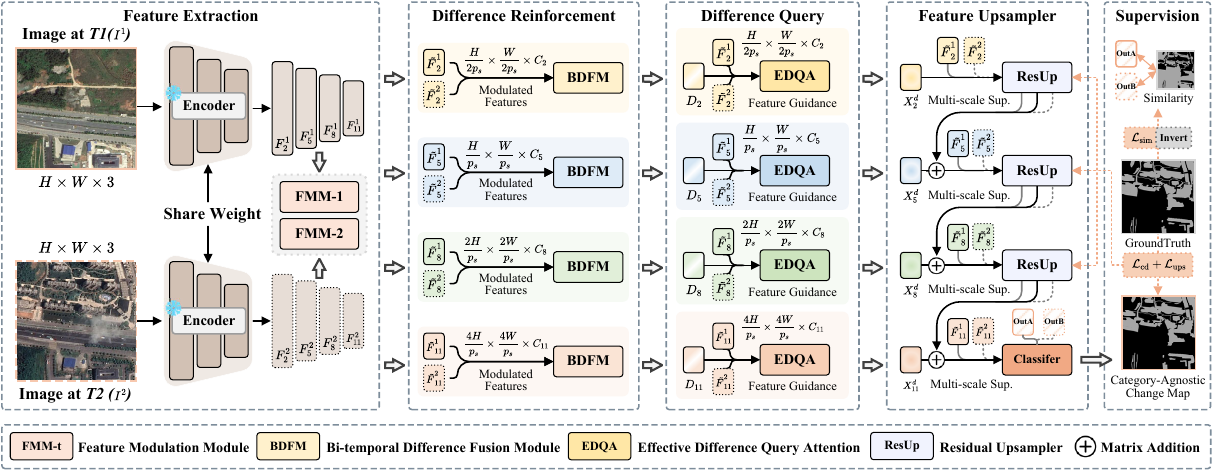}
\caption{Category-Agnostic Change Head (CACH). Bi-temporal image features are extracted and processed through feature differences fusion and difference query modules to obtain calibrated discrepancies, which are then concatenated across multiple dimensions via our residual upsampler to produce the final category-agnostic change map.}
\label{CACH}
\end{figure*}
\begin{figure*}[t]
  \centering
  \includegraphics[width=0.98\textwidth]{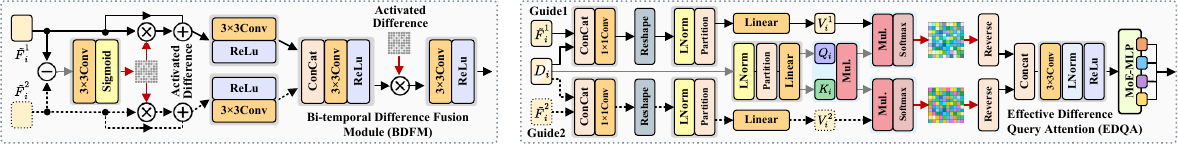}
  \caption{Bi-temporal Difference Fusion Module (BDFM) and Effective Difference Query Attention (EDQA) Module.}
\label{BDFM}
\end{figure*}

\section{Methodology}

\subsection{Analysis of Previous OVCD Paradigms}
Previous studies \cite{ucd_scm, anychange, instceg, li2025dynamic_earth} have constructed OVCD frameworks based on existing foundation models, achieving open-vocabulary capability to some extent. However, they still rely on instance-level change decision. Specifically, a mask generator (\textit{e.g.}, the SAM family \cite{sam1, sam2}) is first applied to generate change proposals, followed by a comparator (\textit{e.g.}, CVA \cite{cva} or OTSU \cite{4310076}) to perform change judgment. Change judgment is performed by comparing the cosine similarity $D$ of bi-temporal features within each proposal against a predefined threshold, as follows:
\begin{align}
D\left(\mathbf{p}\right)=-\frac{\mathbf{z^{1}}[\mathbf{p}]}{\left\|\mathbf{z^{1}}[\mathbf{p}]\right\|_{2}} \cdot \frac{\mathbf{z^{2}}[\mathbf{p}]}{\left\|\mathbf{z^{2}}[\mathbf{p}]\right\|_{2}},
\beta = \text{cos}(\theta \cdot \frac{\pi}{180}),
\end{align}
where $\mathbf{z}^{t}[\mathbf{p}]$ represents the average feature embedding of the bi-temporal images $I^{t}$, $t \in \{1,2\}$, obtained from an image encoder (\textit{e.g.}, ResNet \cite{7780459}, SAM \cite{sam1}, or DINO \cite{dinov2}) at the location corresponding to proposal $\mathbf{p}$. The angle $\theta$ is a predefined hyperparameter. After conversion to radians, its cosine value serves as the threshold $\beta$ \cite{anychange}. Proposals with similarity scores exceeding the threshold $\beta$ are classified as change masks and the rest are dropped.

These paradigms rely on mask generators to produce change proposals. Although they offer strong generalization and fine-grained segmentation, they often perform uncontrolled segmentation, splitting large objects into multiple fragments. Moreover, the predefined change threshold $\beta$ does not generalize well across instances and typically requires careful manual tuning. From an efficiency standpoint, these paradigms must traverse all proposals to determine changes, and such traversal cannot be parallelized.

% Reliance on predefined instance-level segmentation paradigms leads to error accumulation during the segmentation process.
% As shown in Table 1, we analyze and compare the inference time and memory consumption of our proposed pixel-based OVCD paradigm with the instance-based paradigm. It can be observed that, due to model stacking and the inherently non-parallelizable instance traversal process, the instance-based OVCD paradigm consumes substantially more computational resources.

% We extend open-vocabulary change detection to the pixel level, substantially enhancing the capability to detect changes without predefined change categories. Meanwhile, the pixel-level paradigm moves beyond simple model stacking by first identifying pixel-wise semantic category changes and then guiding them toward specified target categories. In terms of efficiency, pixel-based OVCD exhibits significantly lower resource consumption than previous approaches.

\subsection{A Category-Agnostic CD Dataset}
Observing geological changes helps improve the Earth's observation and environmental monitoring \cite{1395984}. In this context, change detection is a core task that analyzes bi-temporal remote sensing images to determine where and what types of changes occur. However, existing change detection datasets typically focus on a limited set of change categories. For example, WHU-CD \cite{whucd} and LEVIR-CD \cite{levircd} concentrate on building changes on the Earth's surface, whereas DSIFN \cite{DSIFN} and CLCD \cite{clcd} focus on land-cover changes. SYSU-CD \cite{sysu} extends the change categories to some extent, including newly built urban buildings, vegetation changes, and road expansion. Despite this extension, the considered categories are still largely constrained to changes related to urbanization. The closest concept to a category-agnostic change map is the binary change map in semantic change detection datasets \cite{second, SC-SCD, liu2025jl1,rs15092464}. Nevertheless, these change maps are restricted to a limited set of predefined categories. For instance, SECOND \cite{second} focuses on bareland, buildings, trees, vegetation, playgrounds, and water, while changes like roads or building renovations are not included.

In this study, we collect bi-temporal images and coarse change maps with limited categories from the training sets of semantic change detection datasets, including SECOND \cite{second}, JL1-CD \cite{liu2025jl1}, and CNAM-CD \cite{rs15092464}. We further re-annotate these coarse change maps to extend them to category-agnostic changes. As shown in Fig. \ref{datasets1}, our CA-CDD dataset broadens the scope of existing datasets to an open and unconstrained range. Fig. \ref{datasets2} illustrates the annotation process for category-agnostic change maps. Notably, we evaluate our method on the test set of the SECOND dataset. No images from the SECOND test set are included in CA-CDD.

\subsection{Semantic Segmentation to Change Detection}
\textbf{Architecture Overview.}
The extraction process of the category-agnostic change map is illustrated in Fig.~\ref{CACH}. Given a pair of bi-temporal images $I^{1}, I^{2} \in \mathbb{R}^{H \times W \times 3}$, we first employ DINOv2~\cite{dinov2} to extract multi-scale features. Specifically, features from selected encoder layers are denoted as $F^{t}_{i} \in \mathbb{R}^{\frac{H}{p_s} \times \frac{W}{p_s} \times C_{\text{dim}}}$, where $i \in \{2, 5, 8, 11\}$ and timestamp $t \in \{1, 2\}$. Here, $C_{\text{dim}}$ and $p_s$ denote the feature dimension and patch size, respectively. The FMM transforms $F^{t}_{i}$ into pyramid modulated features $\tilde{F}_i^t \in \mathbb{R}^{\frac{2^{k-1}H}{p_s} \times \frac{2^{k-1}W}{p_s} \times C_i}$, where $k$ indexes the feature scale. Subsequently, the bi-temporal difference features are extracted and refined by BDFM and EDQA. These calibrated features are then fed into a residual upsampling module (ResUp) to produce the category-agnostic change map $M_{\text{ca}}$. In parallel, $I^1$ and $I^2$, together with the textual prompt $\mathbf{x}_{\text{text}}$, are input into an OVSS model to obtain semantic maps $M^{1}$ and $M^{2}$. The final change map is derived by combining semantic predictions with the category-agnostic change map as:
\begin{align}
{M}_{\text{ch}} = {M}_{\text{ca}} \cdot {M}^1 + M_{\text{ca}} \cdot {M}^2.
\end{align}

% we employ convolutional activations to extract spatial changes from the modulated features $\tilde{F}_i^1$, $\tilde{F}_i^2$ to generate $i$-th layer difference attention:

\textbf{Bi-temporal Difference Fusion Module (BDFM).}
A careful review of previous change detection methods \cite{10234560, 10285430, li2022cd, 10965808, 10504297, jiang2025LGCANet, 10034787} reveals that difference enhancement based solely on arithmetic operations \cite{9491802, 10185449, 10285430} and cascade convolutional blocks \cite{9883686, Yuan31122022} neglects the guidance of bi-temporal features on difference features. To address this, we propose BDFM, as illustrated in Fig. \ref{BDFM}. BDFM generates feature difference activations to emphasize regions in the bi-temporal images that correspond to these differences as:
\begin{align}
Att_i = \sigma(\mathrm{Conv_{3 \times 3}} * |\tilde{F}_i^1 - \tilde{F}_i^2|),
\end{align}
where $\mathrm{Conv_{3 \times 3}}$ is a 2D convolution with a kernel size of $3 \times 3$, $*$ represents the convolution operation, and $\sigma$ denotes the $\texttt{Sigmoid}$ function. To inject richer spatial discrepancy representations and facilitate precise localization of change regions, we incorporate the modulated features into the activated features.

Specifically, we fuse the activated feature differences from the bi-temporal images. By employing difference attention $Att_i$, the discrepancy regions are further enhanced:
\begin{align}
X_i^t = \gamma\mathrm{Conv_{3 \times 3}} &* (\tilde{F}_i^t + Att_i \cdot \tilde{F}^t_i), \\ D_i = \gamma\mathrm{Conv_{3 \times 3}} * (\gamma&\mathrm{Conv_{3 \times 3}} * (X_i^1 || X_i^2) \cdot Att_i),
\end{align}
where $X_i^t$ are bi-temporal feature differences, and $D_i$ denotes the fused feature differences at the $i$-th layer. $\gamma$ and $||$ are, respectively, the $\texttt{ReLu}$ activation function and the concatenation operation. Nevertheless, the fused feature differences still contain non-negligible noise. Since these noisy responses are not consistently reflected in the original modulated features, a subsequent calibration step is necessary to suppress noise and improve robustness.

\textbf{Effective Difference Query Attention (EDQA).} 
Owing to differences in acquisition time and imaging sensors of bi-temporal images, the captured imagery is strongly influenced by task-irrelevant factors, including seasonal shifts, illumination variations, and structural modifications \cite{rs15082092, rs12101688, LI2019197}. Moreover, changed regions typically occupy only a small portion of the scene, leading to severe foreground–background imbalance. Although BDFM enhances discrepancy representations, it inevitably introduces pseudo-change artifacts due to feature amplification. As depicted in Fig. \ref{BDFM}, we propose EDQA, which explicitly calibrates noisy difference features via cross-temporal guidance and adaptive expert modeling.

% The modulated feature guidance $\tilde{G}_i^t$ are partitioned into $N_w$ sliding windows of size $S_w \times S_w$ and reshaped into $\mathbb{R}^{N_w \times S_w \times S_w \times C_i}$, followed by linear projections $\phi_{v_i}^t$ to obtain $V^1_i$ and $V^2_i$, respectively. Similarly, the feature differences $D_i$ are partitioned into sliding windows and linearly projected into the query ($Q_i$) and key ($K_i$) via $\phi_{q_i}$ and $\phi_{k_i}$. This process mitigates the influence of pseudo-change artifacts introduced during feature difference fusion:

First, the feature differences $D_i$ are concatenated with the modulated features $\tilde{F}_i^t$, respectively, and mapped through a $1 \times 1$ convolution to feature guidance $\tilde{G}_i^t$. To efficiently model cross-temporal dependencies, we adopt a sliding-window attention strategy~\cite{9710580}. Specifically, $\tilde{G}_i^t$ is partitioned into $N_w$ windows of size $S_w \times S_w$ and reshaped into $\mathbb{R}^{N_w \times S_w \times S_w \times C_i}$. Linear projections $\phi_{v_i}^t$ are applied to obtain value embeddings $V_i^t$. Meanwhile, the difference features $D_i$ are partitioned in the same manner and projected into query and key embeddings ($Q_i$ and $K_i$) via $\phi_{q_i}$ and $\phi_{k_i}$, respectively. The calibrated difference features are computed as:
\begin{align}
\tilde{D}_i^t=\phi_{\text{proj}}(\mathrm{Softmax}((\phi_{q_i}(D_i) \cdot \phi_{k_i}(D_i)^\top) / \sqrt{d_i} + b_i)\cdot \phi_{v_i}^t(\tilde{G}_i^t)),
\end{align}
where $d_i$ denotes the attention head dimension, and $b_i$ represents the relative positional bias. This operation leverages discrepancy-aware queries to selectively aggregate reliable information from modulated features. The calibrated bi-temporal features are then reconstructed by reversing the window partition and fused as:
\begin{align}
\tilde{D}_i = \gamma\mathrm{Conv_{3 \times 3}} * (\gamma\mathrm{Conv_{3 \times 3}} * (\tilde{D}_i^1 || \tilde{D}_i^2)),
\label{eq:query_fuse}
\end{align}
 
Owing to differences in dataset sources, variations in imaging conditions and seasonal characteristics inevitably arise. To accommodate bi-temporal images from heterogeneous sources, we augment the MLP with a Mixture-of-Experts (MoE) \cite{Jia_M4oE_MICCAI2024}. The calibrated bi-temporal feature differences $\tilde{D}_i$ are utilized to estimate expert weights and to compute the outputs of each expert:
\begin{align}
\text{weight}_G = \text{Softmax}(W_g \cdot \tilde{D}_i &+ \mathbf{b}_g), \\ \text{expert}_j(\tilde{D}_i) = W_{o,j} \cdot \text{GELU}(W_{h,j} \cdot \tilde{D}_i &+ \mathbf{b}_{h,j}) + \mathbf{b}_{o,j},
\end{align}
where $\text{weight}_G$ denotes the weight assigned to each expert, and $\text{expert}_j(\tilde{D}_i)$ represents the output of expert-$j$ given the feature difference $\tilde{D}_i$. $W_g$ and $\mathbf{b}_g$ are the learnable parameters used to compute the expert weights. $W_{h,j}$, $\mathbf{b}_{h,j}$, and $W_{o,j}$, $\mathbf{b}_{o,j}$ are the parameters of the two linear mapping layers for the output of expert-$j$.

Finally, the weighted outputs of the $N_e$ experts are aggregated to produce the refined difference feature $X_i^d$, as described in Eq. (\ref{eq:xd}):
\begin{align}
X_i^d = \textstyle\sum_{j=1}^{N_e} \text{weight}_j \cdot \text{expert}_j(\tilde{D}_i).
\label{eq:xd}
\end{align}

\begin{figure*}[t]
  \centering
  \includegraphics[width=\textwidth]{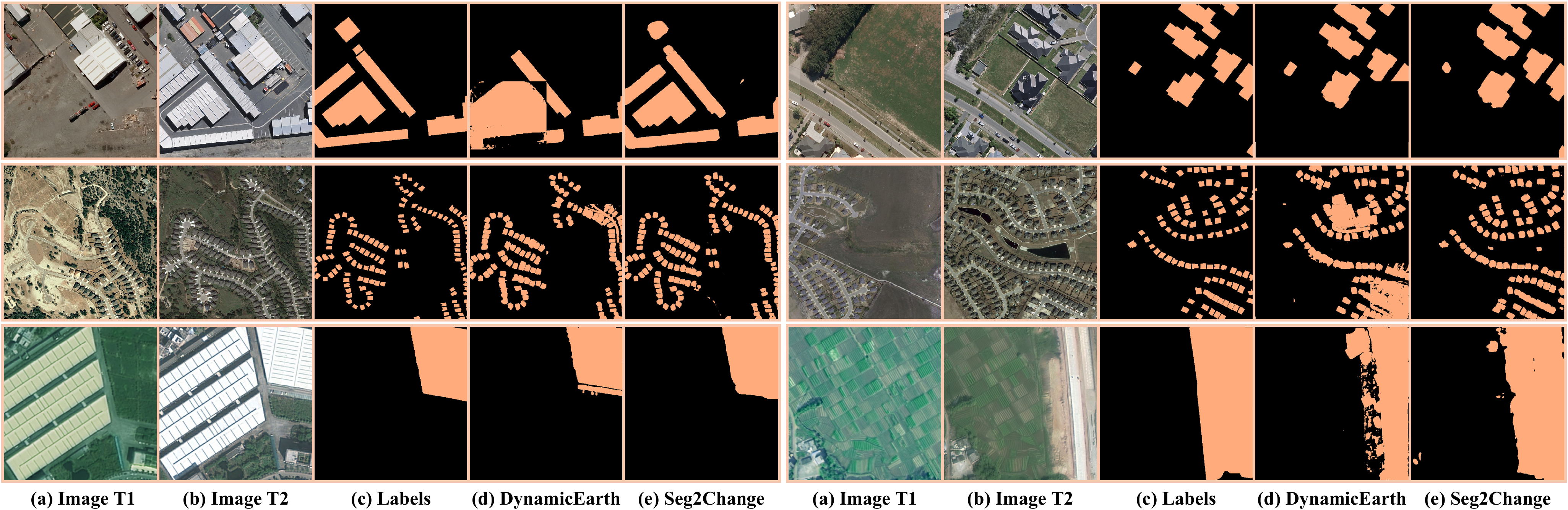}
  \caption{Open-vocabulary building and land-cover change detection examples. In each group: images at T1 ($I^1$), T2 ($I^2$), labels, the results of DynamicEarth and our Seg2Change. Color rendering: \textcolor[HTML]{FFAB7C}{\textit{"Building/Land-cover"}}.}
  \label{bcd_vis}
\end{figure*}

\begin{table*}[ht]
\setlength\tabcolsep{2.5pt}
\centering
\caption{Performance comparison on binary OVCD datasets. $\dagger$ and $\ddagger$ denote the \textbf{M}–\textbf{C}–\textbf{I} and \textbf{I}–\textbf{M}–\textbf{C} configurations of DynamicEarth. * denotes the variants implemented with SegEarth-OV3 (SegE-OV3 for short). The best and the second-best results are, respectively, marked in \textbf{BOLD} and in \underline{underline}. Except for Kappa, all results are expressed in percentage (\%).}
\scalebox{0.95}{
\begin{tabular}{c|c|c|ccc|ccc|ccc|ccc}
\hline
\multirow{2}{*}{Method} & \multirow{2}{*}{Identifier} & \multirow{2}{*}{Comparator} & \multicolumn{3}{c|}{WHU-CD} & \multicolumn{3}{c|}{LEVIR-CD} & \multicolumn{3}{c|}{DSIFN} & \multicolumn{3}{c}{CLCD} \\
\cline{4-15}
& & & F1$^c$ & IoU$^c$ & Kappa & F1$^c$ & IoU$^c$ & Kappa & F1$^c$ & IoU$^c$ & Kappa & F1$^c$ & IoU$^c$ & Kappa \\
\midrule
PCA\_KMeans \cite{pca_km} & / & KMeans \cite{pca_km} & 14.33 & 7.72 & 0.0846 & 9.96 & 5.24 & 0.0116 & 35.48 & 21.57 & 0.1865 & 18.42 & 10.14 & 0.0780 \\ 
CVA \cite{cva} & / & CVA Match \cite{cva}  & 7.17 & 3.72 & 0.0023 & 9.27 & 4.86 & 0.0017 & 27.88 & 16.20 & 0.0142 & 13.39 & 7.17 & 0.0017 \\
DCVA \cite{dcva} & / & CVA Match \cite{cva}  & 20.55 & 11.45 & 0.1537 & 13.85 & 7.44 & 0.0507 & 38.52 & 23.85 & 0.1758 & 20.81 & 11.62 & 0.0918 \\
UCD-SCM \cite{ucd_scm} & SAM & OTSU \cite{4310076} & 32.13 & 19.14 & 0.2792 & 32.36 & 19.30 & 0.2734 & 40.13 & 25.10 & 0.2443 & 23.31 & 13.19 & 0.1522 \\
AnyChange \cite{anychange} & SAM & Latent Match \cite{anychange} & 28.13 & 16.37 & 0.2330 & 32.68 & 19.53 & 0.2672 & 39.19 & 24.37 & 0.2280 & 31.96 & 19.02 & 0.2342 \\
AnyChange* \cite{anychange} & SegE-OV3 & Latent Match \cite{anychange} & 69.25 & 52.96 & 0.6790 & \underline{72.27} & \underline{56.58} & \underline{0.7043} & \underline{54.69} & \underline{37.64} & \underline{0.4798} & 27.55 & 15.98 & 0.2276 \\
Inst-CEG \cite{instceg} & APE & CEG \cite{instceg} & 62.54 & 45.49 & 0.6074 & 63.29 & 46.30 & 0.6084 & 31.81 & 18.91 & 0.2430 & 6.76 & 3.50 & 0.0020 \\
Inst-CEG* \cite{instceg} & SegE-OV3 & CEG \cite{instceg} & 71.35 & 55.46 & 0.7016 & 70.62 & 54.58 & 0.6912 & 47.21 & 30.90 & 0.3985 & 10.09 & 5.32 & 0.0648 \\
DynamicEarth$\dagger$ \cite{li2025dynamic_earth} & SAM2 & DINOv2 \cite{dinov2} & 57.35 & 40.20 & 0.5541 & 46.43 & 30.23 & 0.4242 & 54.35 & 37.32 & 0.4741 & 23.83 & 13.52 & 0.1916 \\
DynamicEarth$\ddagger$ \cite{li2025dynamic_earth} & APE & DINOv2 \cite{dinov2} & 75.85 & 61.09 & 0.7488 & 69.70 & 53.50 & 0.6789 & 26.42 & 15.22 & 0.2260 & 14.97 & 8.09 & 0.1382 \\
DynamicEarth* \cite{li2025dynamic_earth} & SegE-OV3 & DINOv2 \cite{dinov2} & \underline{79.66} & \underline{66.20} & \underline{0.7879} & 71.97 & 56.21 & 0.7031 & 39.30 & 24.45 & 0.3423 & \underline{38.16} & \underline{23.58} & \underline{0.3535} \\
\midrule
\rowcolor[HTML]{EBEBEB}\textbf{Seg2Change} & SegE-OV3 & CACH (ours) & \textbf{86.18} & \textbf{75.72} & \textbf{0.8562} & \textbf{78.72} & \textbf{64.91} & \textbf{0.7742} & \textbf{58.56} & \textbf{41.40} & \textbf{0.5075}  & \textbf{47.89} & \textbf{31.48} & \textbf{0.4239} \\
\hline
\end{tabular}
}
\label{tab:results_bcd}
\end{table*}

\subsection{Model Detail}
% \textbf{Residual Upsampler (ResUp).} 
\textbf{Loss Function.}
First, we aggregate the refined feature differences $\{X_i^d\}$ across all layers and upsample them using the proposed ResUp (details in the Appendix) to obtain a category-agnostic change map. We compute the binary cross-entropy loss between the predicted change map and the ground-truth change labels:
\begin{align}
\mathcal{L}_{\text{cd}} = \mathcal{L}_\text{bce}(\delta_\uparrow(X^d_2, X^d_5, X^d_8, X^d_{11}), y^l),
\end{align}
where $y^l$ denotes the change detection labels in CA-CDD, and $\delta_\uparrow$ represents our upsampler ResUp. Then, we upsample the refined feature difference outputs $X_i^d$ from each layer to obtain $N_L$ layer-wise predictions (at layers 2, 5, 8, and 11), and compute the feature upsampling loss between the upsampled multi-dimensional feature differences and the change detection labels:
\begin{align}
\mathcal{L}_{\text{ups}} = \textstyle\sum_{i\in N_L} \mathcal{L}_\text{bce}(\delta_\uparrow(X^d_i), y^l),
\end{align}
% \begin{align}
% \mathcal{L}_{\text{ups}} = \textstyle\sum_{i=2}^{N_L} \mathcal{L}_\text{bce}(\delta_\uparrow(X^d_i), y^l),
% \end{align}

To suppress pseudo-change responses, we introduce a similarity constraint for unchanged regions. Specifically, we first obtain semantic representations by aggregating and upsampling the modulated features $\tilde{F}^1$ and $\tilde{F}^2$. The cosine similarity between the semantic representation outputs of the bi-temporal images is then calculated, followed by the computation of its complementary probability. By inverting the change detection labels, we obtain the labels for the unchanged regions. The loss for these unchanged (similar) regions is subsequently computed to penalize inconsistent predictions:
\begin{align}
\mathcal{L}_{\text{sim}} = [1 - \text{cos}(\delta_\uparrow(\tilde{F}^1), \delta_\uparrow(\tilde{F}^2))] \cdot \tilde{y}^l,
\end{align}
where $\tilde{F}^1$ and $\tilde{F}^2$ denote the modulated multi-level pyramid features at time $t$ and time $t+1$, respectively. $\tilde{y}^l$ represents the unchanged-region labels obtained by inverting the change detection labels. Finally, the overall loss function is formulated as a weighted sum of the individual loss terms:
\begin{align}
\mathcal{L}_{\text{total}} = \alpha \mathcal{L}_{\text{cd}} + \beta \mathcal{L}_{\text{ups}} + \upsilon \mathcal{L}_{\text{sim}},
\end{align}
where $\alpha$, $\beta$, and $\upsilon$ denote the weights for the category-agnostic change map loss, the feature upsampling loss, and the unchanged-region loss, respectively.

\section{Experiments}

\subsection{Datasets \& Benchmarks}
To assess the performance of Seg2Change, we conduct comprehensive experimental evaluations on two representative datasets for each of three OVCD tasks: building change detection (WHU-CD \cite{whucd}, LEVIR-CD \cite{levircd}), land-cover change detection (DSIFN \cite{DSIFN}, CLCD \cite{clcd}), and semantic change detection (SC-SCD \cite{TAN2025374}, SECOND \cite{second}). We select classical algebra-based unsupervised change detection methods as well as VLM-based unsupervised change detection methods for comparison, including PCA\_KMeans \cite{pca_km}, CVA \cite{cva}, DCVA \cite{dcva}, UCD-SCM \cite{ucd_scm}, and AnyChange \cite{anychange}. We further compare our method with the current state-of-the-art OVCD approaches, including Inst-CEG \cite{instceg} and DynamicEarth \cite{li2025dynamic_earth}. We reproduce the results by either reimplementing these methods or utilizing their publicly available code. For a fair comparison, we replace the original identifiers (\textit{e.g.}, SegEarth-OV \cite{li2025segearthov} and APE \cite{APE}) in these methods with the current state-of-the-art OVSS model (\textit{i.e.}, SegEarth-OV3 \cite{li2025segearthov3}) for remote sensing community.

\subsection{Implementation Details}
\textbf{Codebase.}
We implement Seg2Change in PyTorch and decouple the pre-training of CACH from its inference pipeline to ensure code clarity and modularity. The visual encoder in CACH is built upon the ViT-Base variant of DINOv2 \cite{dinov2}, where features from layers $\{2, 5, 8, 11\}$ are selected as inputs, with $C_{dim} = 768$ and $p_s = 14$. For the BDFM and EDQA modules, the channel dimensions are set to $C_2=48$, $C_5=64$, $C_8=80$, and $C_{11}=96$. The sliding window size $S_w$ is set to 9. The number of experts $N_e$ is set to 4. Following \cite{ucd_scm, instceg}, we leverage multiple sub-classes to enhance prediction confidence, \textit{e.g.}, rename "building" to \{"building", "house", "roof"\}. Detailed class names for all datasets are listed in the Appendix.

\textbf{Setup.} For a fair comparison, we follow the OVCD experimental settings adopted in DynamicEarth \cite{li2025dynamic_earth}. All experiments are conducted on a single NVIDIA RTX 4090 GPU. We crop both the pre-training and testing datasets to 512 × 512. Accordingly, in the Seg2Change framework, the height ($H$) and width ($W$) are set to 512 for both pre-training and inference. CACH is trained with an initial learning rate of $1\times10^{-3}$, a batch size of 4, and 20 epochs. The loss weights $\alpha$, $\beta$, and $\upsilon$ are set to 0.8, 0.1, and 0.1, respectively.

% \textbf{Evaluation.} For all datasets, we evaluate performance using the F1-score, intersection over union (IoU), and the Kappa coefficient. Following the standard CD setting, the F1-score (F1$^c$) and IoU (IoU$^c$) are reported for the changed class only. In contrast, the Kappa coefficient measures overall agreement by incorporating both changed and unchanged pixels. For semantic CD tasks, we report mean metric values (\textit{e.g.}, $\text{mIoU}^c$) across all semantic categories. In addition, we include overall accuracy (OA) as a complementary metric to reflect multi-class classification performance.

% $\dagger$ and $\ddagger$ denote the \textbf{M}–\textbf{C}–\textbf{I} and \textbf{I}–\textbf{M}–\textbf{C} configurations of DynamicEarth, respectively. * denotes the variants implemented with SegEarth-OV3.

% The best and the second-best results are, respectively, marked in \textbf{BOLD} and in \underline{underline}. Except for mKappa, all results are expressed in percentage (\%).

\begin{figure*}
  \centering
  \includegraphics[width=\textwidth]{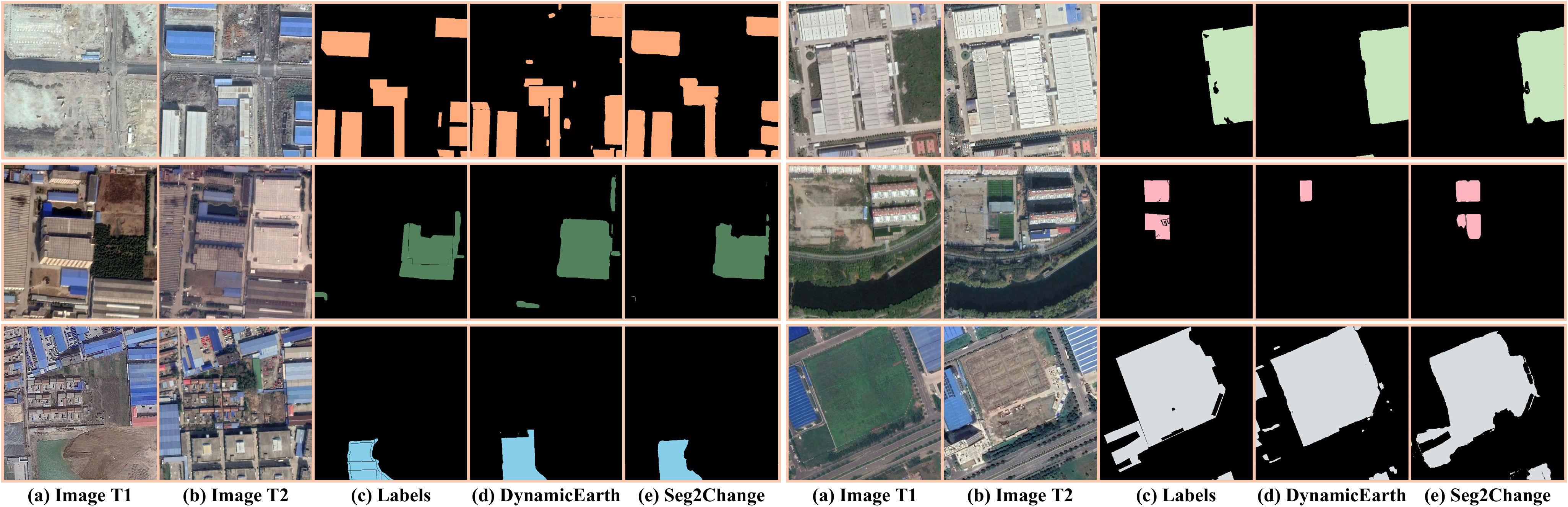}
  \caption{Open-vocabulary semantic change detection examples. In each group: images at T1 ($I^1$), T2 ($I^2$), labels, the results of DynamicEarth and our Seg2Change. Color rendering: \textcolor[HTML]{FFAB7C}{\textit{"Building"}}, \textcolor[HTML]{C7E5BD}{\textit{"Vegetation"}}, \textcolor[HTML]{54815E}{\textit{"Tree"}}, \textcolor[HTML]{FFB5C0}{\textit{"Playground"}}, \textcolor[HTML]{87CEEB}{\textit{"Water"}}, \textcolor[HTML]{D7DCE1}{\textit{"Bareland"}}.}
  \label{scd_vis}
\end{figure*}

\begin{table*}[htb]
\setlength\tabcolsep{4pt}
\centering
\caption{Performance, computational resource usage, and inference time cost comparison on semantic OVCD datasets. GPU memory usage and inference time are evaluated on a pair of bi-temporal remote sensing images ($[2, 3, 512, 512]$).}
\scalebox{0.98}{
\begin{tabular}{c|c|c|cccc|cccc}
\hline
\multirow{2}{*}{Method} & GPU Memory & Inference & \multicolumn{4}{c|}{SC-SCD} & \multicolumn{4}{c}{SECOND} \\
\cline{4-11}
& Usage (GB) $\downarrow$ & (ms/sample) $\downarrow$ & mF1$^c$ & mIoU$^c$ & mOA & mKappa & mF1$^c$ & mIoU$^c$ & mOA & mKappa \\
\midrule
UCD-SCM \cite{ucd_scm} & 9.46 (100\%) & 3225 (100\%) & 12.06 & 6.51 & 85.95 & 0.0619 & 14.63 & 8.40 & 78.61 & 0.0795 \\
AnyChange \cite{anychange} & \underline{6.59} (69\%) & 3988 (124\%) & 16.51 & 9.32 & 78.27 & 0.1056 & 19.54 & 11.84 & 76.07 & 0.1252 \\
AnyChange* \cite{anychange} & 10.15 (107\%) & \underline{2683} (83\%) & 16.64 & 9.48 & 88.06 & 0.1192 & 21.30 & 13.81 & 77.97 & 0.1492 \\
Inst-CEG \cite{instceg} & 14.95 (158\%) & 5672 (176\%) & 6.90 & 3.68 & 94.40 & 0.0551 & 17.82 & 10.46 & 92.80 & 0.1604 \\
Inst-CEG* \cite{instceg} & 7.08 (75\%) & 2928 (91\%) & 23.57 & 14.35 & 94.49 & 0.2139 & 29.35 & 18.40 & 93.62 & 0.2670 \\
DynamicEarth$\dagger$ \cite{li2025dynamic_earth} & 7.33 (77\%) & 5035 (156\%) & \underline{29.11} & \underline{17.97} & 91.37 & \underline{0.2532} & \underline{37.51} & \underline{23.58} & 91.88 & \underline{0.3297} \\
DynamicEarth$\ddagger$ \cite{li2025dynamic_earth} & 15.33 (162\%) & 6784 (210\%) & 9.19 & 5.16 & \underline{95.25} & 0.0877 & 22.17 & 13.72 & 93.76 & 0.2109 \\
DynamicEarth* \cite{li2025dynamic_earth} & 11.19 (118\%) & 2892 (89\%) & 19.87 & 11.47 & 95.03 & 0.1804 & 25.43 & 15.24 & \underline{93.83} & 0.2286 \\
\midrule
\rowcolor[HTML]{EBEBEB}\textbf{Seg2Change} & \textbf{6.08} (64\%) & \textbf{1521} (47\%) & \textbf{35.68} & \textbf{23.22} & \textbf{95.82} & \textbf{0.3385} & \textbf{42.89} & \textbf{29.08} & \textbf{95.17} & \textbf{0.4045} \\
\hline
\end{tabular}
}
\label{tab:results_scd}
\end{table*}

\subsection{Comparison to State-of-the-art}
To guarantee experimental fairness and emphasize the effectiveness of CACH in capturing category-agnostic changes, we replace the identifiers in three representative methods (AnyChange \cite{anychange}, Inst-CEG \cite{instceg}, and DynamicEarth \cite{li2025dynamic_earth}) with the recently released SegEarth-OV3 \cite{li2025segearthov3}. The modified versions are denoted with an asterisk (*), while their comparators are retained.

\textbf{Evaluations of Building Change Detection.}
As shown in the left panel of Table \ref{tab:results_bcd}, we evaluate Seg2Change on two widely adopted benchmark datasets for building CD: WHU-CD \cite{whucd} and LEVIR-CD \cite{levircd}. Conventional algebra-based unsupervised CD methods exhibit limited practical utility, while VLM-based methods drive unsupervised CD toward supporting specified-class open-vocabulary settings. Notably, replacing the SAM-based identifier with SegEarth-OV3 (SAM3-based) yields consistent performance gains across building CD benchmarks. Semantic segmentation is a critical component of CD, and its quality directly influences the accurate delineation of changed objects. Under the same identifier setting, our category-agnostic change map strategy improves the average IoU$^c$ across the two datasets by nearly 9\%. From the perspective of Kappa, which jointly reflects change and invariant pixel accuracy, Seg2Change more effectively suppresses pseudo-changes.

% Notably, by using category-agnostic change maps to guide open-vocabulary semantic segmentation models, we achieve a new performance breakthrough on WHU-CD. For the change class, our method achieves an $\text{F1}$ score of 85.24. Meanwhile, on the LEVIR-CD dataset, our method also demonstrates strong performance, surpassing the previous leading approach by 11.17\% in change-class $\text{IoU}$.

\textbf{Evaluations of Land-Cover Change Detection.}
In practical applications, user interests are not limited to building categories alone but extend to multi-category scenarios. Land-cover CD focuses on land-use transitions, \textit{e.g.}, "vegetation" converted to "tilled land", "bareland" transformed into "roads", "lakes", or "buildings". Instance-level approaches built upon SAM suffer from coarse boundary delineation, limiting their ability to detect intact object contours. Through a pixel-level change modeling strategy, we alleviate the cascading error propagation in VLM-based approaches. Further, the category-agnostic change representation generated by CACH naturally aligns with land-cover CD scenarios. With the integration of tailored language prompts, our method attains notable gains, improving $\text{F1}^c$ by 3.87\% on DSIFN and 9.73\% on CLCD.

% \begin{table*}[htb]
% \setlength\tabcolsep{2.5pt}
% \centering
% \caption{Performance comparison of different semantic change detection methods on HRSCD and SECOND datasets. $\dagger$ denotes the results with our reimplementation. The best results are \textbf{bolded} and the second-best results are \underline{underlined}. All results of the four evaluation metrics are described as percentages (\%).}
% \scalebox{0.9}{
% \begin{tabular}{c|c|c|ccc|ccc}
% \hline
% \multirow{2}{*}{Method} & \multirow{2}{*}{FLOPs(G)} & Inference & \multicolumn{3}{c|}{In-Domain Test} & \multicolumn{3}{c}{Out-Domain Test} \\
% \cline{4-9}
% & & (s/sample) & F1 & IoU & OA & F1 & IoU & OA \\
% \midrule
% UCD-SCM & 175.41 & 269.4 & 12.22 & 6.51 & 74.51 & 16.05 & 8.72 & 76.64 \\
% AnyChange & 178.04 & 259.3 & 17.07 & 9.33 & 77.03 & 19.43 & 10.76 & 80.16 \\
% Inst-CEG & 166.71 & 146.3 & 12.48 & 6.65 & 73.33 & 10.56 & 5.57 & 67.46 \\
% DynamicEarth $\dagger$ & 206.78 & 278.9 & 30.07 & 17.70 & 82.77 & 31.23 & 18.50 & 88.16 \\
% DynamicEarth $\ddagger$ & 221.46 & 274.1 & 5.46 & 2.81 & 54.03 & 12.57 & 6.70 & 59.46 \\
% \midrule
% \textbf{Seg2Change} & 128.74 & \textbf{124.1} & \textbf{45.59} & \textbf{29.53} & \textbf{91.31} & \textbf{44.81} & \textbf{28.87} & \textbf{92.03} \\
% \hline
% \end{tabular}
% }
% \label{tab:results_bandon}
% \vspace{-0.8em}
% \end{table*}

\textbf{Evaluations of Semantic Change Detection.} As shown in Table \ref{tab:results_scd}, our method achieves improvements of 5.25\% and 5.50\% in class-wise $\text{mIoU}^c$ across all categories on the SC-SCD and SECOND datasets, respectively. Meanwhile, benefiting from our pixel-level inference paradigm, GPU memory consumption and inference latency are reduced by 36\% and 53\%, respectively. Notably, replacing the identifier in DynamicEarth with SegEarth-OV3 does not lead to performance improvement. DynamicEarth extracts visual features of change proposals using DINOv2 and determines changes by computing cosine similarity. Due to its suboptimal fixed-threshold change judgment strategy, a stronger segmentation model introduces more false-positive change instances. By leveraging CACH to identify category-agnostic change maps, we effectively harness the advantages of stronger semantic segmentation models.

\begin{figure}[t]
  \centering
\includegraphics[width=0.95\linewidth]{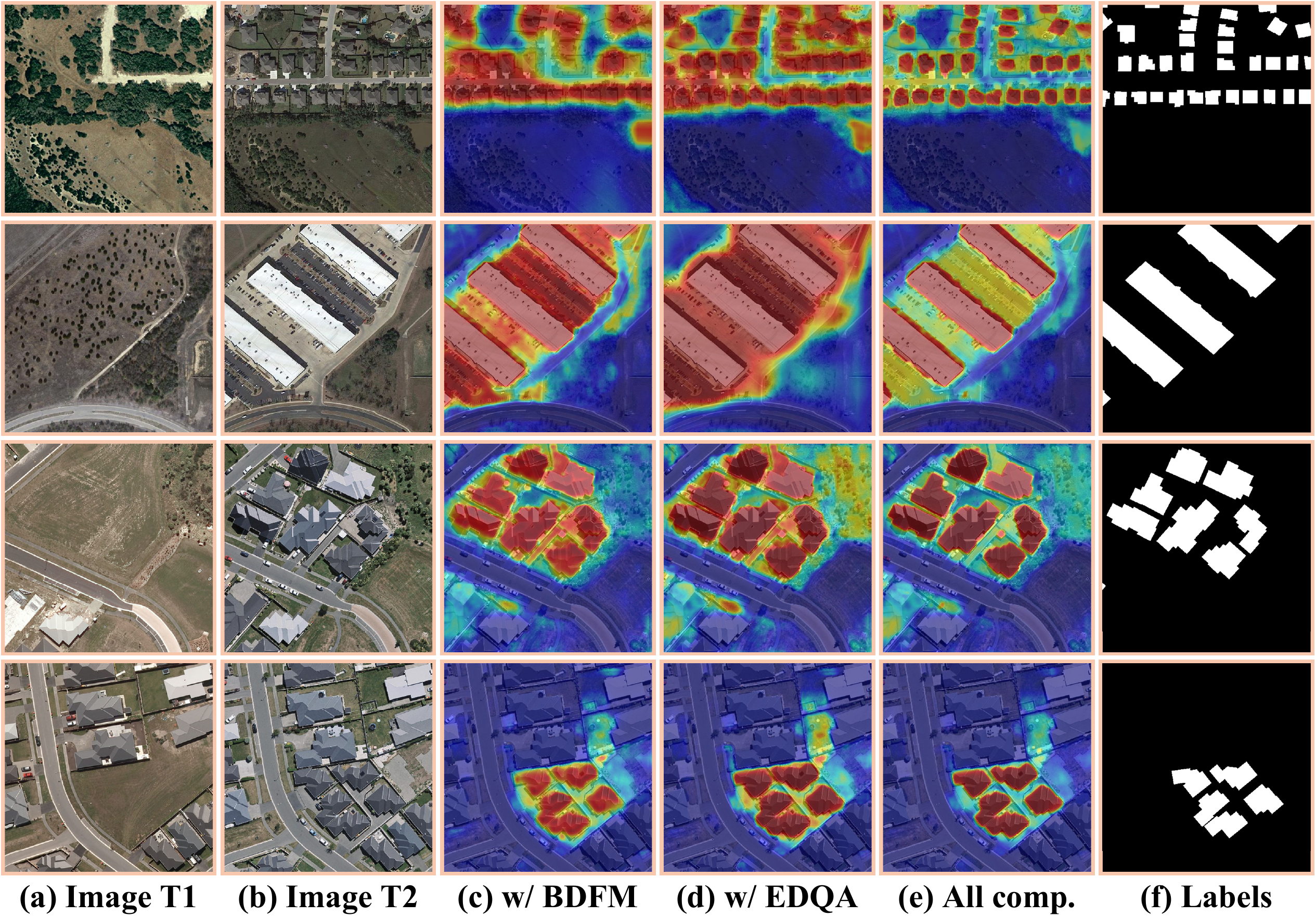}
  \caption{Visualization of feature outputs across our modules. Panel (e) presents the output features after all components, illustrating the progressive refinement of difference features.}
\label{heat_map}
\end{figure}

% Components in CACH

\begin{table}[t]
\setlength\tabcolsep{2pt}
\centering
\caption{Investigation on the impact of the proposed CACH components on the WHU-CD and DSIFN datasets.}
\scalebox{0.9}{
\begin{tabular}{cccc|ccc|ccc}
\hline
\multicolumn{4}{c|}{Components in CACH} &  \multicolumn{3}{c|}{WHU-CD} & \multicolumn{3}{c}{DSIFN} \\
\hline
\multicolumn{10}{c}{The Effectiveness of Difference Modules in CACH}  \\
\hline
FMM & BDFM & EDQA & ResUp & F1$^c$ & IoU$^c$ & Kappa & F1$^c$ & IoU$^c$ & Kappa \\
\hline
\ding{51} & \ding{55} & \ding{55} & \ding{55} & 69.68 & 53.47 & 0.6870 & 49.65 & 33.02 & 0.3581 \\
\ding{51} & \ding{51} & \ding{55} & \ding{55} & 78.10 & 64.06 & 0.7732 & 53.22 & 36.26 & 0.4758 \\
\ding{51} & \ding{51} & \ding{51} & \ding{55} & 84.18 & 72.68 & 0.8354 & 55.57 & 38.47 & 0.4966 \\
\rowcolor[HTML]{EBEBEB} \ding{51} & \ding{51} & \ding{51} & \ding{51} & \textbf{86.18} & \textbf{75.72} & \textbf{0.8562} & \textbf{58.56} & \textbf{41.40} & \textbf{0.5075} \\
\hline
\multicolumn{10}{c}{The Effectiveness of Backbone Features in CACH}  \\
\hline
\multicolumn{4}{l|}{Shallow: $\{F_0,F_2,F_3,F_5\}$} & 78.13 & 64.12 & 0.7736 & 49.78 & 33.14 & 0.4780 \\
\multicolumn{4}{l|}{Deep: $\{F_6,F_8,F_9,F_{11}\}$} & 81.41 & 68.64 & 0.8069 & 54.35 & 37.32 & 0.5318 \\
\multicolumn{4}{l|}{Comb1: $\{F_1,F_4,F_6,F_{10}\}$} & 82.92 & 70.82 & 0.8224 & 55.93 & 38.82 & 0.5390 \\
\rowcolor[HTML]{EBEBEB} \multicolumn{4}{l|}{Comb2: $\{F_2,F_5,F_8,F_{11}\}$} & \textbf{86.18} & \textbf{75.72} & \textbf{0.8562} & \textbf{58.56} & \textbf{41.40} & \textbf{0.5075} \\
\hline
\multicolumn{10}{c}{The Effectiveness of Loss Functions in CACH}  \\
\hline
\multicolumn{4}{l|}{Change Map Loss ($\mathcal{L}_{\text{cd}}$)} & 84.31 & 72.88 & 0.8371 & 52.69 & 35.77 & 0.7593 \\
\multicolumn{4}{l|}{+ Upsample Loss ($\mathcal{L}_{\text{ups}}$)} & 85.58 & 74.79 & 0.8500 & 55.32 & 38.24 & 0.4942 \\
\rowcolor[HTML]{EBEBEB} \multicolumn{4}{l|}{+ Unchanged Loss ($\mathcal{L}_{\text{sim}}$)} & \textbf{86.18} & \textbf{75.72} & \textbf{0.8562} & \textbf{58.56} & \textbf{41.40} & \textbf{0.5075}\\
\hline
\end{tabular}
}
\label{tab:ablation_CACH}
\end{table}

\textbf{Visualization Results.}
We qualitatively compare Seg2Change with the best-performing configuration of DynamicEarth. As shown in Fig. \ref{bcd_vis}–\ref{scd_vis}, Seg2Change consistently delivers accurate change detection across all categories, whereas DynamicEarth requires switching configurations to achieve optimal performance for building or land-cover change detection. In addition, we alleviate the difficulty in determining change boundaries caused by fixed thresholds.

% \textbf{Evaluations of OV Cross-Domain CD.}
% Cross-domain change detection represents an extreme scenario in change detection applications. In practice, annotating dedicated data and training scene-specific models for each scenario is costly. Therefore, models are expected to perform well on both in-domain and out-of-domain data, exhibiting strong generalization ability. To evaluate Seg2Change’s capability to handle out-of-domain data with viewpoint shifts, we conduct experiments on both in-domain and out-of-domain test sets of the BANDON dataset. As illustrated in the figure, the in-domain and out-of-domain data differ in terms of viewing angles and imaging quality. As reported in Table 2, DynamicEarth and previous OVCD methods achieve IoU scores below 20\% on both in-domain and out-of-domain data, whereas Seg2Change—without training on either domain—improves performance to 29.53\% and 44.81\%, respectively.
% To investigate the impact of each component, we conduct an ablation study on two binary change detection datasets in Table \ref{tab:ablation_CACH} and visualize the contribution of each component in CACH to the difference features in Fig. \ref{heat_map}. As depicted in Table \ref{tab:ablation_CACH}, we examine the effectiveness of various combinations of backbone features and further investigate the impact of the proposed loss functions on OVCD performance. By incorporating the proposed multi-scale feature upsampling and unchanged pixel supervision, OVCD performance is further improved.

\subsection{Ablation Study}
\textbf{Impact of Components in CACH.} 
Table \ref{tab:ablation_CACH} and Fig. \ref{heat_map} present a quantitative and qualitative analysis of the contributions of each component in CACH. The results show that progressively adding BDFM, EDQA, and ResUp consistently improves performance, demonstrating their effectiveness and complementarity. Among different backbone features, the "Comb2" configuration achieves the best results, highlighting the importance of proper multi-level feature selection. Furthermore, incorporating the upsample loss and unchanged-region loss yields additional performance gains, with the latter effectively mitigating pseudo-changes.

\textbf{Impact of Different CACH Pre-train Dataset.}
We investigate the impact of the CACH pre-train dataset on the semantic change detection performance in Table \ref{tab:ablation_CACD}. The original bi-temporal images and coarse annotations of CA-CDD are derived from the SECOND \cite{second}, JL1-CD \cite{liu2025jl1}, and CNAM-CD \cite{rs15092464} datasets. We further expand and re-annotate the original coarse labels. Benefiting from enriched category-agnostic change annotations, Seg2Change achieves superior performance across all OVCD tasks.

% Any semantic category transition is recorded as a change in the corresponding change map.

\begin{table}[t]
\setlength\tabcolsep{1pt}
\centering
\caption{Ablation experimental results ($\text{mF1}^c$) of the different CACH pre-train datasets on the SECOND dataset.}
\scalebox{0.89}{
\begin{tabular}{l|cccccc|c}
\hline
\multirow{2}{*}{Pre-train} & \multicolumn{7}{c}{SECOND (Semantic Change Detection Categories)} \\
\cline{2-8}
 & \small{Building} & \small{Tree} & \small{Water} & \small{Vegetation} & \small{Bareland} & \small{Playground} & \small{$\text{mF1}^c$}  \\
\hline
SECOND \cite{second} & 75.50 & 21.58 & 28.32 & 25.36 & 49.88 & 27.15 & 37.96 \\
JL1-CD \cite{liu2025jl1} & 70.46 & 16.96 & 24.90 & 23.54 & 48.02 & 28.78 & 35.44 \\
CNAM-CD \cite{rs15092464} & 66.46 & 22.94 & 28.09 & 26.85 & 46.33 & 24.49 & 35.86 \\
\rowcolor[HTML]{EBEBEB} CA-CDD (ours) & \textbf{76.36} & \textbf{27.73} & \textbf{37.23} & \textbf{31.49} & \textbf{54.29} & \textbf{30.23} & \textbf{42.89} \\
\hline
\end{tabular}
}
\label{tab:ablation_CACD}
\end{table}

\begin{table}[t]
\setlength\tabcolsep{1pt}
\centering
\caption{Investigation on the impact of different implementation of OVSS model and the effectiveness of our CACH.}
\scalebox{0.89}{
\begin{tabular}{l|cccc|cccc}
\hline
\multirow{2}{*}{OVSS Model} &  \multicolumn{4}{c|}{SC-SCD} & \multicolumn{4}{c}{SECOND} \\
\cline{2-9}
& mF1$^c$ & mIoU$^c$ & mOA & mKappa & mF1$^c$ & mIoU$^c$ & mOA & mKappa \\
\hline
SegE-OV \cite{li2025segearthov}  & 20.74 & 12.28 & 95.03 & 0.1866 & 34.24 & 22.25 & 94.49 & 0.3175 \\
RSKT-Seg \cite{li2025exploring}  & 24.90 & 14.22 & 91.34 & 0.2192 & 34.98 & 23.27 & 94.44 & 0.3219 \\
SAM3 \cite{carion2025sam3segmentconcepts}  & 26.07 & 15.49 & 94.92 & 0.2379 & 35.73 & 23.81 & 94.37 & 0.3303 \\
\rowcolor[HTML]{EBEBEB} SegE-OV3 \cite{li2025segearthov3} & \textbf{35.68} & \textbf{23.22} & 95.82 & \textbf{0.3385} & \textbf{42.89} & \textbf{29.08} & 95.17 & \textbf{0.4045} \\
\hline
Method & \multicolumn{8}{c}{Replace Comparator with our Proposed CACH}  \\
\hline
UCD-SCM & 12.06 & 6.51 & 85.95 & 0.0619 & 14.63 & 8.40 & 78.61 & 0.0795 \\
\rowcolor[HTML]{EBEBEB} + CACH (ours) & 19.45 & 10.78 & \underline{97.67} & 0.1877 & 20.73 & 11.56 & \textbf{97.58} & 0.1990 \\
Inst-CEG & 6.90 & 3.68 & 94.40 & 0.0551 & 17.82 & 10.46 & 92.80 & 0.1604 \\
\rowcolor[HTML]{EBEBEB} + CACH (ours) & 11.77 & 6.25 & 94.77 & 0.0943 & 22.63 & 12.76 & \underline{97.53} & 0.2169 \\
DynamicEarth$\dagger$ & 29.11 & 17.97 & 91.37 & 0.2532 & 37.51 & 23.58 & 91.88 & \underline{0.3297} \\
\rowcolor[HTML]{EBEBEB} + CACH (ours) & \underline{33.46} & \underline{20.09} & \textbf{99.52} & \underline{0.3326} & \underline{40.54} & \underline{25.42} & 83.38 & 0.3094 \\
DynamicEarth$\ddagger$ & 9.19 & 5.16 & 95.25 & 0.0877 & 22.17 & 13.72 & 93.76 & 0.2109 \\
\rowcolor[HTML]{EBEBEB} + CACH (ours) & 13.72 & 7.37 & 91.09 & 0.1184 & 24.33 & 13.85 & 99.17 & 0.2393 \\
\hline
\end{tabular}
}
\label{tab:ablation_OVSS}
\end{table}

\textbf{Impact on Different Implementations of OVSS Model and the Effectiveness of the Proposed CACH.}
Table \ref{tab:ablation_OVSS} investigates the impact of different implementations of the OVSS model in Seg2Change and replaces the comparators of other OVCD methods with our CACH. Our CACH fully unleashes the potential of the OVSS model for further performance enhancement. Existing approaches range from OTSU \cite{4310076} in UCD-SCM to change event generation in Inst-CEG \cite{instceg}, and further to DynamicEarth \cite{li2025dynamic_earth}, which utilizes the DINOv2 image encoder to extract feature discrepancies. Despite their differences, these approaches rely on a predefined threshold, whose optimal value varies substantially among instances and is therefore challenging to generalize.

% \begin{table}[t]
% \setlength\tabcolsep{1.5pt}
% \centering
% \caption{Investigation on the impact of our feature upsample and unchanged losses on the WHU-CD and DSIFN datasets.}
% \scalebox{0.9}{
% \begin{tabular}{c|c|ccc|ccc}
% \hline
% \multirow{2}{*}{No.} & \multirow{2}{*}{Loss Function} &  \multicolumn{3}{c|}{WHU-CD} & \multicolumn{3}{c}{DSIFN} \\
% \cline{3-8}
% & & F1$^c$ & IoU$^c$ & Kappa & F1$^c$ & IoU$^c$ & Kappa \\
% \hline
% \#01 & w/ change detection loss & 84.31 & 72.88 & 0.8371 & 52.69 & 35.77 & 0.7593 \\
% \#02 & \#01 w/ feat. upsample loss & 85.58 & 74.79 & 0.8500 & 55.32 & 38.24 & 0.4942 \\
% \rowcolor[HTML]{EBEBEB} \#03 & \#02 w/ unchanged loss & \textbf{86.18} & \textbf{75.72} & \textbf{0.8562} & \textbf{58.61} & \textbf{41.45} & \textbf{0.5075} \\
% \hline
% \end{tabular}
% }
% \label{tab:ablation_LOSS}
% \end{table}

\section{Conclusion}
In this paper, we present Seg2Change, a novel approach for adapting open-vocabulary semantic segmentation models for remote sensing bi-temporal change detection. We first analyze the limitations of existing OVCD methods, which mainly rely on mask generators and fixed thresholds. By constructing a category-agnostic change detection dataset and designing a change estimation head that enhances and calibrates feature discrepancies, our method can generate change maps for arbitrary category transitions. Furthermore, we build a bridge to directly leverage powerful OVSS models for open-vocabulary change detection. Finally, we develop a simple yet efficient adaptation architecture (\textit{i.e.}, Seg2Change) that achieves state-of-the-art performance across various OVCD tasks.

\appendix

\makeatletter
\twocolumn[{
  \begin{center}
    \Huge \bfseries \@titlefont \textbf{Seg2Change: Adapting Open-Vocabulary Semantic Segmentation Model for Remote Sensing Change Detection}\\
  \end{center}
  \begin{center}
    \huge Supplementary Material\\[1em]
  \end{center}
}]
\makeatother

\textbf{\Large Supplementary Material Contents}
\begin{itemize}
\item \textbf{Section \textcolor[HTML]{3E94EF}{\ref{More_Experimental_Details}} - More Experimental Details}.

\item \textbf{Section \textcolor[HTML]{3E94EF}{\ref{More_Diagnostic_Experiments}} - More Diagnostic Experiments}.

\item \textbf{Section \textcolor[HTML]{3E94EF}{\ref{More_Detailed_Experiment_Results}} - More Detailed Experiment Results}.

\item \textbf{Section \textcolor[HTML]{3E94EF}{\ref{Detailed_Architecture}} - Detailed Architecture of Seg2Change}.

\item \textbf{Section \textcolor[HTML]{3E94EF}{\ref{CA-CDD}} - A Category-Agnostic CD Dataset}.

\item \textbf{Section \textcolor[HTML]{3E94EF}{\ref{More_Qualitative_Results}} - More Qualitative Results}.
\end{itemize}

\section{More Experimental Details}
\label{More_Experimental_Details}
\subsection{Dataset Description}
Below is a detailed description of all the datasets used to validate our Seg2Change. Unless otherwise specified, we use the test sets of the following datasets for validation.

\textbf{Building Change Detection:} (1) The WHU-CD \cite{whucd} is a large-scale, very high-resolution remote sensing building change detection dataset provided by Wuhan University. It comprises two aerial datasets from Christchurch, New Zealand, captured in April 2012 and 2016, with a spatial resolution of 0.3 m/pixel. The WHU-CD dataset consists of more than 220, 000 independent buildings. We applied a sliding window with a uniform stride to the testing image pairs with a resolution of 11265 $\times$ 15354, respectively, generating 660 testing samples of size 512 $\times$ 512. (2) The LEVIR-CD \cite{levircd} is a large-scale very high-resolution remote sensing building change detection dataset provided by Beihang University. These bi-temporal images with time span of 5 to 14 years have significant construction changes, with a spatial resolution of 0.5 m/pixel. This dataset covers various types of buildings and contains a total of 31,333 individual building change instances, each with dimensions of 1024 × 1024. We applied a sliding-window with uniform stride to the testing image pairs, generating 512 testing samples of size 512 × 512.

\textbf{Land-cover Change Detection:} (1) DSIFN \cite{DSIFN} is a dataset for land-cover change detection. It comprises 3,988 pairs of remote sensing images from six major cities in China. The dataset contains the change of multiple kinds of land cover objects, such as roads, buildings, croplands, and water bodies. Each image has a resolution of 512 $\times$ 512. Following the default splits, we divided the dataset into training, validation, and testing sets, consisting of 3,000, 340, and 48 image pairs, respectively. (2) The CLCD \cite{clcd} dataset consists of 600 pairs of cropland change images, including 360 pairs for training, 120 for validation, and 120 for testing. The bi-temporal images in CLCD were collected by Gaofen-2 in Guangdong Province, China, in 2017 and 2019, respectively, with spatial resolutions ranging from 0.5 to 2 m. Each sample pair consists of two 512 × 512 images and an associated binary label indicating cropland change.

\begin{table}[ht]
\centering
\setlength\tabcolsep{1pt}
\caption{The prompt class names for the evaluation datasets. \{\} indicates multiple prompt vocabularies for a single class. Foreground classes are highlighted in \textbf{BOLD}.}
\scalebox{0.8}{
\begin{tabular}{c|c|c}
\hline
Dataset & Category & Class Name  \\
\midrule
\multirow{3}{*}{\makecell{WHU-CD \cite{whucd} \\ (Single-Category)}}  & \multirow{3}{*}{Building} & \{bareland,barren\},grass,car, \\ 
&  & \{tree,forest\},\{water,river\}, \\
&  & cropland,\textbf{\{building,roof,house\}}\\
\hline
\multirow{3}{*}{\makecell{LEVIR-CD \cite{levircd} \\ (Single-Category)}}  & \multirow{3}{*}{Building} & \{bareland,barren\},grass,car, \\ 
&  & \{tree,forest\},\{water,river\}, \\
&  & cropland,\textbf{\{building,roof,house\}}\\
\hline

\multirow{3}{*}{\makecell{DSIFN \cite{DSIFN} \\ (Multi-Category)}}  & \multirow{3}{*}{Land-Cover} & \textbf{\{building,roof,house\},garden,playground,} \\ 
&  & \textbf{\{construction,apartment,residential\},} \\
&  & \textbf{materials},\{tree,forest\}, \{water,river\}\\
\hline
\multirow{3}{*}{\makecell{CLCD \cite{clcd} \\ (Multi-Category)}}  & \multirow{3}{*}{Land-Cover} & \textbf{\{bareland,barren, grass\}}, car, \\ 
&  &  \textbf{\{tree,forest\},\{water,river\}} \\
&  & \textbf{cropland,\{building,roof,house\}}\\
\hline
\multirow{6}{*}{\makecell{SC-SCD \cite{SC-SCD} \\ (Multi-Category)}}  & Bareland & \textbf{bareland,barren,ground,floor,soil} \\
& Water & \textbf{water,river,pond} \\
& Building & \textbf{building,roof,house} \\
& Structure & \textbf{structure,construction,greenhouse}  \\
& Farmland & \textbf{farmland,cropland,terrace} \\
& Vegetation & \textbf{grass,vegetation,tree,forest} \\
& Road & \textbf{road} \\
\hline
\multirow{6}{*}{\makecell{SECOND \cite{second} \\ (Multi-Category)}}  & Building & \textbf{building,roof,house} \\
& Tree & \textbf{tree,forest} \\
& Water & \textbf{water,river} \\
& Low vegetation & \textbf{grass,cropland}  \\
& N.v.g surface & \textbf{bareland,barren,ground} \\
& Playground & \textbf{sports field} \\
\hline
\end{tabular}
}
\label{prompts}
\end{table}

\begin{table}[t]
\setlength\tabcolsep{2.5pt}
\centering
\caption{Investigation on the impact of the CACH backbone on the WHU-CD and DSIFN datasets.}
\scalebox{0.92}{
\begin{tabular}{l|ccc|ccc}
\hline
\multirow{2}{*}{CACH Backbone} & \multicolumn{3}{c|}{WHU-CD} & \multicolumn{3}{c}{DSIFN} \\
\cline{2-7}
& F1$^c$ & IoU$^c$ & Kappa & F1$^c$ & IoU$^c$ & Kappa \\
\hline
ResNet-18 \cite{he2015deepresiduallearningimage} & 81.28 & 68.47 & 0.8062 & 46.38 & 30.19 & 0.4519 \\
ResNet-50 \cite{he2015deepresiduallearningimage} & 84.16 & 72.66 & 0.8355 & 55.84 & 38.74 & 0.5238 \\
DINOv2-Small \cite{dinov2} & 84.44 & 73.07 & 0.8384 & 54.99 & 37.93 & 0.5148 \\
\rowcolor[HTML]{EBEBEB} DINOv2-Base \cite{dinov2} & 86.18 & 75.72 & 0.8562 & 58.56 & 41.40 & 0.5075 \\
DINOv2-Large \cite{dinov2} & \textbf{86.58} & \textbf{76.33} & \textbf{0.8603} & \textbf{58.87} & \textbf{41.71} & \textbf{0.5693} \\
\hline
\end{tabular}
}
\label{tab:ablation_backbone}
\end{table}

% \begin{table*}[ht]
% \setlength\tabcolsep{1.5pt}
% \centering
% \caption{Investigation on the impact of different implementation of OVSS model on SC-SCD and SECOND datasets.}
% \scalebox{0.9}{
% \begin{tabular}{l|ccccccc|ccccccc}
% \hline
% \multirow{2}{*}{OVSS Model} & \multicolumn{7}{c|}{SC-SCD (Semantic Change Detection Categories)} & \multicolumn{7}{c}{SECOND (Semantic Change Detection Categories)} \\
% \cline{2-15}
%  & \small{Building} & \small{Tree} & \small{Water} & \small{Vegetation} & \small{Bareland} & \small{Playground} & \small{Avg.} & \small{Building} & \small{Tree} & \small{Water} & \small{Vegetation} & \small{Bareland} & \small{Playground} & \small{Avg.} \\
% \hline
% SegE-OV \cite{li2025segearthov} & 75.50 & 21.58 & 28.32 & 25.36 & 49.88 & 27.15 & 37.96 & 75.50 & 21.58 & 28.32 & 25.36 & 49.88 & 27.15 & 37.96 \\
% RSKT-Seg \cite{li2025exploring} & 70.46 & 16.96 & 24.90 & 23.54 & 48.02 & 28.78 & 35.44  & 75.50 & 21.58 & 28.32 & 25.36 & 49.88 & 27.15 & 37.96 \\
% SAM3 \cite{carion2025sam3segmentconcepts} & 66.46 & 22.94 & 28.09 & 26.85 & 46.33 & 24.49 & 35.86 & 75.50 & 21.58 & 28.32 & 25.36 & 49.88 & 27.15 & 37.96 \\
% \rowcolor[HTML]{EBEBEB} SegE-OV3 \cite{li2025segearthov3} & \textbf{76.36} & \textbf{27.73} & \textbf{37.23} & \textbf{31.49} & \textbf{54.29} & \textbf{30.23} & \textbf{42.89} & 75.50 & 21.58 & 28.32 & 25.36 & 49.88 & 27.15 & 37.96 \\
% \hline
% \end{tabular}
% }
% \label{tab:ablation_ovss_scd}
% \end{table*}

\textbf{Semantic Change Detection:} (1) The SC-SCD \cite{SC-SCD} consists of a pair of bi-temporal satellite images acquired from the Pléiades platform in 2014 and the Beijing-2 platform in 2016, with a spatial resolution of 0.5 m and a full size of 19,613 × 19,990 pixels (about 98.02 km$^2$), depicting urban and rural transitions in southern China region. The SC-SCD dataset includes seven land-cover categories: bareland, water, building, structure, farmland, vegetation, and road. Full coverage of landcover pixels are first annotated from bi-temporal RSIs, followed by difference analysis between the pair of land-cover maps to obtain pixels with different categories. The bi-temporal RSIs, land-cover maps and the corresponding BCD map are cropped into squares of 512 × 512 pixels. The training dataset consists of 1400 samples and the test dataset consists of 322 samples. (2) SECOND \cite{second} is a semantic change detection dataset comprising 4,662 pairs of aerial images (2,968 for training and 1,694 for testing) collected from multiple sensors. Each image has a size of 512 × 512 pixels with a spatial resolution of 0.5–3 m/pixel and includes pixel-level annotations. The SECOND dataset focuses on six major land cover categories—building, tree, water, low vegetation, non-vegetated ground (N.v.g) surface, and playground—which frequently undergo natural or human-induced anthropogenic changes.

\begin{table}[t]
\setlength\tabcolsep{1.5pt}
\centering
\caption{Investigation on the impact of different implementation of OVSS model on WHU-CD and LEVIR-CD datasets.}
\scalebox{0.92}{
\begin{tabular}{l|ccc|ccc}
\hline
\multirow{2}{*}{OVSS Model} &  \multicolumn{3}{c|}{WHU-CD} & \multicolumn{3}{c}{LEVIR-CD} \\
\cline{2-7}
& F1$^c$ & IoU$^c$ & Kappa & F1$^c$ & IoU$^c$ & Kappa \\
\hline
SegEarth-OV \cite{li2025segearthov}  & 74.53 & 59.40 & 0.7367 & 65.97 & 49.22 & 0.6438 \\
RSKT-Seg \cite{li2025exploring}  & 79.13 & 65.47 & 0.7840 & 72.40 & 56.75 & 0.7103 \\
SAM3 \cite{carion2025sam3segmentconcepts}  & 83.17 & 71.19 & 0.8255 & 77.82 & 63.69 & 0.7658 \\
\rowcolor[HTML]{EBEBEB} SegEarth-OV3 \cite{li2025segearthov3} & \textbf{86.18} & \textbf{75.72} & \textbf{0.8562} & \textbf{78.72} & \textbf{64.91} & \textbf{0.7742} \\
\hline
\end{tabular}
}
\label{tab:ablation_OVSS_bcd}
\end{table}

\begin{table}[t]
\setlength\tabcolsep{1.5pt}
\centering
\caption{Investigation on the impact of different implementation of OVSS model on DSIFN and CLCD datasets.}
\scalebox{0.92}{
\begin{tabular}{l|ccc|ccc}
\hline
\multirow{2}{*}{OVSS Model} &  \multicolumn{3}{c|}{DSIFN} & \multicolumn{3}{c}{CLCD} \\
\cline{2-7}
& F1$^c$ & IoU$^c$ & Kappa & F1$^c$ & IoU$^c$ & Kappa \\
\hline
SegEarth-OV \cite{li2025segearthov}  & 42.47 & 26.96 & 0.4051 & 33.81 & 20.35 & 0.3209 \\
RSKT-Seg \cite{li2025exploring}  & 50.97 & 34.20 & 0.4900 & 43.31 & 27.64 & 0.4210 \\
SAM3 \cite{carion2025sam3segmentconcepts}  & 52.00 & 35.13 & 0.5004 & 43.42 & 27.73 & 0.4228 \\
\rowcolor[HTML]{EBEBEB} SegEarth-OV3 \cite{li2025segearthov3} & \textbf{58.56} & \textbf{41.40} & \textbf{0.5075} & \textbf{47.89} & \textbf{31.48} & \textbf{0.4239} \\
\hline
\end{tabular}
}
\label{tab:ablation_OVSS_lcd}
\end{table}

% \begin{table}[t]
% \setlength\tabcolsep{1pt}
% \centering
% \caption{Ablation experimental results of the Different Composition of CACD Dataset.}
% \scalebox{0.9}{
% \begin{tabular}{c|ccccccc}
% \hline
% \multirow{2}{*}{Pre-train} & \multicolumn{7}{c}{SECOND (Semantic Change Detection Categories)} \\
% \cline{2-8}
%  & \small{Building} & \small{Tree} & \small{Water} & \small{Vegetation} & \small{Bareland} & \small{Playground} & \small{Avg.}  \\
% \hline
% SECOND \cite{second} & 75.50 & 21.58 & 28.32 & 25.36 & 49.88 & 27.15 & 37.96 \\
% JL1-CD \cite{liu2025jl1} & 70.46 & 16.96 & 24.90 & 23.54 & 48.02 & 28.78 & 35.44 \\
% CNAM-CD \cite{rs15092464} & 66.46 & 22.94 & 28.09 & 26.85 & 46.33 & 24.49 & 35.86 \\
% \rowcolor[HTML]{EBEBEB} CACD (ours) & \textbf{76.36} & \textbf{27.73} & \textbf{37.23} & \textbf{31.49} & \textbf{54.29} & \textbf{30.23} & \textbf{42.89} \\
% \hline
% \end{tabular}
% }
% \label{tab:ablation_CACD}
% \end{table}

\subsection{Evaluation metrics.} To compare the proposed method's performance with SOTA methods, we selected the most widely used evaluation metrics for change detection in remote sensing images: intersection over the union (IoU$^c$), F1-score (F1$^c$), Precision (Pre$^c$), Recall (Rec$^c$), Overall Accuracy (OA), and the Kappa coefficient (Kappa). These metrics are formulated as follows:
\begin{align}
\text{F1}^c &= (2 \times \text{Pre}^c \times \text{Rec}^c)/(\text{Pre}^c + \text{Rec}^c)      \\
\text{IoU}^c &= \text{TP}/(\text{TP} + \text{FN} + \text{FP}) \\
\text{Pre}^c &= \text{TP}/(\text{TP} + \text{FP})            \\
\text{Rec}^c &= \text{TP}/(\text{TP} + \text{FN})            \\
\text{OA} &= (\text{TP} + \text{TN})/(\text{TP} + \text{TN} + \text{FP} + \text{FN})    \\
\text{Kappa} &= (\text{OA} - \text{PRE}^c)/(1 - \text{PRE}^c)    \\
\begin{split}
\text{PRE}^c &= (\text{TP} + \text{FN})\times(\text{TP} + \text{FP})/(\text{TP} + \text{TN} + \text{FP} + \text{FN})^2 \\
     &+ (\text{TN} + \text{FP})\times(\text{TN} + \text{FN})/(\text{TP} + \text{TN} + \text{FP} + \text{FN})^2
\end{split}
\end{align} 

\subsection{Textual Description}
In Table \ref{prompts}, we present category definitions and corresponding class name mappings across multiple change detection datasets. We explicitly organize each semantic category into a set of fine-grained subclasses to capture richer semantic variations. Following \cite{ucd_scm, instceg}, we represent the building category as a collection of related subclasses, \textit{i.e.}, building: \{building, roof, house\}, and similarly group other categories, such as tree: \{tree, forest\} and water: \{water, river\}. This hierarchical formulation enables the unification of heterogeneous annotations across datasets while preserving semantic diversity. By leveraging such subclass-level representations, we enhance the flexibility of open-vocabulary change detection and improve the generalization capability of the model. Notably, we do not perform extensive prompt engineering to optimize performance.

\begin{table}[t]
\setlength\tabcolsep{1pt}
\centering
\caption{Investigation on the impact of different input channel at BDFM and EDQA on SC-SCD and SECOND datasets.}
\scalebox{0.87}{
\begin{tabular}{l|cccc|cccc}
\hline
Feature Channel &  \multicolumn{4}{c|}{SC-SCD} & \multicolumn{4}{c}{SECOND} \\
\cline{2-9}
($C_2, C_5, C_8, C_{11}$) & mF1$^c$ & mIoU$^c$ & mOA & mKappa & mF1$^c$ & mIoU$^c$ & mOA & mKappa \\
\hline
(16, 32, 48, 64) & 27.18 & 15.72 & 80.85 & 0.2224 & 32.29 & 19.25 & 90.47 & 0.2764 \\
(32, 48, 54, 80) & 27.64 & 16.04 & 80.83 & 0.2273 & 42.49 & 26.98 & 87.59 & 0.3670 \\

\rowcolor[HTML]{EBEBEB} (48, 64, 80, 96) & \textbf{35.68} & \textbf{23.22} & 95.82 & \textbf{0.3385} & 42.89 & 29.08 & \textbf{95.17} & 0.4045 \\
(32, 64, 128, 256) 
& 30.89 & 18.27 & \textbf{97.44} & 0.2970 & \textbf{47.30} & \textbf{30.98} & 87.65 & \textbf{0.4092} \\
\hline
\end{tabular}
}
\label{tab:ablation_channel}
\end{table}

\begin{table}[t]
\setlength\tabcolsep{1pt}
\centering
\caption{Investigation on the impact of different implementations of the feature difference module.}
\scalebox{0.9}{
\begin{tabular}{l|cccc|cccc}
\hline
Feature & \multicolumn{4}{c|}{SC-SCD} & \multicolumn{4}{c}{SECOND} \\
\cline{2-9}
Difference & mF1$^c$ & mIoU$^c$ & mOA & mKappa & mF1$^c$ & mIoU$^c$ & mOA & mKappa \\
\hline
feature abs & 28.93 & 16.91 & 99.52 & 0.2875 & 33.21 & 19.91 & \textbf{99.52} & 0.3301 \\
TFIM \cite{TFIM} & 28.10 & 16.35 & 99.53 & 0.2793 & 33.17 & 19.88 & \textbf{99.52} & 0.3296 \\
CDEM \cite{CDEM}  & 29.68 & 17.42 & 99.52 & 0.2949 & 35.44 & 21.54 & 54.41 & 0.1545 \\
MDFM \cite{MDFM} & 33.58 & 20.18 & 99.52 & 0.3338 & 44.10 & 28.29 & 87.70 & 0.3821 \\
\rowcolor[HTML]{EBEBEB} BDFM (ours) & \textbf{35.68} & \textbf{23.22} & \textbf{95.82} & \textbf{0.3385} & \textbf{42.89} & \textbf{29.08} & 95.17 & \textbf{0.4045} \\

\hline
\end{tabular}
}
\label{tab:ablation_bdfm}
\end{table}

\begin{table}[tb]
\setlength\tabcolsep{1.5pt}
\centering
\caption{Computational resource usage, and inference time cost comparison on semantic OVCD datasets. GPU memory usage and inference time are evaluated on a pair of bi-temporal remote sensing images ($[2, 3, 512, 512]$).}
\scalebox{0.8}{
\begin{tabular}{c|c|c|c|c}
\hline
Method & \makecell{SC-SCD \\ ($\text{mF1}^c$)$\uparrow$} & \makecell{GPU Memory \\ Usage (GB) $\downarrow$} & \makecell{Inference Time\\ (ms/sample) $\downarrow$}  &  \makecell{ Learnable \\ Parameters (M)} \\
\hline
DynamicEarth$\dagger$ \cite{li2025dynamic_earth} & 29.11 & 7.33 & 5035 & / \\
DynamicEarth$\ddagger$ \cite{li2025dynamic_earth} & 9.19  & 15.33 & 6784 & / \\
\hline
\rowcolor[HTML]{EBEBEB} Seg2Change & \textbf{35.68} & \textbf{6.08} & \textbf{1521} & \textbf{3.9} \\
\hline
\end{tabular}
}
\label{tab:parameters}
\end{table}

\section{More Diagnostic Experiments}
\label{More_Diagnostic_Experiments}
The main contributions of this study include a Category-Agnostic Change Detection Dataset (\textbf{CA-CDD}) and a Category-Agnostic Change Head (\textbf{CACH}), which comprises the Feature Modulation Module (\textbf{FMM}), the Bi-temporal Difference Fusion Module (\textbf{BDFM}), and the Effective Difference Query Attention (\textbf{EDQA}). We therefore adapted the open-vocabulary semantic \textbf{Seg}mentation model \textbf{to} the open-vocabulary \textbf{change} detection task (\textbf{Seg2Change}). We conducted a series of ablation studies to validate the effectiveness of each component proposed in this paper.

\textbf{Investigation on the impact of the CACH backbone.}
\label{backbone}
We experiment with different backbones, where CACH can be regarded as a lightweight change detection network. As shown in Table \ref{tab:ablation_backbone}, CACH consistently benefits from stronger backbone performance. To achieve a balance between performance and efficiency, we finally adopt DINOv2-Base \cite{dinov2} as the feature extraction backbone. In this work, we do not focus on selecting increasingly powerful backbones to boost category-agnostic change detection performance. Instead, following previous work \cite{unimatchv2}, we choose the DINOv2.

\begin{table*}[ht]
\setlength\tabcolsep{2.5pt}
\centering
\caption{Performance comparison on building change detection datasets. $\dagger$ and $\ddagger$ denote the \textbf{M}–\textbf{C}–\textbf{I} and \textbf{I}–\textbf{M}–\textbf{C} configurations of DynamicEarth. * denotes the variants implemented with SegEarth-OV3. The best and the second-best results are, respectively, marked in \textbf{BOLD} and in \underline{underline}. Except for Kappa, all results are expressed in percentage (\%).}
\scalebox{0.95}{
\begin{tabular}{c|cccccc|cccccc}
\hline
\multirow{2}{*}{Method} & \multicolumn{6}{c|}{WHU-CD} & \multicolumn{6}{c}{LEVIR-CD}  \\
\cline{2-13}
& F1$^c$ & IoU$^c$ & Pre$^c$ & Rec$^c$  & OA & Kappa & F1$^c$ & IoU$^c$ & Pre$^c$ & Rec$^c$  & OA & Kappa \\
\midrule

PCA\_KMeans \cite{pca_km}
& 14.33 & 7.72 & 8.24 & 55.05 & 75.71 & 0.0846 
& 9.96 & 5.24 & 5.70 & 39.50 & 63.60 & 0.0116 \\ 

CVA \cite{cva}
& 7.17 & 3.72 & 3.81 & 62.03 & 40.74 & 0.0023 
& 9.27 & 4.86 & 5.01 & 61.27 & 38.88 & 0.0017 \\ 

DCVA \cite{dcva}
& 20.55 & 11.45 & 12.40 & 59.96 & 82.88 & 0.1537 
& 13.85 & 7.44 & 7.63 & 75.50 & 52.17 & 0.0507 \\ 

UCD-SCM \cite{ucd_scm}
& 32.13 & 19.14 & 20.31 & 76.81 & 88.02 & 0.2792 
& 32.36 & 19.30 & 23.87 & 50.21 & 89.30 & 0.2734 \\ 

AnyChange \cite{anychange}
& 28.13 & 16.37 & 16.48 & \textbf{96.16} & 81.86 & 0.2330 
& 32.68 & 19.53 & 20.46 & 81.06 & 82.98 & 0.2672 \\ 

AnyChange* \cite{anychange}
& 69.25 & 52.96 & 60.83 & 80.38 & 97.36 & 0.6790 
& \underline{72.27} & \underline{56.58} & 59.28 & \textbf{92.54} & 96.38 & \underline{0.7043} \\ 

Inst-CEG \cite{instceg}
& 62.54 & 45.49 & 54.35 & 73.62 & 96.50 & 0.6074 
& 63.29 & 46.30 & 51.55 & 81.96 & 95.16 & 0.6084 \\ 

Inst-CEG* \cite{instceg}
& 71.35 & 55.46 & 66.52 & 76.93 & 97.72 & 0.7016 
& 70.62 & 54.58 & \textbf{74.25} & 67.33 & \underline{97.15} & 0.6912 \\

DynamicEarth$\dagger$ \cite{li2025dynamic_earth}
& 57.35 & 40.20 & 52.33 & 63.43 & 96.26 & 0.5541 
& 46.43 & 30.23 & 33.97 & 73.33 & 91.38 & 0.4242 \\ 

DynamicEarth$\ddagger$ \cite{li2025dynamic_earth}
& 75.85 & 61.09 & \underline{77.84} & 73.96 & 98.13 & 0.7488 
& 69.70 & 53.50 & 62.94 & 78.09 & 96.54 & 0.6789 \\ 

DynamicEarth* \cite{li2025dynamic_earth}
&  \underline{79.66} &  \underline{66.20} & 71.91 & 89.28 & \underline{98.32} & \underline{0.7879}
& 71.97 & 56.21 & 65.71 & 79.55 & 96.84 & 0.7031 \\ 

\midrule

\rowcolor[HTML]{EBEBEB} Seg2Change (Ours)
& \textbf{86.18} & \textbf{75.72} & \textbf{80.86} & \underline{92.25} & \textbf{98.91} & \textbf{0.8562}
& \textbf{78.72} & \textbf{64.91} & \underline{69.73} & \underline{90.38} & \textbf{97.51} & \textbf{0.7742} \\ 

\hline
\end{tabular}
}
\label{tab:results_detail_bcd}
\end{table*}

\begin{table*}[ht]
\setlength\tabcolsep{2.5pt}
\centering
\caption{Performance comparison on land-cover change detection datasets. $\dagger$ and $\ddagger$ denote the \textbf{M}–\textbf{C}–\textbf{I} and \textbf{I}–\textbf{M}–\textbf{C} configurations of DynamicEarth. * denotes the variants implemented with SegEarth-OV3. The best and the second-best results are, respectively, marked in \textbf{BOLD} and in \underline{underline}. Except for Kappa, all results are expressed in percentage (\%).}
\scalebox{0.95}{
\begin{tabular}{c|cccccc|cccccc}
\hline
\multirow{2}{*}{Method} & \multicolumn{6}{c|}{DSIFN} & \multicolumn{6}{c}{CLCD}  \\
\cline{2-13}
& F1$^c$ & IoU$^c$ & Pre$^c$ & Rec$^c$ & OA & Kappa & F1$^c$ & IoU$^c$ & Pre$^c$ & Rec$^c$ & OA & Kappa \\
\midrule

PCA\_KMeans \cite{pca_km}
& 35.48 & 21.57 & 29.14 & 45.35 & 71.98 & 0.1865
& 18.42 & 10.14 & 11.90 & 40.76 & 73.13 & 0.0780 \\

CVA \cite{cva}
& 27.88 & 16.20 & 17.65 & \underline{66.29} & 41.72 & 0.0142
& 13.39 & 7.17 & 7.53 & 60.42 & 41.82 & 0.0017 \\

DCVA \cite{dcva}
& 38.52 & 23.85 & 25.76 & \textbf{76.31} & 58.60 & 0.1758
& 20.81 & 11.62 & 12.09 & \textbf{74.77} & 57.67 & 0.0918 \\

UCD-SCM \cite{ucd_scm}
& 40.13 & 25.10 & 32.82 & 51.61 & 73.83 & 0.2443
& 23.31 & 13.19 & 18.18 & 32.49 & 84.09 & 0.1522 \\

AnyChange \cite{anychange}
& 39.19 & 24.37 & 31.36 & 52.25 & 72.45 & 0.2280
& 31.96 & 19.02 & 21.33 & 63.63 & 79.84 & 0.2342 \\

AnyChange* \cite{anychange}
& \underline{54.69} & \underline{37.64} & 72.04 & 44.07 & \textbf{87.59} & \underline{0.4798}
& 27.55 & 15.98 & 33.05 & 23.62 & 90.76 & 0.2276 \\

Inst-CEG \cite{instceg}
& 31.81 & 18.91 & 54.50 & 22.46 & 83.64 & 0.2430
& 6.76 & 3.50 & 9.73 & 5.18 & 89.37 & 0.0020 \\

Inst-CEG* \cite{instceg}
& 47.21 & 30.90 & 65.56 & 36.88 & 85.98 & 0.3985
& 10.09 & 5.32 & 19.41 & 6.82 & 90.96 & 0.0648 \\

DynamicEarth$\dagger$ \cite{li2025dynamic_earth}
& 54.35 & 37.32 & 69.96 & 44.44 & \underline{87.32} & 0.4741
& 23.83 & 13.52 & 30.70 & 19.47 & 90.74 & 0.1916 \\

DynamicEarth$\ddagger$ \cite{li2025dynamic_earth}
& 26.42 & 15.22 & \textbf{90.96} & 15.45 & 85.37 & 0.2260
& 14.97 & 8.09 & \textbf{83.52} & 8.22 & \underline{93.05} & 0.1382 \\

DynamicEarth* \cite{li2025dynamic_earth}
& 39.30 & 24.45 & \underline{86.75} & 25.40 & 86.66 & 0.3423
& \underline{38.16} & \underline{23.58} & \underline{65.39} & 26.94 & \textbf{93.50} & \underline{0.3535} \\

\midrule

\rowcolor[HTML]{EBEBEB} Seg2Change (Ours)
& \textbf{58.56} & \textbf{41.40} & 62.76 & 54.88 & 86.80 & \textbf{0.5075}
& \textbf{47.89} & \textbf{31.48} & 37.32 & \underline{66.79} & 89.18 & \textbf{0.4239} \\

\hline
\end{tabular}
}
\label{tab:results_detail_bcd_2}
\end{table*}

% \textbf{Impact of Different CACH Pretrain Dataset.}
% We investigate the impact of the category-agnostic CD pretrain dataset on the OV-SCD performance in Table \ref{tab:ablation_CACD}.  The original bi-temporal images and coarse annotations of CACD are derived from the SECOND \cite{second}, JL1-CD \cite{liu2025jl1}, and CNAM-CD \cite{rs15092464} datasets. We further expand and re-annotate the original coarse labels. These datasets have limited records of change categories, \textit{e.g.}, only 6 categories recorded by SECOND. We generalize the category definitions in these datasets to a class-agnostic setting. Any semantic category transition is recorded as a change in the corresponding change map. Benefiting from the enriched class-agnostic change annotations, Seg2Change attains superior performance in OVCD tasks.

\textbf{Impact on Different Implementation of OVSS Model.}
Tables \ref{tab:ablation_OVSS_bcd}-\ref{tab:ablation_OVSS_lcd} examine the impact of different OVSS model implementations on the performance of Seg2Change in the context of building and land-cover change detection. Through our class-agnostic change detection paradigm, Seg2Change is able to fully leverage the increasingly powerful OVSS models. Our approach does not require threshold fine-tuning to adapt to different OVSS models. A single architecture can be applied to all change detection tasks without the need to fine-tune the combination of OVSS models. Our CACH can be regarded as a post-processing module that is compatible with any OVSS model.

\textbf{Investigation on the impact of $(C_2, C_5, C_8, C_{11})$.} 
We analyze how the input channel dimensions at different processing layers of BDFM and EDQA affect change detection performance, by employing layers ranging from low to high dimensionality. Table \ref{tab:ablation_channel} indicates that higher-dimensional settings of $(C_2, C_5, C_8, C_{11})$ yield better performance, but also result in increased parameter counts and GPU memory consumption. To balance efficiency and performance, we adopt the configuration (48, 64, 80, 96).

\begin{table*}[ht]
\setlength\tabcolsep{2.2pt}
\centering
\caption{Performance comparison on the SC-SCD dataset. The best and the second-best results are, respectively, marked in \textbf{BOLD} and in \underline{underline}. Except for Kappa, all results are expressed in percentage (\%).}
\scalebox{0.88}{
\begin{tabular}{c|cc|cc|cc|cc|cc|cc|cc|cccc}
\hline
\multirow{2}{*}{Method}
& \multicolumn{2}{c|}{Bareland}
& \multicolumn{2}{c|}{Water}
& \multicolumn{2}{c|}{Building}
& \multicolumn{2}{c|}{Structure}
& \multicolumn{2}{c|}{Farmland}
& \multicolumn{2}{c|}{Vegetation}
& \multicolumn{2}{c|}{Road}
& \multicolumn{4}{c}{Average} \\
\cline{2-19}
& F1$^c$ & IoU$^c$
& F1$^c$ & IoU$^c$
& F1$^c$ & IoU$^c$
& F1$^c$ & IoU$^c$
& F1$^c$ & IoU$^c$
& F1$^c$ & IoU$^c$
& F1$^c$ & IoU$^c$
& mF1$^c$ & mIoU$^c$ & mOA & mKappa \\
\midrule

UCD-SCM \cite{ucd_scm}
& 16.13 & 8.77
& 3.32 & 1.69
& 15.00 & 8.11
& \underline{18.33} & \underline{10.09}
& 9.18 & 4.81
& 16.52 & 9.01
& 5.95 & 3.06
& 12.06 & 6.51 & 85.95 & 0.0619 \\

AnyChange \cite{anychange}
& 28.01 & 16.29
& 4.22 & 2.16
& 19.88 & 11.03
& 18.30 & 10.07
& 11.19 & 5.93
& 30.29 & 17.85
& 3.71 & 1.89
& 16.51 & 9.32 & 78.27 & 0.1056 \\

AnyChange* \cite{anychange}
& 24.38 & 13.88
& 3.42 & 1.74
& 35.91 & 21.88
& 17.94 & 9.85
& 9.27 & 4.86
& 21.92 & 12.31
& 3.63 & 1.85
& 16.64 & 9.48 & 88.06 & 0.1192 \\

Inst-CEG \cite{instceg}
& 0.00 & 0.00
& 14.68 & 7.92
& 10.34 & 5.45
& 0.00 & 0.00
& 0.00 & 0.00
& 12.71 & 6.79
& 10.58 & 5.58
& 6.90 & 3.68 & 94.40 & 0.0551 \\

Inst-CEG* \cite{instceg}
& 18.41 & 10.14
& \underline{38.14} & \underline{23.57}
& \underline{53.88} & \underline{36.87}
& 2.48 & 1.25
& \underline{16.53} & \underline{9.01}
& 21.88 & 12.28
& 13.70 & 7.36
& 23.57 & 14.35 & 94.49 & 0.2139 \\

DynamicEarth$\dagger$ \cite{li2025dynamic_earth}
& \underline{36.85} & \underline{22.59}
& \textbf{49.78} & \textbf{33.14}
& 44.42 & 28.55
& \textbf{23.33} & \textbf{13.21}
& 10.39 & 5.48
& \underline{33.01} & \underline{19.76}
& 5.97 & 3.08
& \underline{29.11} & \underline{17.97} & 91.37 & \underline{0.2532} \\

DynamicEarth$\ddagger$ \cite{li2025dynamic_earth}
& 0.00 & 0.00
& 21.97 & 12.34
& 23.87 & 13.55
& 0.00 & 0.00
& 0.00 & 0.00
& 0.00 & 0.00
& 18.54 & 10.22
& 9.19 & 5.16 & \underline{95.25} & 0.0877 \\

DynamicEarth* \cite{li2025dynamic_earth}
& 16.90 & 9.23
& 19.43 & 10.76
& 44.17 & 28.35
& 12.16 & 6.47
& 6.98 & 3.62
& 19.00 & 10.50
& \underline{20.43} & \underline{11.38}
& 19.87 & 11.47 & 95.03 & 0.1804 \\

\midrule

\rowcolor[HTML]{EBEBEB} Seg2Change (Ours)
& \textbf{43.15} & \textbf{27.51}
& 35.92 & 21.89
& \textbf{64.13} & \textbf{47.20}
& 1.53 & 0.77
& \textbf{20.47} & \textbf{11.40}
& \textbf{45.97} & \textbf{29.84}
& \textbf{38.60} & \textbf{23.91}
& \textbf{35.68} & \textbf{23.22} & \textbf{95.82} & \textbf{0.3385} \\

\hline
\end{tabular}
}
\label{tab:results_scscd}
\end{table*}

\begin{table*}[ht]
\setlength\tabcolsep{2.2pt}
\centering
\caption{Performance comparison on the SECOND dataset. The best and the second-best results are, respectively, marked in \textbf{BOLD} and in \underline{underline}. Except for Kappa, all results are expressed in percentage (\%).}
\scalebox{0.88}{
\begin{tabular}{c|cc|cc|cc|cc|cc|cc|cccc}
\hline
\multirow{2}{*}{Method}
& \multicolumn{2}{c|}{Building}
& \multicolumn{2}{c|}{Tree}
& \multicolumn{2}{c|}{Water}
& \multicolumn{2}{c|}{Vegetation}
& \multicolumn{2}{c|}{N.v.g Surface}
& \multicolumn{2}{c|}{Playground}
& \multicolumn{4}{c}{Average} \\
\cline{2-17}
& F1$^c$ & IoU$^c$
& F1$^c$ & IoU$^c$
& F1$^c$ & IoU$^c$
& F1$^c$ & IoU$^c$
& F1$^c$ & IoU$^c$
& F1$^c$ & IoU$^c$
& mF1$^c$ & mIoU$^c$ & mOA & mKappa \\
\midrule

UCD-SCM \cite{ucd_scm}
& 36.15 & 22.06
& 6.50 & 3.36
& 2.20 & 1.11
& 18.77 & 10.35
& 22.83 & 12.88
& 1.31 & 0.66
& 14.63 & 8.40 & 78.61 & 0.0795 \\

AnyChange \cite{anychange}
& 42.79 & 27.22
& 7.49 & 3.89
& 2.19 & 1.11
& 24.82 & 14.17
& 38.90 & 24.15
& 1.03 & 0.52
& 19.54 & 11.84 & 76.07 & 0.1252 \\

AnyChange* \cite{anychange}
& \underline{58.58} & \underline{41.42}
& 3.93 & 2.01
& 0.97 & 0.49
& 21.46 & 12.02
& 41.69 & 26.33
& 1.17 & 0.59
& 21.30 & 13.81 & 77.97 & 0.1492 \\

Inst-CEG \cite{instceg}
& 38.27 & 23.66
& 17.59 & 9.64
& 18.24 & 10.03
& 1.14 & 0.57
& 0.00 & 0.00
& 31.70 & 18.84
& 17.82 & 10.46 & 92.79 & 0.1604 \\

Inst-CEG* \cite{instceg}
& 63.61 & 46.63
& \textbf{28.01} & \textbf{16.28}
& 22.67 & 12.78
& 25.61 & 14.69
& 14.12 & 7.60
& 22.08 & 12.41
& 29.35 & 18.40 & 93.62 & 0.2670 \\

DynamicEarth$\dagger$ \cite{li2025dynamic_earth}
& 55.70 & 38.60
& 26.89 & 15.54
& \underline{26.64} & \underline{15.37}
& \textbf{35.06} & \textbf{21.25}
& \underline{43.80} & \underline{28.04}
& \underline{36.97} & \underline{22.67}
& \underline{37.51} & \underline{23.58} & 91.88 & \underline{0.3297} \\

DynamicEarth$\ddagger$ \cite{li2025dynamic_earth}
& 48.19 & 31.74
& 19.64 & 10.89
& 21.97 & 12.34
& 0.79 & 0.40
& 0.00 & 0.00
& \textbf{42.43} & \textbf{26.93}
& 22.17 & 13.72 & 93.76 & 0.2109 \\

DynamicEarth* \cite{li2025dynamic_earth}
& 51.96 & 35.10
& 13.74 & 7.38
& 22.02 & 12.37
& 23.21 & 13.13
& 26.27 & 15.12
& 15.37 & 8.32
& 25.43 & 15.24 & \underline{93.83} & 0.2286 \\

\midrule

\rowcolor[HTML]{EBEBEB} Seg2Change (Ours)
& \textbf{76.36} & \textbf{61.76}
& \underline{27.73} & \underline{16.10}
& \textbf{37.23} & \textbf{22.88}
& \underline{31.49} & \underline{18.69}
& \textbf{54.29} & \textbf{37.26}
& 30.23 & 17.81
& \textbf{42.89} & \textbf{29.08} & \textbf{95.17} & \textbf{0.4045} \\

\hline
\end{tabular}
}
\label{tab:results_second}
\end{table*}

\textbf{Investigation on the impact of different implementations of the feature difference module.}
We evaluate the effect of various difference feature extraction modules on the final change detection performance, including absolute feature subtraction, the temporal feature interaction module (TFIM) \cite{TFIM}, the cross-domain difference enhancement module (CDEM) \cite{CDEM}, and the multiscale difference fusion module (MDFM) \cite{MDFM}. As shown in Table \ref{tab:ablation_bdfm}, our BDFM achieves superior performance and enhances the guidance of feature differences.

\textbf{Investigation on the performance, computational resource usage, and inference time cost.}
As shown in Table \ref{tab:parameters}, Seg2Change exhibits notable superiority over existing methods with respect to both performance and computational efficiency. On the SC-SCD dataset, Seg2Change attains the highest accuracy, markedly surpassing DynamicEarth, while also reducing GPU memory consumption and inference latency. In contrast, prior approaches either suffer from inferior accuracy or require substantially greater computational resources. Seg2Change introduces only a lightweight set of learnable parameters (3.9M), underscoring its efficiency. These results indicate that Seg2Change achieves a superior balance between effectiveness and efficiency, making it a more practical and scalable solution for real-world OVCD tasks.

\section{More Detailed Experiment Results}
\label{More_Detailed_Experiment_Results}
\textbf{Building Change Detection.} As shown in Table \ref{tab:results_detail_bcd}, Seg2Change consistently achieves the best performance on both the WHU-CD and LEVIR-CD datasets, demonstrating strong effectiveness and robustness in building change detection. Specifically, on the WHU-CD dataset, our method surpasses all competitors, achieving an F1$^c$ of 86.18\% and an IoU$^c$ of 75.72\%, and outperforming the previous best method, DynamicEarth*, by a clear margin. Meanwhile, Seg2Change achieves the highest OA of 98.91\% and Kappa of 0.8562, indicating more reliable and consistent predictions. On the LEVIR-CD dataset, our method similarly attains the best F1$^c$ of 78.72\% and IoU$^c$ of 64.91\%, while maintaining competitive precision and achieving notably high recall of 90.38\%. Compared with existing OVCD methods, the superior recall suggests that Seg2Change is more effective at capturing complete change regions.

% These improvements can be attributed to the proposed decoupled design, which effectively leverages open-vocabulary semantic representations and mitigates error propagation, thereby enhancing both accuracy and generalization across datasets.

\begin{figure*}[t]
\centering
\includegraphics[width=0.98\textwidth]{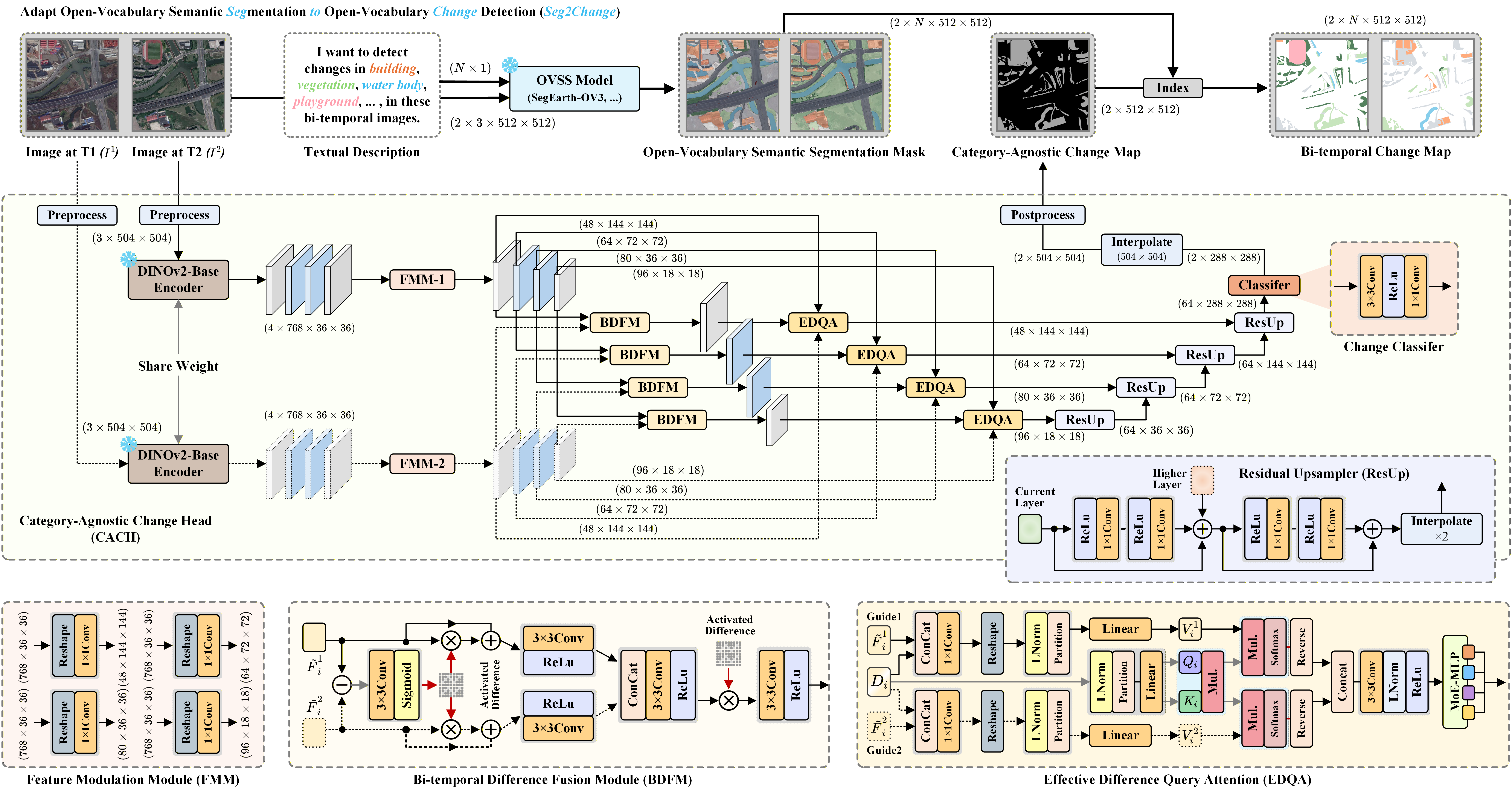}
\caption{Overall of Seg2Change. The bi-temporal remote sensing images are first fed into the category-agnostic change head (CACH) to generate a category-agnostic change map. In parallel, the same images, along with textual descriptions, are fed into an OVSS model to generate semantic segmentation maps. The final change map is derived by combining them.}
\label{Detailed}
\end{figure*}

\textbf{Land-Cover Change Detection.} As demonstrated in Table \ref{tab:results_detail_bcd_2}, Seg2Change consistently outperforms the second-best methods across both datasets, demonstrating clear quantitative advantages. On the DSIFN dataset, compared with the second-best method AnyChange*, Seg2Change improves F1$^c$ from 54.69\% to 58.56\%, achieving an absolute gain of 3.87\%, and increases IoU$^c$ from 37.64\% to 41.40\%, yielding a gain of 3.76\%. Although AnyChange* attains the highest OA, its relatively low recall limits its ability to capture complete change regions, whereas Seg2Change improves recall by 10.81\%, leading to more comprehensive detection results.

On the more challenging CLCD dataset, Seg2Change surpasses the second-best DynamicEarth* by a larger margin. Specifically, F1$^c$ increases from 38.16\% to 47.89\%, corresponding to an improvement of 9.73\%, and IoU$^c$ rises from 23.58\% to 31.48\%, with a gain of 7.90\%. Notably, Seg2Change improves recall by 39.85\%, indicating a significantly stronger ability to capture diverse and complex change patterns. These improvements demonstrate that Seg2Change achieves a better balance between precision and recall, resulting in more reliable and complete change detection performance.

\textbf{Semantic Change Detection.} As shown in Tables \ref{tab:results_scscd} and \ref{tab:results_second}, Seg2Change demonstrates clear and consistent advantages over the second-best methods on both the SC-SCD and the SECOND datasets, highlighting its strong capability for semantic change detection across diverse categories.

On the SC-SCD dataset, compared with the second-best method DynamicEarth$\dagger$, Seg2Change improves the average F1$^c$ from 29.11\% to 35.68\%, achieving a gain of 6.57\%, and increases mIoU$^c$ from 17.97\% to 23.22\%, yielding an improvement of 5.25\%. Meanwhile, Kappa is improved by 0.0853. At the category level, Seg2Change achieves substantial gains, particularly on building, where F1$^c$ improves by 19.71\%, and on vegetation and road, with gains of 12.96\% and 18.17\%, respectively. Although DynamicEarth$\dagger$ performs strongly on certain categories such as water and structure, Seg2Change provides more balanced performance across categories, leading to superior overall results.

On the SECOND dataset, Seg2Change further enlarges its advantage over the second-best DynamicEarth$\dagger$. Specifically, mF1$^c$ increases from 37.51\% to 42.89\%, corresponding to an improvement of 5.38\%, and mIoU$^c$ rises from 23.58\% to 29.08\%, with a gain of 5.50\%. Notably, Seg2Change achieves significant improvements on key categories such as building and non-vegetated ground surface, with F1$^c$ gains of 20.66\% and 10.49\%, respectively. These results indicate that Seg2Change is more effective at capturing complex and large-scale changes. Overall, the consistent improvements across both datasets demonstrate that Seg2Change achieves a better trade-off between category-wise discrimination and overall consistency, resulting in more robust and reliable open-vocabulary semantic change detection performance.

\section{Detailed Architecture}
\label{Detailed_Architecture}
As illustrated in Fig. \ref{Detailed}, we present the detailed architecture of Seg2Change. This framework aims to adapt open-vocabulary semantic segmentation models for remote sensing to the task of open-vocabulary change detection. Our category-agnostic change head is employed to extract changes of arbitrary categories from bi-temporal images. Semantic segmentation maps are generated using bi-temporal images, textual descriptions, and an open-vocabulary semantic segmentation model. Finally, by integrating the category-agnostic change map, the corresponding semantic changes for each temporal image are derived.

We extend conventional change detection to support open-vocabulary capability. Traditional change detection methods are generally restricted to specific or single-category scenarios. Despite the advancement of open-vocabulary semantic segmentation models, there remains a lack of a simple and effective paradigm to directly apply these models to change detection.

\subsection{Overall of Seg2Change}
The input of open-vocabulary change detection consists of a pair of bi-temporal images $I^1, I^2$ and a text description of interest $\mathbf{x}_{text}$. We feed the bi-temporal images $I^1, I^2$ into CACH to obtain the category-agnostic change map $M_{\text{ca}}$. Meanwhile, the bi-temporal images $I^1, I^2$, together with the text description $\mathbf{x}_{text}$, are input into the OVSS model (\textit{i.e.}, SegEarth-OV3 \cite{li2025segearthov3}) to generate the semantic segmentation maps $M^1, M^2$. For open-vocabulary semantic change detection tasks (\textit{e.g.}, SC-SCD \cite{SC-SCD} and SECOND \cite{second}), semantic change maps for bi-temporal images are modeled as follows:
\begin{align}
M_{\text{ch}}^t = M_{\text{ca}} \cdot M^t,
\end{align}
where $t \in \{1, 2\}$ denotes the temporal index. For open-vocabulary binary change detection tasks (\textit{e.g.}, WHU-CD \cite{whucd}, LEVIR-CD \cite{levircd}, DSIFN \cite{DSIFN}, and CLCD \cite{clcd}), the binary change maps for bi-temporal images need to be integrated:
\begin{align}
M_{\text{ch}} = M_{\text{ca}} \cdot M^1 + M_{\text{ca}} \cdot M^2.
\end{align}

\subsection{Category-Agnostic Change Map}
In the following, we present the procedure for extracting the category-agnostic change map. As the DINOv2 \cite{dinov2} encoder adopts a patch size of 14, the input image dimensions are required to be divisible by 14. We thus preprocess the data into a tensor of size $(2 \times 3 \times 504 \times 504)$, where the leading dimension of 2 corresponds to the bi-temporal inputs. Subsequently, pyramid features are obtained using the DINOv2-Base image encoder $\Phi_{\text{DINO}}$:
\begin{align}
\{F_2^t, F_5^t, F_8^t, F_{11}^t\} = \Phi_{\text{DINO}}(I^t).
\end{align}

We adopt the feature representations from the 2nd, 5th, 7th, and 11th layers of DINOv2 image encoder, as previously analyzed in Sec. \ref{backbone} and thus omitted here for brevity. Rather than heavily fine-tuning $\Phi_{\text{DINO}}$ with additional parameters, we introduce a simple Feature Modulation Module (\textbf{FMM}) to better align the features with remote sensing characteristics, as formulated in Eq. (\ref{resize}):
\begin{align}
\tilde{F}_i^t = \text{Resize}_{i} (\mathrm{Conv_{1 \times 1}} * (F_i^t)),
\label{resize}
\end{align}
where $\mathrm{Conv_{1 \times 1}}$ denotes a 2D convolution with a kernel size of $1 \times 1$, and $*$ represents the convolution operation. $\text{Resize}_i$ represents the scaling operation for the $i$-th layer, defined as ${\mathrm{Deconv_{4 \times 4}(\cdot)}, \mathrm{Deconv_{2 \times 2}(\cdot)}, \mathrm{Identity}(\cdot), \mathrm{Conv}_{3\times3}(\cdot)}$, corresponding to upsampling by a factor of 4, upsampling by a factor of 2, keeping the scale unchanged, and downsampling by a factor of 2 (with stride 2), respectively. $\mathrm{Deconv_{d \times d}}$ denotes a 2D transposed convolution with a kernel size of $\mathrm{d \times d}$, while $\mathrm{Identity}(\cdot)$ indicates no change in scale. The modulated features $\tilde{F}_i^t$ are then subjected to difference feature enhancement and calibration. We utilize the proposed Bi-temporal Difference Fusion Module (\textbf{BDFM}) to extract and enhance difference features from bi-temporal images. 

First, we employ convolutional activations to extract spatial variations from the modulated features $\tilde{F}_i^1$ and $\tilde{F}_i^2$, generating the $i$-th layer difference attention:
\begin{align}
Att_i = \sigma(\mathrm{Conv_{3 \times 3}} * |\tilde{F}_i^1 - \tilde{F}_i^2|),
\end{align}
where $\sigma$ denotes the $\texttt{Sigmoid}$ function. Subsequently, the difference attention $Att_i$ is used to enhance the discrepant regions of the bi-temporal images, respectively:
\begin{align}
X_i^t = \gamma\mathrm{Conv_{3 \times 3}} &* (\tilde{F}_i^t + Att_i \cdot \tilde{F}^t_i), \\ D_i = \gamma\mathrm{Conv_{3 \times 3}} * (\gamma&\mathrm{Conv_{3 \times 3}} * (X_i^1 || X_i^2) \cdot Att_i),
\end{align}
where $X_i^t$ are bi-temporal feature differences, $D_i$ denotes the fused feature differences at the $i$-th layer. $\gamma$ and $||$ are, respectively, the $\texttt{ReLu}$ activation function and the concatenation operation. The enhancement of feature differences inevitably introduces noise and pseudo changes. To mitigate this issue, we further employ the proposed Effective Difference Query Attention (\textbf{EDQA}) module to calibrate the feature differences. Here, we transform the modulated features $\tilde{F}_i^t$ into the guidance features $\tilde{G}_i^t$, as they can provide effective guidance for refining the difference features:
\begin{align}
\tilde{G}_i^t = \mathrm{Conv_{1 \times 1}} * (D_i || \tilde{F}_i^t).
\end{align}

A sliding-window attention mechanism \cite{9710580} is adopted to model the attention between the difference features $D_i$ and the guidance features $\tilde{G}_i^t$. For simplicity, the sliding-window processing is omitted, and the formulation is expressed in Eq. (\ref{window_att}):
\begin{align}Q_i, K_i = \phi_{qk_i}(D_i)&, V_i^t = \phi_{v_i}^t(\tilde{G}_i^t), \\
\tilde{D}_i^t = \phi_{\text{proj}}^t(\text{Softmax}((Q_i &\cdot K_i) / \sqrt{d_i}) \cdot V_i^t),
\label{window_att}
\end{align}
where $\phi_{qk_i}$ are linear projections for $D_i$, while $\phi_{v_i}^t$ and $\phi_{\text{proj}}^t$ are linear projections for $\tilde{G}_i^t$ and $\tilde{D}_i^t $ at time $t$. $\tilde{D}_i^t$ denotes the query calibrated difference at the $i$-th layer of the image at time $t$. The calibrated bi-temporal differences obtained from the query are then reconstructed by reversing the window partition and fused as:
\begin{align}
\tilde{D}_i = \gamma\mathrm{Conv{3 \times 3}} * (\gamma\mathrm{Conv_{3 \times 3}} * (\tilde{D}_i^1 || \tilde{D}_i^2)),
\label{fuse_att}
\end{align}
where $\tilde{D}_i$ denotes the calibrated bi-temporal feature difference at the $i$-th layer. Owing to differences in dataset sources, variations in imaging conditions and seasonal characteristics inevitably arise. To accommodate bi-temporal images from heterogeneous sources, we augment the MLP with a Mixture-of-Experts (MoE) module \cite{Jia_M4oE_MICCAI2024}. The calibrated bi-temporal feature difference $\tilde{D}_i$ are utilized to estimate expert weights and to compute the outputs of experts:
\begin{align}
\text{weight}_G = \text{Softmax}(W_g \cdot \tilde{D}_i &+ \mathbf{b}_g), \\ \text{expert}_j(\tilde{D}_i) = W_{o,j} \cdot \text{GELU}(W_{h,j} \cdot \tilde{D}_i &+ \mathbf{b}_{h,j}) + \mathbf{b}_{o,j},
\end{align}
where $\text{weight}_G$ denotes the weight assigned to each expert, and $\text{expert}_j(\tilde{D}_i)$ represents the output of expert-$j$ given the feature difference $\tilde{D}_i$. $W_g$ and $\mathbf{b}_g$ are the learnable parameters used to compute expert weights. $W_{h,j}$, $\mathbf{b}_{h,j}$ and $W_{o,j}$, $\mathbf{b}_{o,j}$ are the parameters of the two linear mapping layers for the output of expert-$j$. Finally, the weighted outputs of the $N_e$ experts are aggregated to produce the feature difference $X_i^d$, which is formulated as Eq. (\ref{eq:xd}):
\begin{align}
X_i^d = \textstyle\sum_{j=1}^{N_e} \text{weight}_j \cdot \text{expert}_j(\tilde{D}_i).
\label{eq:xd}
\end{align}

\begin{table*}[ht]
\setlength\tabcolsep{3pt}
\centering
\caption{Information on the evaluation benchmark datasets and the CA-CDD coarse label source datasets. * denotes that the initially coarse, category-restricted change annotations are refined to produce category-agnostic change maps. ** indicates that all images have been preprocessed and cropped to a resolution of 512$\times$512.}
\scalebox{0.95}{
\begin{tabular}{l|c|c|c|c|c}
\hline
Dataset & Study Site & Number of Image Pairs & Image Size** & Resolution & Evaluation Task \\
\hline
WHU-CD \cite{whucd} & Christchurch, New Zealand & 660 (Test)  & $512 \times 512$ & 0.3 m & Building CD \\
LEVIR-CD \cite{levircd} & Texas, U.S. & 512 (Test) & $512 \times 512$ & 0.5 m& Building CD \\
DSIFN \cite{DSIFN} & Xi'an, China & 48 (Test) & $512 \times 512$ & 0.03--1 m& Land-cover CD \\
CLCD \cite{clcd} & Guangdong, China & 120 (Test)  & $512 \times 512$ & 0.5--2 m & Land-cover CD \\
SECOND \cite{second} & Hangzhou, Chengdu, Shanghai, China & 1694 (Test) & $512 \times 512$ & 0.5--3 m & Semantic CD \\
SC-SCD \cite{SC-SCD} & Longwen, Zhangzhou, China & 322 (Test)  & $512 \times 512$ & 0.5 m & Semantic CD \\
\hline
SECOND \cite{second} & Hangzhou, Chengdu, Shanghai, China & 2968* (Train) & $512 \times 512$ & 0.5--3 m & Semantic CD \\
JL1-CD \cite{liu2025jl1} & Multiple regions in China & 1000* (Train) & $512 \times 512$ & 0.5--0.75 m & Land-cover CD \\
CNAM-CD \cite{rs15092464} & 23 SLNAs \cite{rs15092464} & 1000* (Train) & $512 \times 512$ & 0.5 m & Land-cover CD \\
\hline
\end{tabular}
}
\label{tab:cacdd}
\end{table*}

\begin{figure*}[t]
\centering
\includegraphics[width=0.84\textwidth]{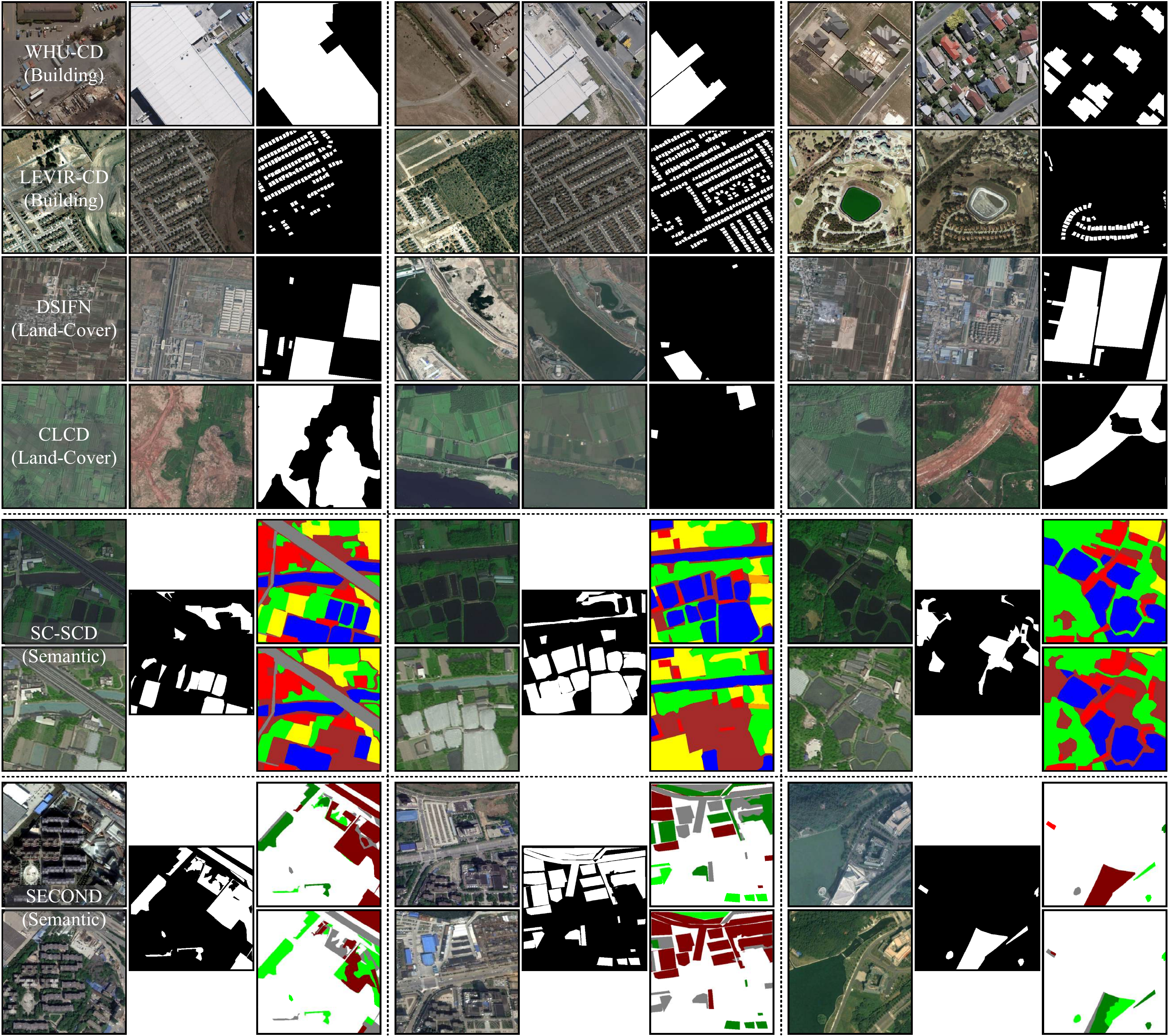}
\caption{Example images from binary change detection datasets (WHU-CD, LEVIR-CD, DSIFN, and CLCD) and semantic change detection datasets (SC-SCD and SECOND). The binary change detection datasets contain only binary change maps for specific categories. In contrast, the semantic change detection datasets include both binary change maps indicating changed regions between bi-temporal images and multi-class change maps that capture category transitions across time.}
\label{Datasets_Test}
\end{figure*}

\begin{figure*}[t]
\centering
\includegraphics[width=0.84\textwidth]{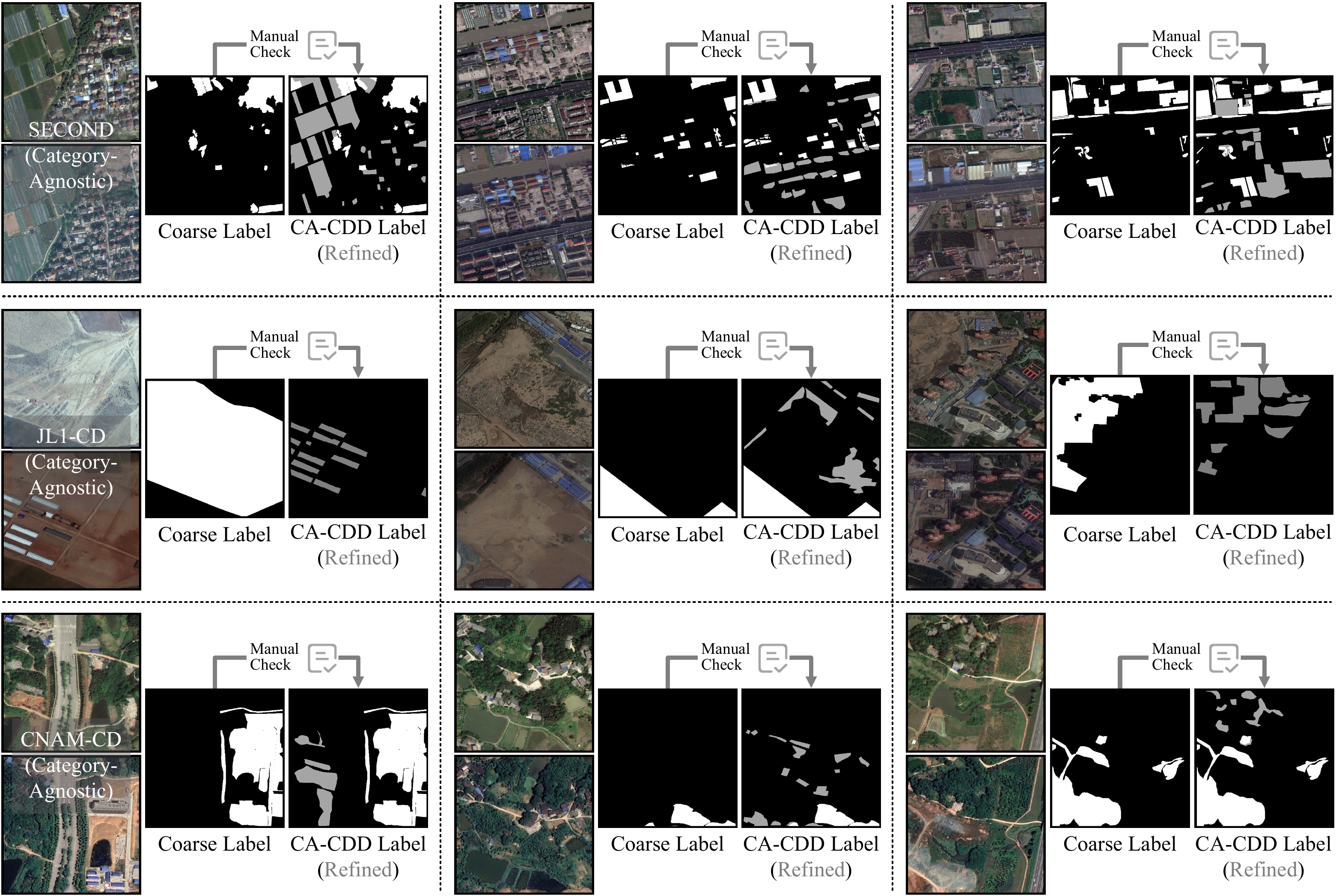}
\caption{Visual comparison between CA-CDD category-agnostic change labels and the original coarse labels. We have improved the limited category range of the original labels. At the same time, we have refined the coarse-grained range annotations in the labels to fine-grained annotations. The \textcolor[HTML]{A5A5A5}{gray} markings in the figure represent the refinements we made to the original labels.}
\label{Datasets_Train}
\end{figure*}

Finally, we upsample the feature differences $X_i^d$ from each layer to obtain the final category-agnostic change map. To ensure compatibility between the inputs and outputs of different modules, the difference features are further modulated:
\begin{align}
\tilde{X}_i^d = \mathrm{Conv_{1 \times 1}} * (X_i^d).
\end{align}

Then, we apply the proposed Residual Upsampler (\textbf{ResUp}) to upsample the features at each layer:
\begin{align}
\tilde{X}_{i}^D = \phi_{2\times}(\mathrm{ResConv}_2(\mathrm{ResConv}_1(\tilde{X}_i^d) + \tilde{X}_{i-1}^D)),
\end{align}
where $\tilde{X}_{i}^D$ denotes the upsampled feature at the $i$-th layer, and $\phi_{2\times}$ represents a bilinear interpolation function that doubles the spatial resolution. During the upsampling process, two residual convolutional units are applied to the modulated difference features, and each residual convolutional unit is defined as follows:
\begin{align}
\mathrm{ResConv}_k = \mathrm{Conv_{1 \times 1}} * (\gamma\mathrm{Conv_{1 \times 1}} *(\gamma (\tilde{X}_k^d))) + \tilde{X}_k^d,
\end{align}
where $\tilde{X}_k^d$ denotes the input tensor to $\text{ResConv}_k$. After progressive upsampling across layers, the aggregated feature $\tilde{X}^D$ is obtained, integrating differences from all levels. Finally, the category-agnostic change prediction $P_{\text{ca}}$ is generated via an output convolutional unit and resized to the original image resolution:
\begin{align}
P_{\text{ca}} = \phi_{H \times W}(\mathrm{Conv_{1 \times 1}} * (\gamma\mathrm{Conv_{3 \times 3}} *(\tilde{X}^D))).
\end{align}

An $\mathrm{arg~max}$ operation is then applied to the prediction to obtain the category-agnostic change map.
\begin{align}
M_{\text{ca}} = \mathrm{arg~max}(P_{\text{ca}}),
\end{align}
where the category-agnostic change prediction $P_{\text{ca}}$ has a dimension of $\mathbb{R}^{2 \times H \times W}$, and the category-agnostic change map $M_{\text{ca}}$ has a dimension of $\mathbb{R}^{1 \times H \times W}$.

\begin{figure*}[t]
\centering
\includegraphics[width=0.92\textwidth]{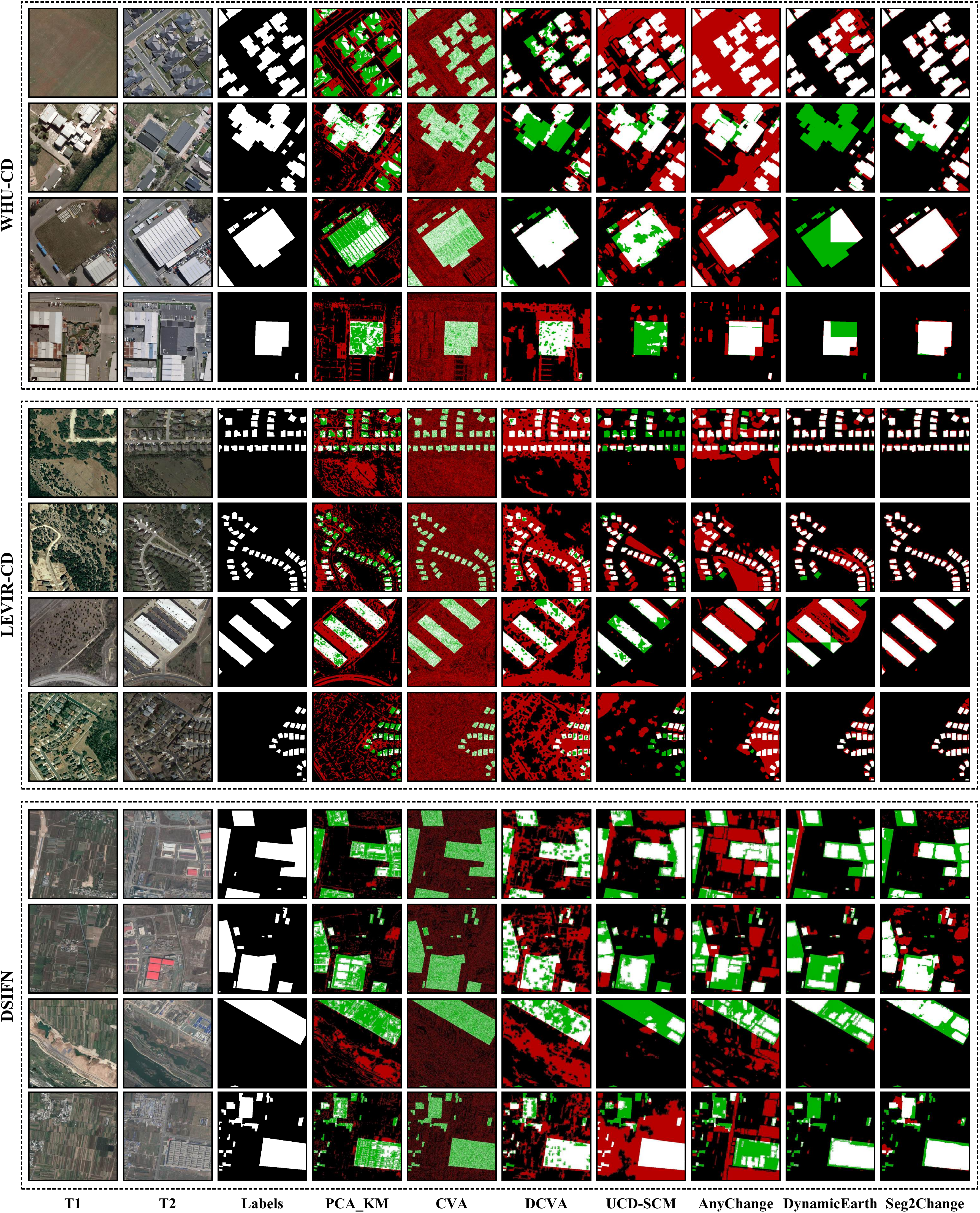}
\caption{Qualitative comparison of several representative UCD and OVCD methods on three binary change detection datasets, \textit{i.e.}, WHU-CD, LEVIR-CD, and DSIFN. White represents a true positive, black is a true negative, \textcolor[HTML]{00B400}{green} indicates a false positive, and \textcolor[HTML]{B80000}{red} is a false negative. Fewer \textcolor[HTML]{00B400}{green} and \textcolor[HTML]{B80000}{red} pixels represent better performance.}
\label{Binary_Vis1}
\end{figure*}
\begin{figure*}[t]
\centering
\includegraphics[width=0.92\textwidth]{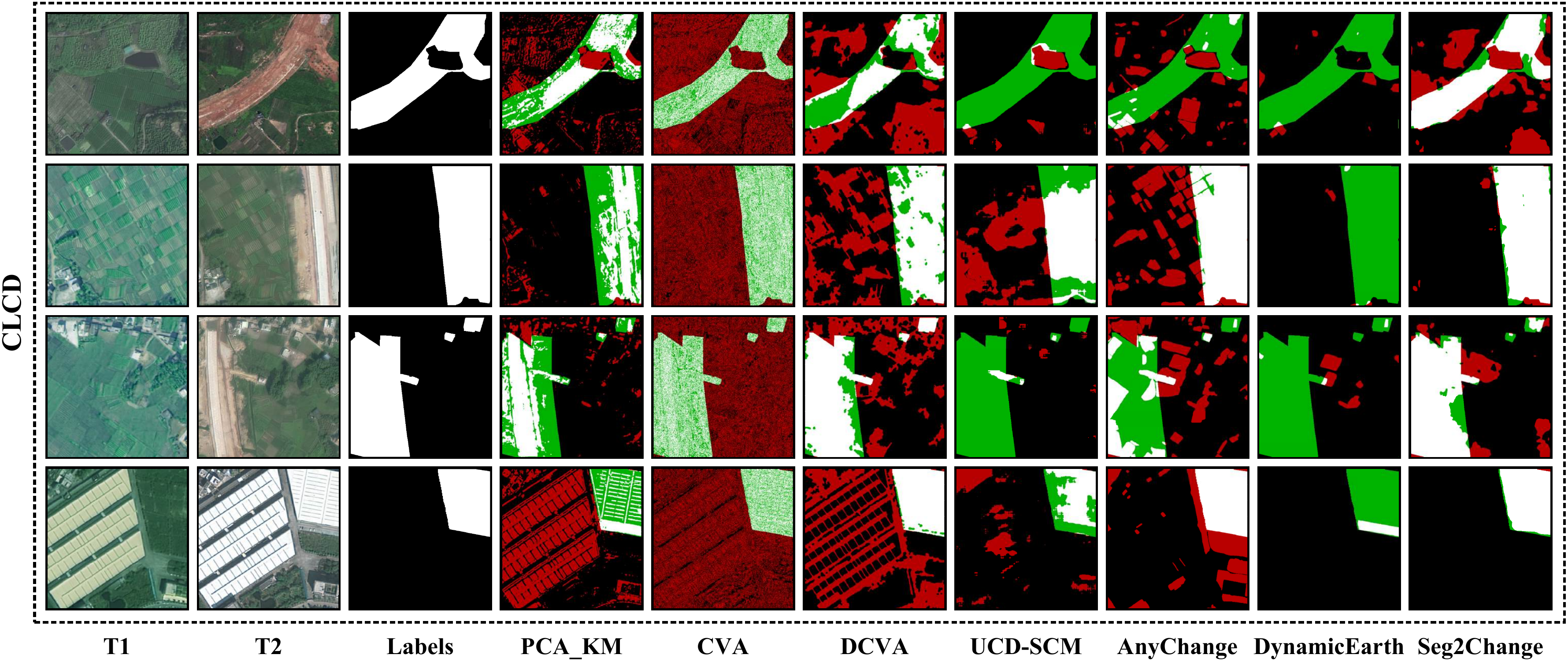}
\caption{Qualitative comparison of several representative UCD and OVCD methods on CLCD.}
\label{Binary_Vis2}
\end{figure*}
\begin{figure*}[t]
\centering
\includegraphics[width=0.92\textwidth]{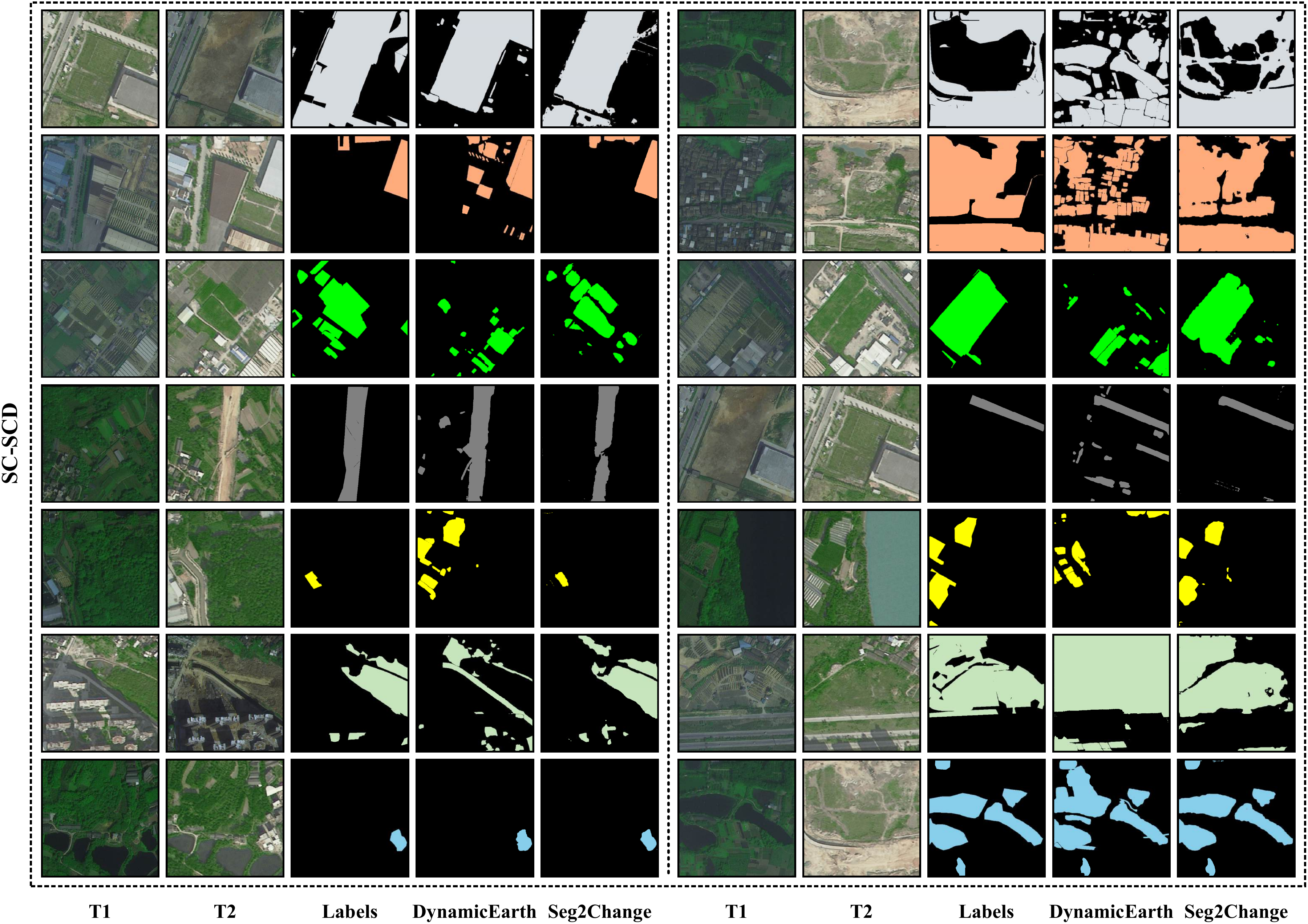}
\caption{Qualitative comparison of the state-of-the-art OVCD method DynamicEarth on a semantic change detection dataset, \textit{i.e.}, SC-SCD.  Black represents non-change.  Color rendering: \textcolor[HTML]{D7DCE0}{\textit{"Bareland"}}, \textcolor[HTML]{FEAC7C}{\textit{"Building"}}, \textcolor[HTML]{00FF00}{\textit{"Farmland"}}, \textcolor[HTML]{808080}{\textit{"Road"}}, \textcolor[HTML]{FFFF00}{\textit{"Structure"}}, \textcolor[HTML]{C6E5BC}{\textit{"Vegetation"}}, \textcolor[HTML]{88CFEB}{\textit{"Water"}}.}
\label{Semantic_Vis1}
\end{figure*}

\section{A Category-Agnostic CD Dataset}
\label{CA-CDD}

Change detection is a fundamental task in remote sensing that aims to analyze bi-temporal imagery to identify both the locations and types of changes. However, existing datasets are typically constrained to a limited set of predefined change categories. For example, WHU-CD \cite{whucd} and LEVIR-CD \cite{levircd} primarily target building changes, while DSIFN \cite{DSIFN} and CLCD \cite{clcd} concentrate on land-cover variations. SYSU-CD \cite{sysu} extends the scope to some extent by including categories such as newly constructed urban structures, vegetation changes, and road expansion. However, it still primarily focuses on urbanization-related changes. Semantic change detection datasets are the closest to category-agnostic ones, yet they differ in several key aspects. Specifically, they usually provide three forms of annotations: semantic segmentation maps at T1 and T2, along with a category-restricted binary change map. These annotations are confined to predefined classes (\textit{e.g.}, SECOND \cite{second} includes building, tree, water, low vegetation, bareland, playground), and any categories beyond this scope are not labeled (\textit{e.g.}, road or building renovation). Our CA-CDD dataset is constructed using initial bi-temporal images and coarse change maps from the SECOND training set (2968 pairs), JL1-CD \cite{liu2025jl1} training set (1000 pairs), and CNAM-CD \cite{rs15092464} training set (1000 pairs). We further refine these coarse annotations through re-labeling to achieve category-agnostic change representations. Importantly, the SECOND test set is exclusively used for evaluation, and no test images from SECOND are included in CA-CDD. Fig. \ref{Datasets_Test} presents a visualization of the evaluation datasets, while Fig. \ref{Datasets_Train} illustrates the annotation process of our CA-CDD dataset. As summarized in Table \ref{tab:cacdd}, we report the collection sites, number of test image pairs, image sizes, resolutions, and task types for all datasets. 

% All images are uniformly cropped to 512 × 512, encompassing both training and testing data across varying resolutions.

\section{More Qualitative Results}
\label{More_Qualitative_Results}
\begin{figure*}[t]
\centering
\includegraphics[width=0.92\textwidth]{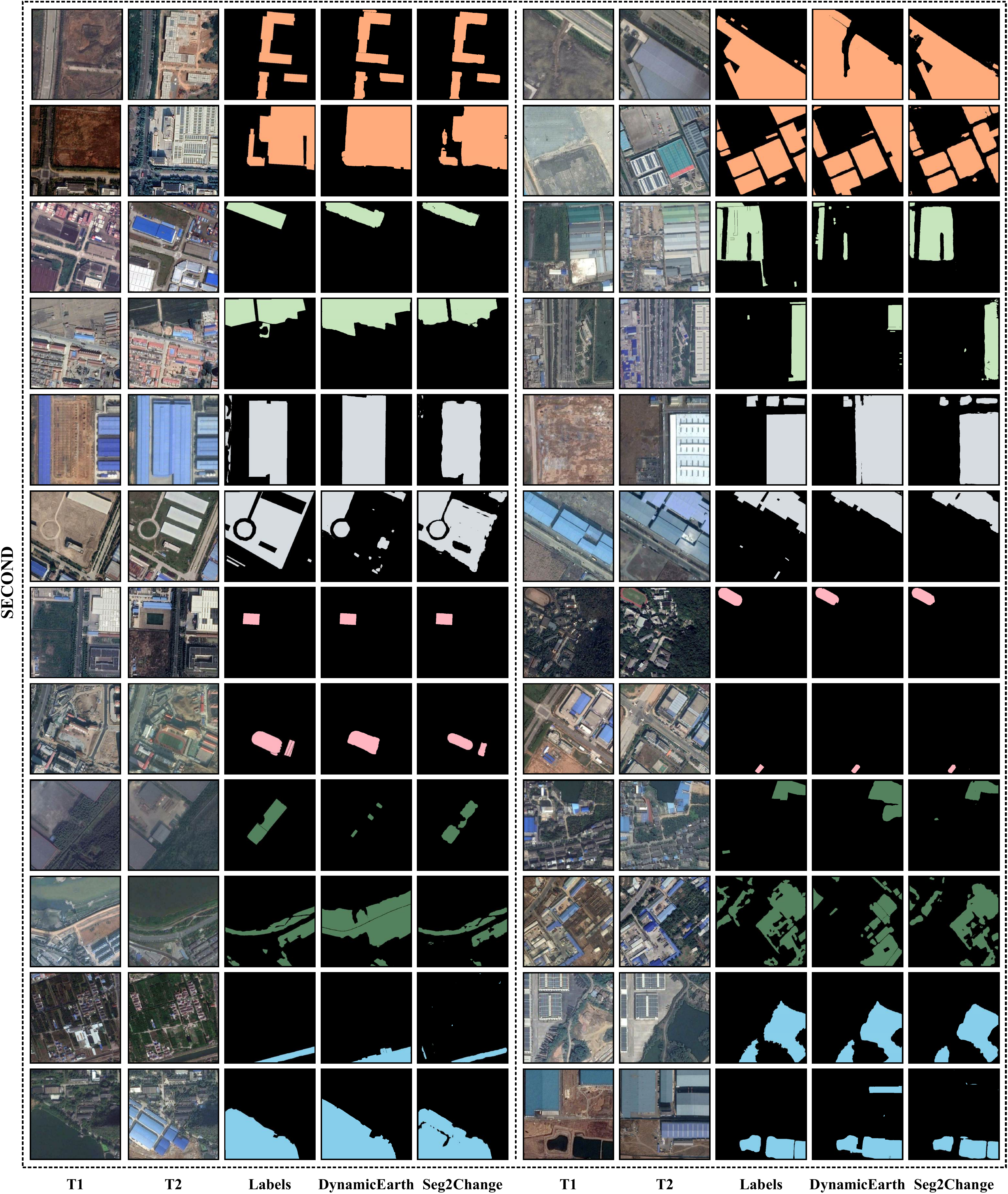}
\caption{Qualitative comparison of the state-of-the-art OVCD method DynamicEarth on a semantic change detection dataset, \textit{i.e.}, SECOND.  Black represents non-change.  Color rendering: \textcolor[HTML]{FFAB7C}{\textit{"Building"}}, \textcolor[HTML]{C7E5BD}{\textit{"Vegetation"}}, \textcolor[HTML]{D7DCE1}{\textit{"Bareland"}}, \textcolor[HTML]{FFB5C0}{\textit{"Playground"}}, \textcolor[HTML]{54815E}{\textit{"Tree"}}, \textcolor[HTML]{87CEEB}{\textit{"Water"}}.}
\label{Semantic_Vis2}
\end{figure*}

To qualitatively compare our method with previous UCD and OVCD approaches \cite{pca_km, cva, dcva, ucd_scm, anychange, li2025dynamic_earth}, we present representative samples selected from six datasets, as illustrated in Fig. \ref{Binary_Vis1}–\ref{Semantic_Vis2}. (1) Fig. \ref{Binary_Vis1}–\ref{Binary_Vis2} show that, in the open-vocabulary binary change detection task, our proposed Seg2Change outperforms all competing methods across various scenarios, including small, large, dense, and sparse changes. Meanwhile, our method directly processes 512$\times$512 images, thereby avoiding the adverse effects introduced by image cropping. (2) Fig. \ref{Semantic_Vis1} demonstrates that mask proposal based methods, such as DynamicEarth (M-C-I or I-M-C) \cite{li2025dynamic_earth}, suffer from accumulated segmentation errors originating from the mask generator. As illustrated in the example on the right side of the second row in Fig. \ref{Semantic_Vis1}, the change detection map produced by DynamicEarth appears fragmented. (3) Fig. \ref{Semantic_Vis2} indicates that Seg2Change enables more precise change identification. This improvement is attributed to our category-agnostic change map decision strategy, which replaces conventional fixed-threshold change determination.

\clearpage

\bibliographystyle{ACM-Reference-Format}
\bibliography{main}

@article{li2025dynamic_earth,
      title={DynamicEarth: How Far are We from Open-Vocabulary Change Detection?}, 
      author={Kaiyu Li and Xiangyong Cao and Yupeng Deng and Chao Pang and Zepeng Xin and Deyu Meng and Zhi Wang},
      journal={ArXiv},
      year={2025},
      volume={abs/2501.12931}
}

@article{dinov2,
  title={Dinov2: Learning robust visual features without supervision},
  author={Oquab, Maxime and Darcet, Timoth{\'e}e and Moutakanni, Th{\'e}o and Vo, Huy and Szafraniec, Marc and Khalidov, Vasil and Fernandez, Pierre and Haziza, Daniel and Massa, Francisco and El-Nouby, Alaaeldin and others},
  journal={ArXiv},
  year={2023},
  volume={abs/2304.07193}
}

@ARTICLE{whucd,
  author={Ji, Shunping and Wei, Shiqing and Lu, Meng},
  journal={IEEE Transactions on Geoscience and Remote Sensing}, 
  title={Fully Convolutional Networks for Multisource Building Extraction From an Open Aerial and Satellite Imagery Data Set}, 
  year={2019},
  volume={57},
  number={1},
  pages={574-586},
}

@Article{levircd,
    AUTHOR = {Chen, Hao and Shi, Zhenwei},
    TITLE = {A Spatial-Temporal Attention-Based Method and a New Dataset for Remote Sensing Image Change Detection},
    JOURNAL = {Remote Sensing},
    VOLUME = {12},
    YEAR = {2020},
    NUMBER = {10},
    ARTICLE-NUMBER = {1662},
    ISSN = {2072-4292},
}

@article{DSIFN,
    title = {A deeply supervised image fusion network for change detection in high resolution bi-temporal remote sensing images},
    journal = {ISPRS Journal of Photogrammetry and Remote Sensing},
    volume = {166},
    pages = {183-200},
    year = {2020},
    issn = {0924-2716},
    author = {Chenxiao Zhang and Peng Yue and Deodato Tapete and Liangcun Jiang and Boyi Shangguan and Li Huang and Guangchao Liu}
}

@ARTICLE{clcd,
    author={Liu, Mengxi and Chai, Zhuoqun and Deng, Haojun and Liu, Rong},
    journal={IEEE J. Sel. Topics Appl. Earth Observ. Remote Sens.}, 
    title={A CNN-Transformer Network With Multiscale Context Aggregation for Fine-Grained Cropland Change Detection}, 
    year={2022},
    volume={15},
    number={},
    pages={4297-4306},
    doi={10.1109/JSTARS.2022.3177235}
}

@article{second,
  author       = {Kunping Yang and
                  Gui{-}Song Xia and
                  Zicheng Liu and
                  Bo Du and
                  Wen Yang and
                  Marcello Pelillo},
  title        = {Asymmetric Siamese Networks for Semantic Change Detection},
  journal      = {CoRR},
  volume       = {abs/2010.05687},
  year         = {2020},
  eprinttype    = {arXiv},
  eprint       = {2010.05687},
}

@ARTICLE{pca_km,
  author={Celik, Turgay},
  journal={IEEE Geoscience and Remote Sensing Letters}, 
  title={Unsupervised Change Detection in Satellite Images Using Principal Component Analysis and $k$-Means Clustering}, 
  year={2009},
  volume={6},
  number={4},
  pages={772-776},
  doi={10.1109/LGRS.2009.2025059}
}

@article{cva,
  title={Automatic analysis of the difference image for unsupervised change detection},
  author={Bruzzone, Lorenzo and Prieto, Diego F},
  journal={IEEE Transactions on Geoscience and Remote sensing},
  volume={38},
  number={3},
  pages={1171--1182},
  year={2002},
  publisher={IEEE}
}

@article{dcva,
  title={Unsupervised change detection using convolutional-autoencoder multiresolution features},
  author={Bergamasco, Luca and Saha, Sudipan and Bovolo, Francesca and Bruzzone, Lorenzo},
  journal={IEEE Transactions on Geoscience and Remote Sensing},
  volume={60},
  pages={1--19},
  year={2022},
  publisher={IEEE}
}

@inproceedings{ucd_scm,
  title={Segment change model (scm) for unsupervised change detection in vhr remote sensing images: a case study of buildings},
  author={Tan, Xiaoliang and Chen, Guanzhou and Wang, Tong and Wang, Jiaqi and Zhang, Xiaodong},
  booktitle={Proc. IEEE Int. Geoscience And Remote Sensing Sym. (IGARSS)},
  pages={8577--8580},
  year={2024},
}

@article{anychange,
  title={Segment any change},
  author={Zheng, Zhuo and Zhong, Yanfei and Zhang, Liangpei and Ermon, Stefano},
  journal={Proc. Adv. Neural Inf. Process. Syst. (NIPS)},
  volume={37},
  pages={81204--81224},
  year={2024}
}

@article{instceg,
  title={SemiCD-VL: Visual-language model guidance makes better semi-supervised change detector},
  author={Li, Kaiyu and Cao, Xiangyong and Deng, Yupeng and Song, Jiayi and Liu, Junmin and Meng, Deyu and Wang, Zhi},
  journal={IEEE Transactions on Geoscience and Remote Sensing},
  year={2024},
  publisher={IEEE}
}

@ARTICLE{4310076,
  author={Otsu, Nobuyuki},
  journal={IEEE Transactions on Systems, Man, and Cybernetics}, 
  title={A Threshold Selection Method from Gray-Level Histograms}, 
  year={1979},
  volume={9},
  number={1},
  pages={62-66},
}

@article{TAN2025374,
title = {TripleS: Mitigating multi-task learning conflicts for semantic change detection in high-resolution remote sensing imagery},
journal = {ISPRS Journal of Photogrammetry and Remote Sensing},
volume = {230},
pages = {374-401},
year = {2025},
issn = {0924-2716},
author = {Xiaoliang Tan and Guanzhou Chen and Xiaodong Zhang and Tong Wang and Jiaqi Wang and Kui Wang and Tingxuan Miao},
}

@article{li2025segearthov3,
  title={SegEarth-OV3: Exploring SAM 3 for Open-Vocabulary Semantic Segmentation in Remote Sensing Images},
  author={Li, Kaiyu and Zhang, Shengqi and Deng, Yupeng and Wang, Zhi and Meng, Deyu and Cao, Xiangyong},
  journal={arXiv preprint arXiv:2512.08730},
  year={2025}
}

@article{li2025exploring,
  title={Exploring Efficient Open-Vocabulary Segmentation in the Remote Sensing},
  author={Li, Bingyu and Dong, Haocheng and Zhang, Da and Zhao, Zhiyuan and Gao, Junyu and Li, Xuelong},
  journal={arXiv preprint arXiv:2509.12040},
  year={2025}
}

@misc{carion2025sam3segmentconcepts,
      title={SAM 3: Segment Anything with Concepts},
      author={Nicolas Carion and Laura Gustafson and Yuan-Ting Hu and Shoubhik Debnath and Ronghang Hu and Didac Suris and Chaitanya Ryali and Kalyan Vasudev Alwala and Haitham Khedr and Andrew Huang and Jie Lei and Tengyu Ma and Baishan Guo and Arpit Kalla and Markus Marks and Joseph Greer and Meng Wang and Peize Sun and Roman Rädle and Triantafyllos Afouras and Effrosyni Mavroudi and Katherine Xu and Tsung-Han Wu and Yu Zhou and Liliane Momeni and Rishi Hazra and Shuangrui Ding and Sagar Vaze and Francois Porcher and Feng Li and Siyuan Li and Aishwarya Kamath and Ho Kei Cheng and Piotr Dollár and Nikhila Ravi and Kate Saenko and Pengchuan Zhang and Christoph Feichtenhofer},
      year={2025},
      eprint={2511.16719},
      archivePrefix={arXiv},
      primaryClass={cs.CV}
}

@inproceedings{li2025segearthov,
  title={Segearth-ov: Towards training-free open-vocabulary segmentation for remote sensing images},
  author={Li, Kaiyu and Liu, Ruixun and Cao, Xiangyong and Bai, Xueru and Zhou, Feng and Meng, Deyu and Wang, Zhi},
  booktitle={Proceedings of the Computer Vision and Pattern Recognition Conference},
  pages={10545--10556},
  year={2025}
}

@article{liu2025jl1,
  title={JL1-CD: A New Benchmark for Remote Sensing Change Detection and a Robust Multi-Teacher Knowledge Distillation Framework},
  author={Liu, Ziyuan and Zhu, Ruifei and Gao, Long and Zhou, Yuanxiu and Ma, Jingyu and Gu, Yuantao},
  journal={arXiv preprint arXiv:2502.13407},
  year={2025}
}

@Article{rs15092464,
AUTHOR = {Zhou, Yanpeng and Wang, Jinjie and Ding, Jianli and Liu, Bohua and Weng, Nan and Xiao, Hongzhi},
TITLE = {SIGNet: A Siamese Graph Convolutional Network for Multi-Class Urban Change Detection},
JOURNAL = {Remote Sensing},
VOLUME = {15},
YEAR = {2023},
NUMBER = {9},
ARTICLE-NUMBER = {2464},
}

@article{sam1,
  title={Segment Anything},
  author={Kirillov, Alexander and Mintun, Eric and Ravi, Nikhila and Mao, Hanzi and Rolland, Chloe and Gustafson, Laura and Xiao, Tete and Whitehead, Spencer and Berg, Alexander C. and Lo, Wan-Yen and Doll{\'a}r, Piotr and Girshick, Ross},
  journal={arXiv:2304.02643},
  year={2023}
}

@article{sam2,
  title={SAM 2: Segment Anything in Images and Videos},
  author={Ravi, Nikhila and Gabeur, Valentin and Hu, Yuan-Ting and Hu, Ronghang and Ryali, Chaitanya and Ma, Tengyu and Khedr, Haitham and R{\"a}dle, Roman and Rolland, Chloe and Gustafson, Laura and Mintun, Eric and Pan, Junting and Alwala, Kalyan Vasudev and Carion, Nicolas and Wu, Chao-Yuan and Girshick, Ross and Doll{\'a}r, Piotr and Feichtenhofer, Christoph},
  journal={arXiv preprint arXiv:2408.00714},
  url={https://arxiv.org/abs/2408.00714},
  year={2024}
}

@inproceedings{APE,
  title={Aligning and Prompting Everything All at Once for Universal Visual Perception},
  author={Shen, Yunhang and Fu, Chaoyou and Chen, Peixian and Zhang, Mengdan and Li, Ke and Sun, Xing and Wu, Yunsheng and Lin, Shaohui and Ji, Rongrong},
  journal={CVPR},
  year={2024}
}

@article{MISHRA2020133,
    title = {Land use and land cover change detection using geospatial techniques in the Sikkim Himalaya, India},
    journal = {The Egyptian Journal of Remote Sensing and Space Science},
    volume = {23},
    number = {2},
    pages = {133-143},
    year = {2020},
    issn = {1110-9823},
    author = {Prabuddh Kumar Mishra and Aman Rai and Suresh Chand Rai},
}

@article{doi:10.1080/014311600210128,
    author = {C. A. Mucher, K. T. Steinnocher, F. P. Kressler and C. Heunks},
    title = {Land cover characterization and change detection for environmental monitoring of pan-Europe},
    journal = {International Journal of Remote Sensing},
    volume = {21},
    number = {6-7},
    pages = {1159--1181},
    year = {2000},
    publisher = {Taylor \& Francis},
    doi = {10.1080/014311600210128},
}

@article{SONG201426,
    title = {Remote sensing of alpine lake water environment changes on the Tibetan Plateau and surroundings: A review},
    journal = {ISPRS Journal of Photogrammetry and Remote Sensing},
    volume = {92},
    pages = {26-37},
    year = {2014},
    issn = {0924-2716},
    author = {Chunqiao Song and Bo Huang and Linghong Ke and Keith S. Richards},
}

@ARTICLE{8641484,
  author={Gao, Feng and Wang, Xiao and Gao, Yunhao and Dong, Junyu and Wang, Shengke},
  journal={IEEE Geoscience and Remote Sensing Letters}, 
  title={Sea Ice Change Detection in SAR Images Based on Convolutional-Wavelet Neural Networks}, 
  year={2019},
  volume={16},
  number={8},
  pages={1240-1244}
}

@ARTICLE{6297453,
  author={Giustarini, Laura and Hostache, Renaud and Matgen, Patrick and Schumann, Guy J.-P. and Bates, Paul D. and Mason, David C.},
  journal={IEEE Transactions on Geoscience and Remote Sensing}, 
  title={A Change Detection Approach to Flood Mapping in Urban Areas Using TerraSAR-X}, 
  year={2013},
  volume={51},
  number={4},
  pages={2417-2430}
}

@article{Xu2019BuildingDD,
  title={Building Damage Detection in Satellite Imagery Using Convolutional Neural Networks},
  author={Joseph Z. Xu and Wenhan Lu and Zebo Li and Pranav Khaitan and Valeriya Zaytseva},
  journal={ArXiv},
  year={2019},
  volume={abs/1910.06444}
}

@inproceedings{NIPS2012_c399862d,
 author = {Krizhevsky, Alex and Sutskever, Ilya and Hinton, Geoffrey E},
 booktitle = {Proc. Adv. Neural Inf. Process. Syst. (NIPS)},
 title = {ImageNet Classification with Deep Convolutional Neural Networks},
 volume = {25},
 year = {2012}
}

@INPROCEEDINGS{7780459,
  author={He, Kaiming and Zhang, Xiangyu and Ren, Shaoqing and Sun, Jian},
  booktitle={Proc. IEEE/CVF Conf. Comput. Vis. Pattern Recognit. (CVPR)}, 
  title={Deep Residual Learning for Image Recognition}, 
  year={2016},
  volume={},
  number={},
  pages={770-778}
}

@ARTICLE{7478072,
  author={Shelhamer, Evan and Long, Jonathan and Darrell, Trevor},
  journal={IEEE Transactions on Pattern Analysis and Machine Intelligence}, 
  title={Fully Convolutional Networks for Semantic Segmentation}, 
  year={2017},
  volume={39},
  number={4},
  pages={640-651}
}

@inproceedings{cheng2021mask2former,
  title={Masked-attention Mask Transformer for Universal Image Segmentation},
  author={Bowen Cheng and Ishan Misra and Alexander G. Schwing and Alexander Kirillov and Rohit Girdhar},
  journal={CVPR},
  year={2022}
}

@inproceedings{clip,
  title={Learning transferable visual models from natural language supervision},
  author={Radford, Alec and Kim, Jong Wook and Hallacy, Chris and Ramesh, Aditya and Goh, Gabriel and Agarwal, Sandhini and Sastry, Girish and Askell, Amanda and Mishkin, Pamela and Clark, Jack and others},
  booktitle={Proc. Int. Conf. Machine Learning. (ICML)},
  pages={8748--8763},
  year={2021}
}

@article{SC-SCD,
title = {TripleS: Mitigating multi-task learning conflicts for semantic change detection in high-resolution remote sensing imagery},
journal = {ISPRS Journal of Photogrammetry and Remote Sensing},
volume = {230},
pages = {374-401},
year = {2025},
issn = {0924-2716},
doi = {https://doi.org/10.1016/j.isprsjprs.2025.09.019},
url = {https://www.sciencedirect.com/science/article/pii/S0924271625003776},
author = {Xiaoliang Tan and Guanzhou Chen and Xiaodong Zhang and Tong Wang and Jiaqi Wang and Kui Wang and Tingxuan Miao}
}

@InProceedings{pmlr-v139-jia21b,
title = 	 {Scaling Up Visual and Vision-Language Representation Learning With Noisy Text Supervision},
author =       {Jia, Chao and Yang, Yinfei and Xia, Ye and Chen, Yi-Ting and Parekh, Zarana and Pham, Hieu and Le, Quoc and Sung, Yun-Hsuan and Li, Zhen and Duerig, Tom},
booktitle = 	 {Proc. Int. Conf. Machine Learning. (ICML)},
pages = 	 {4904--4916},
year = 	 {2021},
volume = 	 {139},
month = 	 {18--24 Jul},
}

@inproceedings{li2022blip,
      title={BLIP: Bootstrapping Language-Image Pre-training for Unified Vision-Language Understanding and Generation}, 
      author={Junnan Li and Dongxu Li and Caiming Xiong and Steven Hoi},
      year={2022},
      booktitle={ICML},
}

@inproceedings{Flamingo,
 author = {Alayrac, Jean-Baptiste and Donahue, Jeff and Luc, Pauline and Miech, Antoine and Barr, Iain and Hasson, Yana and Lenc, Karel and Mensch, Arthur and Millican, Katherine and Reynolds, Malcolm and Ring, Roman and Rutherford, Eliza and Cabi, Serkan and Han, Tengda and Gong, Zhitao and Samangooei, Sina and Monteiro, Marianne and Menick, Jacob L and Borgeaud, Sebastian and Brock, Andy and Nematzadeh, Aida and Sharifzadeh, Sahand and Bi\'{n}kowski, Miko\l aj and Barreira, Ricardo and Vinyals, Oriol and Zisserman, Andrew and Simonyan, Kar\'{e}n},
 booktitle = {Advances in Neural Information Processing Systems},
 editor = {S. Koyejo and S. Mohamed and A. Agarwal and D. Belgrave and K. Cho and A. Oh},
 pages = {23716--23736},
 publisher = {Curran Associates, Inc.},
 title = {Flamingo: a Visual Language Model for Few-Shot Learning},
 volume = {35},
 year = {2022}
}

@article{liu2023grounding,
  title={Grounding dino: Marrying dino with grounded pre-training for open-set object detection},
  author={Liu, Shilong and Zeng, Zhaoyang and Ren, Tianhe and Li, Feng and Zhang, Hao and Yang, Jie and Li, Chunyuan and Yang, Jianwei and Su, Hang and Zhu, Jun and others},
  journal={arXiv preprint arXiv:2303.05499},
  year={2023}
}

@article{liu2024remoteclip,
  title={Remoteclip: A vision language foundation model for remote sensing},
  author={Liu, Fan and Chen, Delong and Guan, Zhangqingyun and Zhou, Xiaocong and Zhu, Jiale and Ye, Qiaolin and Fu, Liyong and Zhou, Jun},
  journal={IEEE Transactions on Geoscience and Remote Sensing},
  volume={62},
  pages={1--16},
  year={2024},
  publisher={IEEE}
}

@article{rskt_seg,
  title={Exploring Efficient Open-Vocabulary Segmentation in the Remote Sensing},
  author={Li, Bingyu and Dong, Haocheng and Zhang, Da and Zhao, Zhiyuan and Gao, Junyu and Li, Xuelong},
  journal={arXiv preprint arXiv:2509.12040},
  year={2025}
}

@article{Chen2023Exchange,
    title = {Exchange means change: An unsupervised single-temporal change detection framework based on intra- and inter-image patch exchange},
    author = {Hongruixuan Chen and Jian Song and Chen Wu and Bo Du and Naoto Yokoya},
    journal = {ISPRS Journal of Photogrammetry and Remote Sensing},
    volume = {206},
    pages = {87-105},
    year = {2023},
    issn = {0924-2716},
    doi = {https://doi.org/10.1016/j.isprsjprs.2023.11.004}
}

@article{zhao2023fast,
  title={Fast Segment Anything},
  author={Xu Zhao and Wenchao Ding and Yongqi An and Yinglong Du and Tao Yu and Min Li and Ming Tang and Jinqiao Wang},
  journal={ArXiv},
  year={2023},
  volume={2306.12156}
}

@ARTICLE{4039609,
  author={Bovolo, Francesca and Bruzzone, Lorenzo},
  journal={IEEE Transactions on Geoscience and Remote Sensing}, 
  title={A Theoretical Framework for Unsupervised Change Detection Based on Change Vector Analysis in the Polar Domain}, 
  year={2007},
  volume={45},
  number={1},
  pages={218-236},
  doi={10.1109/TGRS.2006.885408}}

@article{DU2020278,
title = {An improved change detection approach using tri-temporal logic-verified change vector analysis},
journal = {ISPRS Journal of Photogrammetry and Remote Sensing},
volume = {161},
pages = {278-293},
year = {2020},
issn = {0924-2716},
doi = {https://doi.org/10.1016/j.isprsjprs.2020.01.026},
author = {Peijun Du and Xin Wang and Dongmei Chen and Sicong Liu and Cong Lin and Yaping Meng},
}

@article{NIELSEN19981,
title = {Multivariate Alteration Detection (MAD) and MAF Postprocessing in Multispectral, Bitemporal Image Data: New Approaches to Change Detection Studies},
journal = {Remote Sensing of Environment},
volume = {64},
number = {1},
pages = {1-19},
year = {1998},
issn = {0034-4257},
doi = {https://doi.org/10.1016/S0034-4257(97)00162-4},
url = {https://www.sciencedirect.com/science/article/pii/S0034425797001624},
author = {Allan A. Nielsen and Knut Conradsen and James J. Simpson},
}

@article{Deng01082008,
author = {J. S. Deng and K. Wang and Y. H. Deng and G. J. Qi},
title = {PCA‐based land‐use change detection and analysis using multitemporal and multisensor satellite data},
journal = {International Journal of Remote Sensing},
volume = {29},
number = {16},
pages = {4823--4838},
year = {2008},
publisher = {Taylor \& Francis},
doi = {10.1080/01431160801950162},
}

@ARTICLE{5196726,
  author={Celik, Turgay},
  journal={IEEE Geoscience and Remote Sensing Letters}, 
  title={Unsupervised Change Detection in Satellite Images Using Principal Component Analysis and $k$-Means Clustering}, 
  year={2009},
  volume={6},
  number={4},
  pages={772-776},
  doi={10.1109/LGRS.2009.2025059}}

@ARTICLE{6553145,
  author={Wu, Chen and Du, Bo and Zhang, Liangpei},
  journal={IEEE Transactions on Geoscience and Remote Sensing}, 
  title={Slow Feature Analysis for Change Detection in Multispectral Imagery}, 
  year={2014},
  volume={52},
  number={5},
  pages={2858-2874},
  doi={10.1109/TGRS.2013.2266673}}

@article{CHEN202399,
title = {Fourier domain structural relationship analysis for unsupervised multimodal change detection},
journal = {ISPRS Journal of Photogrammetry and Remote Sensing},
volume = {198},
pages = {99-114},
year = {2023},
issn = {0924-2716},
doi = {https://doi.org/10.1016/j.isprsjprs.2023.03.004},
url = {https://www.sciencedirect.com/science/article/pii/S092427162300062X},
author = {Hongruixuan Chen and Naoto Yokoya and Marco Chini},
}

@article{XIAN20091133,
title = {Updating the 2001 National Land Cover Database land cover classification to 2006 by using Landsat imagery change detection methods},
journal = {Remote Sensing of Environment},
volume = {113},
number = {6},
pages = {1133-1147},
year = {2009},
issn = {0034-4257},
doi = {https://doi.org/10.1016/j.rse.2009.02.004},
url = {https://www.sciencedirect.com/science/article/pii/S0034425709000340},
author = {George Xian and Collin Homer and Joyce Fry},
}

@ARTICLE{4539638,
  author={Bovolo, Francesca and Bruzzone, Lorenzo and Marconcini, Mattia},
  journal={IEEE Transactions on Geoscience and Remote Sensing}, 
  title={A Novel Approach to Unsupervised Change Detection Based on a Semisupervised SVM and a Similarity Measure}, 
  year={2008},
  volume={46},
  number={7},
  pages={2070-2082},
  doi={10.1109/TGRS.2008.916643}}

@ARTICLE{6841049,
  author={Hoberg, Thorsten and Rottensteiner, Franz and Feitosa, Raul Queiroz and Heipke, Christian},
  journal={IEEE Transactions on Geoscience and Remote Sensing}, 
  title={Conditional Random Fields for Multitemporal and Multiscale Classification of Optical Satellite Imagery}, 
  year={2015},
  volume={53},
  number={2},
  pages={659-673},
  doi={10.1109/TGRS.2014.2326886}}

@ARTICLE{1036009,
  author={Kasetkasem, T. and Varshney, P.K.},
  journal={IEEE Transactions on Geoscience and Remote Sensing}, 
  title={An image change detection algorithm based on Markov random field models}, 
  year={2002},
  volume={40},
  number={8},
  pages={1815-1823},
  doi={10.1109/TGRS.2002.802498}}

@article{HUSSAIN201391,
title = {Change detection from remotely sensed images: From pixel-based to object-based approaches},
journal = {ISPRS Journal of Photogrammetry and Remote Sensing},
volume = {80},
pages = {91-106},
year = {2013},
issn = {0924-2716},
doi = {https://doi.org/10.1016/j.isprsjprs.2013.03.006},
url = {https://www.sciencedirect.com/science/article/pii/S0924271613000804},
author = {Masroor Hussain and Dongmei Chen and Angela Cheng and Hui Wei and David Stanley},
}

@article{GILYEPES201677,
title = {Description and validation of a new set of object-based temporal geostatistical features for land-use/land-cover change detection},
journal = {ISPRS Journal of Photogrammetry and Remote Sensing},
volume = {121},
pages = {77-91},
year = {2016},
issn = {0924-2716},
doi = {https://doi.org/10.1016/j.isprsjprs.2016.08.010},
url = {https://www.sciencedirect.com/science/article/pii/S0924271616303434},
author = {Jose L. Gil-Yepes and Luis A. Ruiz and Jorge A. Recio and Ángel Balaguer-Beser and Txomin Hermosilla},
}

@ARTICLE{sysu,
  author={Shi, Qian and Liu, Mengxi and Li, Shengchen and Liu, Xiaoping and Wang, Fei and Zhang, Liangpei},
  journal={IEEE Transactions on Geoscience and Remote Sensing}, 
  title={A Deeply Supervised Attention Metric-Based Network and an Open Aerial Image Dataset for Remote Sensing Change Detection}, 
  year={2021},
  volume={},
  number={},
  pages={1-16},
  doi={10.1109/TGRS.2021.3085870}}

@ARTICLE{1395984,
  author={Radke, R.J. and Andra, S. and Al-Kofahi, O. and Roysam, B.},
  journal={IEEE Transactions on Image Processing}, 
  title={Image change detection algorithms: a systematic survey}, 
  year={2005},
  volume={14},
  number={3},
  pages={294-307},
  doi={10.1109/TIP.2004.838698}}

@ARTICLE{10234560,
  author={Han, Chengxi and Wu, Chen and Guo, Haonan and Hu, Meiqi and Li, Jiepan and Chen, Hongruixuan},
  journal={IEEE Journal of Selected Topics in Applied Earth Observations and Remote Sensing}, 
  title={Change Guiding Network: Incorporating Change Prior to Guide Change Detection in Remote Sensing Imagery}, 
  year={2023},
  volume={16},
  number={},
  pages={8395-8407},
  doi={10.1109/JSTARS.2023.3310208}}

@ARTICLE{10285430,
  author={Chen, ZiJian and Song, YongHong and Ma, Yue and Li, GuoFu and Wang, Rui and Hu, Hao},
  journal={IEEE Transactions on Geoscience and Remote Sensing}, 
  title={Interaction in Transformer for Change Detection in VHR Remote Sensing Images}, 
  year={2023},
  volume={61},
  number={},
  pages={1-12},
  doi={10.1109/TGRS.2023.3324025}}

@article{li2022cd,
  author={Li, Zhenglai and Tang, Chang and Wang, Lizhe and Zomaya, Albert Y.},
  journal={IEEE Transactions on Geoscience and Remote Sensing}, 
  title={Remote Sensing Change Detection via Temporal Feature Interaction and Guided Refinement}, 
  year={2022},
  volume={60},
  number={},
  pages={1-11},
  doi={10.1109/TGRS.2022.3199502}
}

@ARTICLE{10965808,
  author={Zhan, Tao and Tian, Qiushi and Zhu, Yuanyuan and Lan, Jie and Dang, Qianlong and Gong, Maoguo},
  journal={IEEE Transactions on Geoscience and Remote Sensing}, 
  title={Difference-Aware Multiscale Feature Aggregation Network for Building Change Detection}, 
  year={2025},
  volume={63},
  number={},
  pages={1-15},
  doi={10.1109/TGRS.2025.3560977}}

@ARTICLE{10504297,
author={Ying, Zilu and Tan, Zijun and Zhai, Yikui and Jia, Xudong and Li, Wenba and Zeng, Junying and Genovese, Angelo and Piuri, Vincenzo and Scotti, Fabio},
journal={IEEE Transactions on Geoscience and Remote Sensing},
title={DGMA2-Net: A Difference-Guided Multiscale Aggregation Attention Network for Remote Sensing Change Detection},
year={2024},
volume={62},
pages={1-16},
doi={10.1109/TGRS.2024.3390206}
}

@article{jiang2025LGCANet,
  author  = {Jiang, Kaixuan and Wu, Chen},
  journal = {IEEE Transactions on Geoscience and Remote Sensing},
  title   = {LGCANet: Local–Global and Change-Aware Network via Segment Anything Model for Remote Sensing Images Change Detection},
  year    = {2025},
  volume  = {63},
  pages   = {1-13},
  doi     = {10.1109/TGRS.2025.3582784}
}

@ARTICLE{10034787, author={Feng, Yuchao and Jiang, Jiawei and Xu, Honghui and Zheng, Jianwei}, journal={IEEE Transactions on Geoscience and Remote Sensing}, title={Change Detection on Remote Sensing Images using Dual-branch Multi-level Inter-temporal Network}, year={2023}, volume={}, number={}, pages={1-1}, doi={10.1109/TGRS.2023.3241257} }

@Article{rs12101688,
    AUTHOR = {Shi, Wenzhong and Zhang, Min and Zhang, Rui and Chen, Shanxiong and Zhan, Zhao},
    TITLE = {Change Detection Based on Artificial Intelligence: State-of-the-Art and Challenges},
    JOURNAL = {Remote Sensing},
    VOLUME = {12},
    YEAR = {2020},
    NUMBER = {10},
    ARTICLE-NUMBER = {1688},
    ISSN = {2072-4292},
    DOI = {10.3390/rs12101688}
}

@article{LI2019197,
    title = {Deep learning based cloud detection for medium and high resolution remote sensing images of different sensors},
    journal = {ISPRS Journal of Photogrammetry and Remote Sensing},
    volume = {150},
    pages = {197-212},
    year = {2019},
    issn = {0924-2716},
    author = {Zhiwei Li and Huanfeng Shen and Qing Cheng and Yuhao Liu and Shucheng You and Zongyi He},
}

@INPROCEEDINGS {9710580,
author = { Liu, Ze and Lin, Yutong and Cao, Yue and Hu, Han and Wei, Yixuan and Zhang, Zheng and Lin, Stephen and Guo, Baining },
booktitle = { 2021 IEEE/CVF International Conference on Computer Vision (ICCV) },
title = {{ Swin Transformer: Hierarchical Vision Transformer using Shifted Windows }},
year = {2021},
volume = {},
ISSN = {},
pages = {9992-10002},
doi = {10.1109/ICCV48922.2021.00986},
url = {https://doi.ieeecomputersociety.org/10.1109/ICCV48922.2021.00986},
publisher = {IEEE Computer Society},
address = {Los Alamitos, CA, USA},
month =Oct}

@InProceedings{Jia_M4oE_MICCAI2024,
author = { Jiang, Yufeng and Shen, Yiqing},
title = { { M4oE: A Foundation Model for Medical Multimodal Image Segmentation with Mixture of Experts } },
booktitle = {proceedings of Medical Image Computing and Computer Assisted Intervention -- MICCAI 2024},
year = {2024},
publisher = {Springer Nature Switzerland},
volume = {LNCS 15012},
month = {October},
page = {621 -- 631}
}

@article{TFIM,
author={Li, Zhenglai and Tang, Chang and Wang, Lizhe and Zomaya, Albert Y.},
journal={IEEE Transactions on Geoscience and Remote Sensing},
title={Remote Sensing Change Detection via Temporal Feature Interaction and Guided Refinement},
year={2022},
volume={60},
number={},
pages={1-11},
doi={10.1109/TGRS.2022.3199502}
}

@ARTICLE{CDEM,
author={Zhan, Tao and Tian, Qiushi and Zhu, Yuanyuan and Lan, Jie and Dang, Qianlong and Gong, Maoguo},
journal={IEEE Transactions on Geoscience and Remote Sensing},
title={Difference-Aware Multiscale Feature Aggregation Network for Building Change Detection},
year={2025},
volume={63},
number={},
pages={1-15},
doi={10.1109/TGRS.2025.3560977}}

@ARTICLE{MDFM,
author={Ying, Zilu and Tan, Zijun and Zhai, Yikui and Jia, Xudong and Li, Wenba and Zeng, Junying and Genovese, Angelo and Piuri, Vincenzo and Scotti, Fabio},
journal={IEEE Transactions on Geoscience and Remote Sensing},
title={DGMA2-Net: A Difference-Guided Multiscale Aggregation Attention Network for Remote Sensing Change Detection},
year={2024},
volume={62},
number={},
pages={1-16},
doi={10.1109/TGRS.2024.3390206}}

@misc{he2015deepresiduallearningimage,
      title={Deep Residual Learning for Image Recognition}, 
      author={Kaiming He and Xiangyu Zhang and Shaoqing Ren and Jian Sun},
      year={2015},
      eprint={1512.03385},
      archivePrefix={arXiv},
      primaryClass={cs.CV},
      url={https://arxiv.org/abs/1512.03385}, 
}

@article{unimatchv2,
  title={UniMatch V2: Pushing the Limit of Semi-Supervised Semantic Segmentation},
  author={Yang, Lihe and Zhao, Zhen and Zhao, Hengshuang},
  journal={TPAMI},
  year={2025}
}

@ARTICLE{9491802,
  author={Chen, Hao and Qi, Zipeng and Shi, Zhenwei},
  journal={IEEE Transactions on Geoscience and Remote Sensing}, 
  title={Remote Sensing Image Change Detection With Transformers}, 
  year={2022},
  volume={60},
  number={},
  pages={1-14},
  doi={10.1109/TGRS.2021.3095166}}

@INPROCEEDINGS{9883686,
  author={Bandara, Wele Gedara Chaminda and Patel, Vishal M.},
  booktitle={IGARSS 2022 - 2022 IEEE International Geoscience and Remote Sensing Symposium}, 
  title={A Transformer-Based Siamese Network for Change Detection}, 
  year={2022},
  volume={},
  number={},
  pages={207-210},
  doi={10.1109/IGARSS46834.2022.9883686}}

@article{Yuan31122022,
author = {Panli Yuan and Qingzhan Zhao and Xingbiao Zhao and Xuewen Wang and Xuefeng Long and Yuchen Zheng},
title = {A transformer-based Siamese network and an open optical dataset for semantic change detection of remote sensing images},
journal = {International Journal of Digital Earth},
volume = {15},
number = {1},
pages = {1506--1525},
year = {2022},
publisher = {Taylor \& Francis},
doi = {10.1080/17538947.2022.2111470}
}

@ARTICLE{10185449,
  author={Tang, Xu and Zhang, Tianxiang and Ma, Jingjing and Zhang, Xiangrong and Liu, Fang and Jiao, Licheng},
  journal={IEEE Transactions on Geoscience and Remote Sensing}, 
  title={WNet: W-Shaped Hierarchical Network for Remote-Sensing Image Change Detection}, 
  year={2023},
  volume={61},
  number={},
  pages={1-14},
  doi={10.1109/TGRS.2023.3296383}}

@Article{rs15082092,
AUTHOR = {Parelius, Eleonora Jonasova},
TITLE = {A Review of Deep-Learning Methods for Change Detection in Multispectral Remote Sensing Images},
JOURNAL = {Remote Sensing},
VOLUME = {15},
YEAR = {2023},
NUMBER = {8},
ARTICLE-NUMBER = {2092},
URL = {https://www.mdpi.com/2072-4292/15/8/2092},
ISSN = {2072-4292},
DOI = {10.3390/rs15082092}
}

% \appendix









% \bibliographystyle{ACM-Reference-Format}
% \bibliography{main}

\end{document}